\documentclass{article}

\usepackage{hyperref}
\usepackage{url}

\usepackage{aligned-overset}
\usepackage{bbm}
\usepackage{bm}
\usepackage{braket}
\usepackage{calrsfs}
\usepackage{centernot}
\usepackage{colonequals}
\usepackage{listofitems}
\usepackage{tikz}
\usetikzlibrary{shapes}
\usetikzlibrary{plotmarks}
\usepackage{pgfplots}
\pgfplotsset{compat=1.18}
\usepackage{microtype}
\usepackage{multirow}
\usepackage{graphicx}
\usepackage{subcaption}
\usepackage{svg}
\usepackage{wrapfig}
\usepackage{yhmath}
\usepackage{xcolor}
\usepackage[rightcaption]{sidecap}

\tikzset{
    cross/.pic = {
    \draw[rotate = 45] (-#1,0) -- (#1,0);
    \draw[rotate = 45] (0,-#1) -- (0, #1);
    }
}

\usepackage{microtype}
\usepackage{graphicx}
\usepackage{booktabs} 

\usepackage{hyperref}

\usepackage[accepted]{icml2024}

\usepackage{amsmath}
\usepackage{amssymb}
\usepackage{mathtools}
\usepackage{amsthm}

\usepackage[capitalize,noabbrev]{cleveref}

\theoremstyle{plain}
\newtheorem{theorem}{Theorem}[section]

\newtheorem{example}[theorem]{Example}
\theoremstyle{definition}
\newtheorem{definition}[theorem]{Definition}

\theoremstyle{remark}

\usepackage[textsize=tiny]{todonotes}

\icmltitlerunning{Bias-Variance-Covariance Decomposition of Kernel Scores}

\begin{document}

\twocolumn[
\icmltitle{A Bias-Variance-Covariance Decomposition \\ 
of Kernel Scores for Generative Models}




\begin{icmlauthorlist}
\icmlauthor{Sebastian G. Gruber}{dktk,dkfz,goethe}
\icmlauthor{Florian Buettner}{dktk,dkfz,goethe,fci}
\end{icmlauthorlist}

\icmlaffiliation{dkfz}{German Cancer Research Center (DKFZ), Heidelberg, Germany}
\icmlaffiliation{dktk}{German Cancer Consortium (DKTK), partner site Frankfurt/Mainz, a partnership between DKFZ and UCT Frankfurt-Marburg, Germany, Frankfurt am Main, Germany}
\icmlaffiliation{goethe}{Goethe University Frankfurt, Germany}
\icmlaffiliation{fci}{Frankfurt Cancer Institute (FCI), Germany}

\icmlcorrespondingauthor{Sebastian G. Gruber}{sebastian.gruber@dkfz.de}

\icmlkeywords{Machine Learning, Statistical Learning Theory, Bias-Variance Decomposition, Uncertainty Estimation, Generative Models, Large Language Models, Kernels, ICML}

\vskip 0.3in
]



\printAffiliationsAndNotice{}  

\begin{abstract}
Generative models, like large language models, are becoming increasingly relevant in our daily lives, yet a theoretical framework to assess their generalization behavior and uncertainty does not exist.
Particularly, the problem of uncertainty estimation is commonly solved in an ad-hoc and task-dependent manner.
For example, natural language approaches cannot be transferred to image generation.
In this paper, we introduce the first bias-variance-covariance decomposition for kernel scores.
This decomposition represents a theoretical framework from which we derive a kernel-based variance and entropy for uncertainty estimation.
We propose unbiased and consistent estimators for each quantity which only require generated samples but not the underlying model itself.
Based on the wide applicability of kernels, we demonstrate our framework via generalization and uncertainty experiments for image, audio, and language generation.
Specifically, kernel entropy for uncertainty estimation is more predictive of performance on CoQA and TriviaQA question answering datasets than existing baselines and can also be applied to closed-source models.
\end{abstract}

\section{Introduction}

\begin{figure}[ht]
\vskip 0.2in
\centering
    \begin{subfigure}{.5\textwidth}
    \centering
    \includegraphics[width=0.9\textwidth]{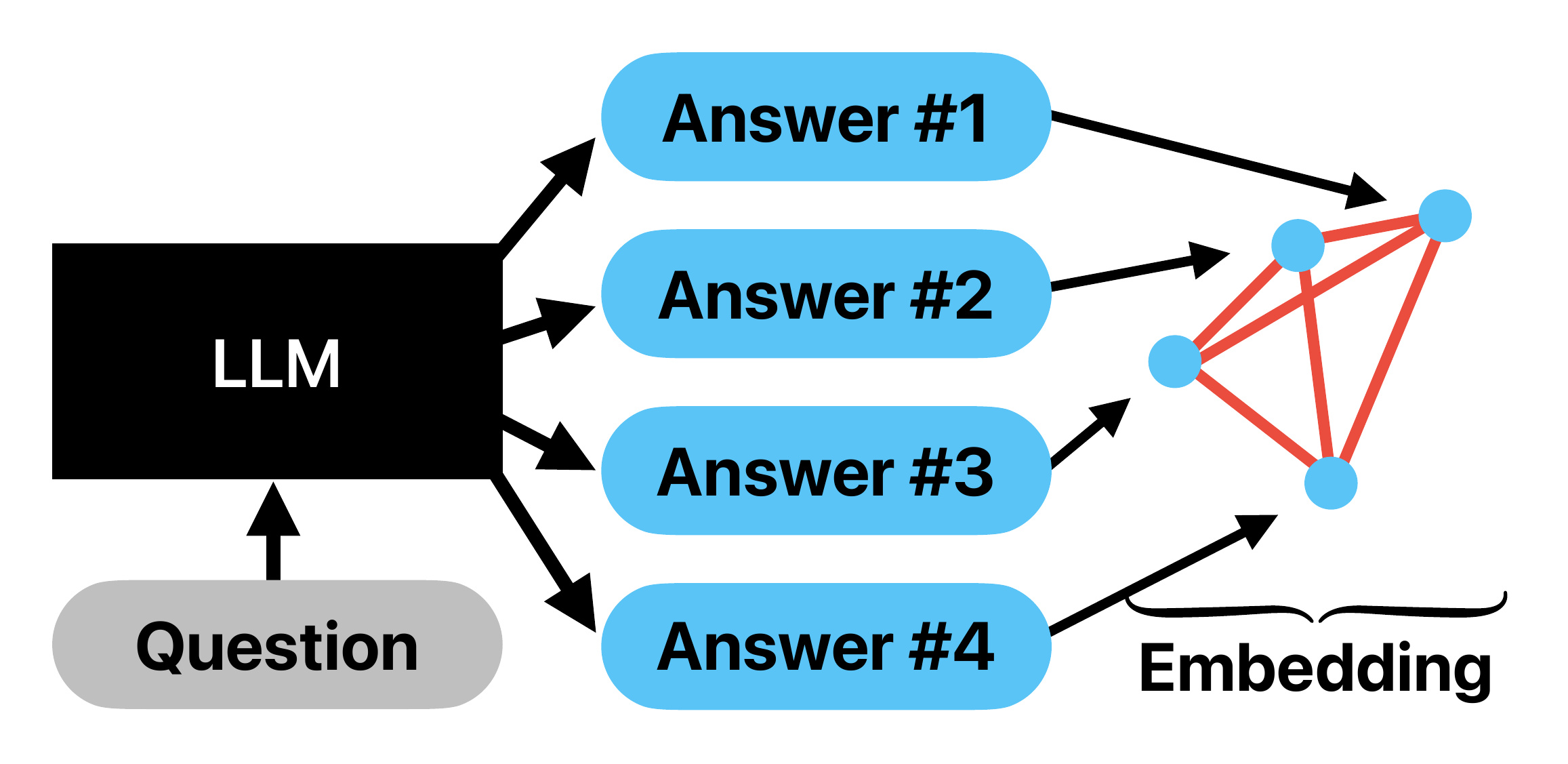}
    \end{subfigure} \\%
    \begin{subfigure}{.5\textwidth}
    \centering
    \includegraphics[width=\textwidth]{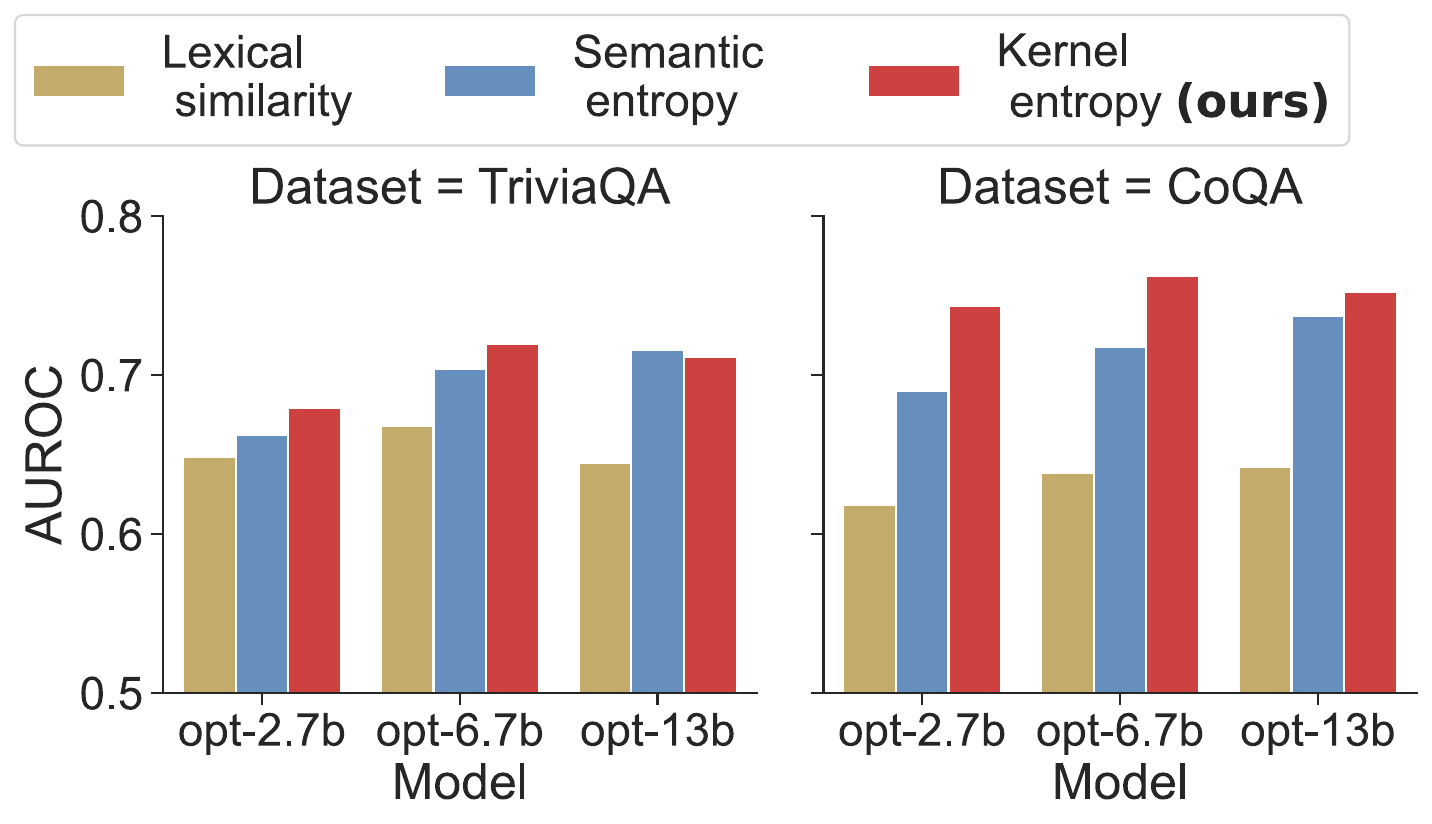}
    \end{subfigure}
\caption{
\textbf{Top:} Illustration of predictive kernel entropy for a generative model.
A kernel measures the pairwise similarities (red lines) of outputs in a vector space. The predictive kernel entropy is then the negative average kernel value.
\textbf{Bottom:} The predictive kernel entropy shows the best performance among uncertainty approaches for single-model settings (c.f. Section \ref{sec:applications_nlg}). 
}
\label{fig:unc_illus}
\vskip -0.2in
\end{figure}

In recent years, generative models have revolutionized daily lives well beyond the field of machine learning \citep{kasneci2023chatgpt, mesko2023imperative}.
These models have found applications in diverse domains, including image creation \citep{ramesh2021zero}, natural language generation \citep{openai2023gpt4}, drug discovery \citep{paul2021artificial}, and speech synthesis \citep{ning2019review}.
While generative models have demonstrated remarkable capabilities in generating data that closely resemble real-world samples, they often fall short in providing the vital and often overlooked aspect of uncertainty estimation \citep{wu2020managing}. Uncertainty estimation in machine learning is a critical component of model performance assessment and deployment \citep{hekler2023test}.
It addresses the inherent limitations and challenges associated with machine learning based decisions.
For generative models, this may include their propensity to generate improbable or nonsensical samples (``hallucinations'').
Even though uncertainty estimation methods for natural language question answering tasks exist \citep{kuhn2022semantic}, they are ad-hoc without theoretical grounding and are not transferable to other data generation tasks .

Predictive uncertainty is an informal concept, but it is implied that it relates to the prediction error without requiring access to target outcomes.
A formal approach to this is the bias-variance decomposition, a central concept in statistical learning theory \citep{bishop2006pattern, hastie2009elements, pml1Book}.
It helps to understand the generalization behavior of models and naturally raises uncertainty terms by isolating the target prediction into a bias term \citep{gruber2023uncertainty}. \\
\citet{ueda1996generalization} discovered the bias-variance-covariance decomposition of the mean squared error, which is the foundation of negative correlation learning \citep{liu1999simultaneous, liu1999ensemble, brown2004diversity} and for reducing correlations in weight averaging \citep{rame2022diverse}.
Though the bias-variance decomposition has been generalized to distributions \citep{gruber2023uncertainty}, the current theory does not include a covariance term and relies on having access to the predicted distribution.
But, many generative models only indirectly fit the training distribution by learning how to generate samples.
Others, such as large language models (LLMs), do explicitly fit the target distribution, but the prevalence of closed source models means the predictive distribution is often not available to the practitioner \citep{openai2023gpt4}.
This makes it infeasible to apply the powerful framework of the bias-variance decomposition in these cases. \\
Contrary, kernels allow to quantify differences between distributions only based on their samples without requiring access to these distributions \citep{JMLR:v13:gretton12a}.
They are used in kernel scores to assess the goodness-of-fit for predicted distributions \citep{gneitingscores}. \\
As \textbf{contribution} in this work, we...
\begin{itemize}
    \item introduce the first extension of the bias-variance-covariance decomposition beyond the mean squared error to kernel scores in Section \ref{sec:BVCD}, and propose unbiased and consistent estimators only requiring generated samples in Section \ref{sec:estimators}.
    \item examine the generalisation behavior of generative models for image and audio generation and investigate how bias, variance and kernel entropy relate to the generalisation error in Section \ref{sec:applications}. This includes evidence that mode collapse of underrepresented minority groups is expressed purely in the bias.
    \item demonstrate how kernel entropy in combination with text embeddings outperforms existing methods for estimating the uncertainty of LLMs on common question answering datasets (c.f. Figure \ref{fig:unc_illus} and Section \ref{sec:applications_nlg}).
\end{itemize}

\section{Background}

In this section, we give a brief introduction into kernel scores, followed up by other bias-variance decompositions and approaches for assessing the uncertainty in natural language generation.

\subsection{Kernel Scores}

Kernel scores are a class of loss functions for distribution predictions \citep{eaton1981method, eaton1996predictive, dawid2007geometry}.
For simplicity, we omit complex-valued kernels.
We refer to a symmetric kernel $k \colon \mathcal{Y} \times \mathcal{Y} \to \mathbb{R}$ defined on a set $\mathcal{Y}$ as \textbf{positive definite} (p.d.) if $\sum_{i=1}^n \sum_{j=1}^n a_i k(x_i, x_j) a_j > 0$ for all $x_1, \dots, x_n \in \mathcal{Y}$ and $a_1, \dots, a_n \neq 0$ with $n \in \mathbb{N}$.
\textbf{Positive semi-definite} (p.s.d.) refers to the case when only '$\geq$' holds. Assume $\mathcal{P}$ is a set of distributions defined on $\mathcal{X}$ such that for a kernel $k$ the operator
$\Braket{P | k | Q} \coloneqq \int_{\mathcal{Y}} \int_{\mathcal{Y}} k \left( x, y \right) \mathrm{d} P \left( x \right) \mathrm{d} Q \left( y \right)$ is finite for all $P, Q \in \mathcal{P}$ \citep{eaton1981method}.
It follows that $\Braket{. | k | .}$ is a symmetric bilinear form and induces the semi-norm $\lVert P \rVert_k = \sqrt{\Braket{P | k | P}}$. \\ 
A \textbf{kernel score} $S_k \colon \mathcal{P} \times \mathcal{Y} \to \mathbb{R}$ based on a p.s.d. kernel $k \colon \mathcal{Y} \times \mathcal{Y} \to \mathbb{R}$ is defined as \citep{steinwart2021strictly}
\begin{equation}
    S_k \left( P, y \right) = \left\lVert P \right\rVert_k^2 - 2 \; \mathbb{E}_{\hat{Y} \sim P} \left[ k \left( \hat{Y}, y \right) \right]. 
\label{eq:ks_def}
\end{equation}
Note that 
\citet{eaton1981method} and \citet{dawid2007geometry} use a slightly less general definition.
If $\mathcal{P}$ only consists of Borel probability measures, then the expected kernel score $\mathbb{E} \left[ S_k \left(P, Y \right) \right]$ based on a target $Y \sim Q \in \mathcal{P}$ is minimized when $P = Q$ \citep{gneitingscores}.
Following \citet{dawid2007geometry}, we refer to $-\left\lVert Q \right\rVert^2_k = \mathbb{E} \left[ S_k \left(Q, Y \right) \right]$ as the \textbf{kernel entropy} function for any $Q$.
The kernel score is connected to the \textbf{maximum mean discrepancy} (MMD) via $\operatorname{MMD}^2_k \left(P, Q \right) = \mathbb{E} \left[ S_k \left(P, Y \right) \right] + \left\lVert Q \right\rVert_k^2$ \citep{steinwart2021strictly}.
The MMD is used for non-parametric two-sample testing \citep{JMLR:v13:gretton12a} and generative image modelling \citep{li2015generative, binkowski2018demystifying}.
Compared to the MMD, kernel scores are applicable to a wider range of scenarios, since one sample of the target distribution is sufficient for evaluation.
For example, MMDs cannot be computed for question-answering pairs when there is only one answer for each question in the dataset.
We offer a more extensive discussion of related work in Appendix \ref{app:ext_MMD}.

\subsection{Bias-Variance (-Covariance) Decompositions}

\citet{ueda1996generalization} introduced the bias-variance-covariance decomposition for the mean squared error.
For a real-valued ensemble prediction $\hat{P}^{\left( n \right)} = \frac{1}{n} \sum_{i=1}^n \hat{P}_i$ with identically distributed $\hat{P}_1, \dots, \hat{P}_n$ and real-valued target $Y$ it is given by
\begin{equation}
\begin{split}
    \underbrace{\mathbb{E} \left[ \left( \hat{P}^{\left( n \right)} - Y \right)^2 \right]}_{\text{Expected Squared Error}} & = \underbrace{\mathbb{V} \left( Y \right)}_{\text{Noise}} + \underbrace{(\mathbb{E} \left[ \hat{P} \right] - \mathbb{E} \left[ Y \right])^2}_{\text{Bias}} \\
    & + \underbrace{\frac{1}{n} \mathbb{V} \left( \hat{P} \right)}_{\text{Variance}} + \underbrace{\frac{n - 1}{n} \operatorname{Cov} \left( \hat{P}, \hat{P}^\prime \right)}_{\text{Covariance}},
\label{eq:sq_bvcd}
\end{split}
\end{equation}
with $\hat{P} \coloneqq \hat{P}_1$ and $\hat{P}^\prime \coloneqq \hat{P}_2$.
We use $\mathbb{V}$ and $\operatorname{Cov}$ to denote the textbook definitions of variance and covariance for real-valued random variables \citep{capinski2004measure}.
Throughout this work, we imply that the pairwise covariance of identically distributed variables is the same.
\citet{rame2022diverse} propose an approximate bias-variance-covariance decomposition for hard-label classification but it only holds in an infinitesimal locality around the prediction.
To our best knowledge, the mean squared error is the only case so far with a non-approximated decomposition.
\citet{gruber2023uncertainty} introduce a bias-variance decomposition for loss functions of general distributions.
They demonstrated that the variance term is a meaningful measure of the model uncertainty similar to confidence scores in classification.
However, they require a loss-specific transformation of the distributions into a dual vector space and a covariance term is not given.

\subsection{Uncertainty in Natural Language Generation}

In the following, we give a brief overview of uncertainty estimation in natural language generation. \\
A common approach is predictive entropy, which is the Shannon entropy $- \int \log \hat{p} \left( y \mid x \right) \mathrm{d} \hat{p} \left( y \mid x \right)$ of the predicted distribution $\hat{p}$ given an input $x$ \citep{malinin2020uncertainty}.
For a generated token sequence $\mathbf{s} = \left(s_1, \dots, s_l \right) \in \mathbb{N}^l$ of length $l \in \mathbb{N}$ it is computed via $\sum_{i=1}^l \log \hat{p} \left( s_i \mid s_1, \dots, s_{i-1} \right)$, where $\hat{p}$ is the predicted distribution of the generating language model.
Note that the predicted distribution is not always available for closed-source models.
The computation also scales linearly with the length of the generated text, making it costly for larger text generations.
\citet{malinin2020uncertainty} propose to use length-normalisation of the predictive entropy since the Shannon entropy is systematically affected by the sequence length.
\citet{kuhn2022semantic} propose \emph{semantic entropy} to ease the computation of the predictive entropy by finding clusters of semantically similar generations.
Another approach is \emph{lexical similarity} \citep{fomicheva2020unsupervised}, which quantifies the average pairwise similarity between generated answers according to a similarity measure, like $\operatorname{RougeL}$ \citep{lin2004automatic, kuhn2022semantic}.
\citet{kadavath2022language} propose the baseline \emph{p(True)}, which asks the model itself if the generated answer is correct.
Alternative approaches exist, which require an ensemble of models \citep{lakshminarayanan2017simple, malinin2020uncertainty}.
However, ensembles are practically less relevant due to the high computational cost of training even a single model.

\section{A Bias-Variance-Covariance Decomposition of Kernel Scores}
\label{sec:BVCD}

In this section, we state our main theoretical contribution.
All proofs are presented in Appendix \ref{app:proofs}.
To highlight the similarity to the mean squared error case, we introduce the novel definitions for distributional variance and distributional covariance.
The latter also implies a distributional correlation, which we define later in Section \ref{sec:estimators}.
Note that textbook variance and covariance are based on multiplication of two scalar components.
Multiplication of two scalars is a special case of an inner product based on $k$.
Thus, we interpret  $\Braket{. | k | .}$ as a generalization of this multiplication, which directly implies the following.

\begin{definition}
    Assume we have a p.s.d. kernel $k$ and random variables $P$ and $Q$ with outcomes in a distribution space as defined above.
    We define the \textbf{distributional variance} generated by $k$ of $P$ as
    \begin{equation}
        \operatorname{Var}_k \left( P \right) = \mathbb{E} \left[ \lVert P - \mathbb{E} \left[ P \right] \rVert_k^2 \right]
    \end{equation}
    and the \textbf{distributional covariance} generated by $k$ between $P$ and $Q$ as
    \begin{equation}
        \operatorname{Cov}_k \left( P, Q \right) = \mathbb{E} \left[ \Braket{P - \mathbb{E} \left[ P \right] | k | Q - \mathbb{E} \left[ Q \right]} \right].
    \end{equation}
\end{definition}

If $P$ is deterministic, i.e. is a random variable with only one outcome, then $\operatorname{Var}_k \left( P \right) = 0$.
Further, we have $\operatorname{Cov}_k \left(P, P \right) = \operatorname{Var}_k \left( P \right)$, and, if $P$ and $Q$ are independent, then $\operatorname{Cov}_k \left(P, Q \right) = 0$.
Note that the terms \emph{kernel variance} and \emph{kernel covariance} already exist in the literature and should not be confused with our definitions \citep{gretton2003kernel}. \\
We now have the necessary tools to state our main theoretical contribution in a concise manner.

\begin{theorem}
    Let $S_k$ be a kernel score based on a p.s.d. kernel $k$ and $\hat{P}$ a predicted distribution for a target $Y \sim Q$, then
\begin{equation}
    \!\!\! \underbrace{\mathbb{E} \! \left[ S_k \! \left( \hat{P}\!, Y \right) \right]}_{\text{Generalization Error}} \!\! = \underbrace{- \! \left\lVert Q \right\rVert_k^2}_{\text{Noise}} + \underbrace{\left\lVert \mathbb{E} \! \left[ \hat{P} \right] \! - \! Q \right\rVert_k^2}_{\text{Bias}} \! + \underbrace{\operatorname{Var}_k \! \left( \hat{P} \right)}_{\text{Variance}}\!.
\end{equation}
If we have an ensemble prediction $\hat{P}^{\left( n \right)} \coloneqq \frac{1}{n} \sum_{i=1}^n \hat{P}_i$ with identically distributed members $\hat{P}_1, \dots, \hat{P}_n$, then
\begin{equation}
    \operatorname{Var}_k \! \left( \hat{P}^{\left( n \right)} \right) \! = \frac{1}{n} \operatorname{Var}_k \! \left( \hat{P}_1 \right) + \frac{n-1}{n} \operatorname{Cov}_k \! \left( \hat{P}_1, \hat{P}_2 \right).
\end{equation}
\label{th:bvcd}
\end{theorem}
In the same sense, a bias-variance-covariance decomposition of the expected MMD is implied since $\mathbb{E} [ \operatorname{MMD}^2_k (\hat{P}, Q ) ] = \mathbb{E} [ S_k (\hat{P}, Y ) ] + \lVert Q \rVert_k^2$.
Theorem \ref{th:bvcd} allows bias-variance and ensemble evaluations of increasingly more common generative models since kernels can be used for almost all data scenarios via vector embeddings \citep{liu2020learning}.
For example, in Section \ref{sec:applications}, we evaluate diffusion models for images, flow-based models for text-to-speech synthesis, and transformers for natural language generation.
It is also possible to offer Theorem \ref{th:bvcd} in terms of an associated reproducing kernel Hilbert space or if the ensemble members are not identically distributed (c.f. Appendix \ref{app:ext_theoretical_results}). \\
To illustrate Theorem \ref{th:bvcd}, we give as example the Brier score used in classification, which also recovers the original decomposition of \citet{ueda1996generalization}.
\begin{example}
    Let $\mathcal{P} = \Delta^l$ be the $l$-dimensional simplex and define $k_\delta \left( x, y \right) \coloneqq \mathbf{1}_{x=y} = 1$ if $x=y$ and $0$ otherwise. Then, for $y \in \left\{1, \dots, l \right\}$ and $P \in \mathcal{P}$, the negative kernel entropy $\left\lVert P \right\rVert^2_{k_\delta} = \sum_{i=1}^l P_i^2 =: \left\lVert P \right\rVert^2$ is the squared euclidean norm and
    $S_{k_\delta} \left( P, y \right) + 1 = \sum_{i=1}^l \left( P_i - \mathbf{1}_{i=y} \right)^2 =: \operatorname{BS} \left( P, y \right)$ is the Brier score \citep{VERIFICATIONOFFORECASTSEXPRESSEDINTERMSOFPROBABILITY}.
    Applying Theorem \ref{th:bvcd} decomposes $\mathbb{E} \left[\operatorname{BS} \left(\hat{P}, Y \right) \right]$ into
    \begin{equation}
        1 - \left\lVert Q \right\rVert^2 + \left\lVert \mathbb{E} \left[ \hat{P} \right] - Q \right\rVert^2 + \sum_{i=1}^l \mathbb{V} \left( \hat{P}_i \right).
    \end{equation}
\label{ex:brier}
\end{example}
The original decomposition in Equation \eqref{eq:sq_bvcd} can be recovered for binary classification since $\operatorname{BS} \left( P, y \right) = 2\left( P_1 - \mathbf{1}_{1=y} \right)^2$ for $l=2$. \\
In the last example, we omitted the covariance term for simplicity.
However, this term has also its practical utility as demonstrated in the next example.
\begin{example}
    Assume we train a generative model and intend to improve the generalization performance via an ensemble approach. Due to computational reasons, we decide to use an ensemble of consecutive training epochs as ensemble members $\hat{P}_1, \dots, \hat{P}_n$.
    We empirically perform such an experiment in Section \ref{sec:applications_image}, which shows that the assumptions of Theorem \ref{th:bvcd} hold approximately after excluding the first 20 epochs (c.f. Figure \ref{fig:metrics_per_epoch} and Equation \eqref{eq:cov_numbers}). We receive as estimates $\operatorname{Var}_k \left( \hat{P}_1 \right) \approx 0.0052$ and $\operatorname{Cov}_k \left( \hat{P}_1, \hat{P}_2 \right) \approx 0.0049$.
    Thus, 
    \begin{equation}
    \begin{split}
        & \overbrace{\mathbb{E} \left[ S_k \left( \hat{P}_1, Y \right) \right]}^{\text{Single Model Error}} - \overbrace{\mathbb{E} \left[ S_k \left( \frac{1}{n} \sum_{i=1} \hat{P}_i, Y \right) \right]}^{\text{Ensemble Error}} \\
        & = \frac{n-1}{n} \left( \operatorname{Var}_k \left( \hat{P}_1 \right) -\operatorname{Cov}_k \left( \hat{P}_1, \hat{P}_2 \right) \right) \\
        & \approx \left(1 - \frac{1}{n} \right) 0.0003.
    \end{split}
    \end{equation}
\label{ex:cov_numeric}
\end{example}
This example demonstrates how to predict the generalization improvement of arbitrary ensemble sizes even when the ensemble members are not independently distributed.
This is not possible without a variance-covariance decomposition. \\
In the following of this work, we use Theorem \ref{th:bvcd} to study the generalization behavior of generative models and to find ways to estimate the uncertainty of generated data.
The presented evaluations and approaches are applicable to almost any data generation task due to the flexibility of kernels and data embeddings.

\paragraph{Predictive Kernel Entropy for Single Models.}
Historically, the bias-variance decomposition had a large impact on the development of some of the most established machine learning algorithms, like Random Forests \citep{breiman2001random} or Gradient Boosting \citep{friedman2002stochastic}.
However, ensemble approaches are not similarly dominant for generative modeling.
Estimating $\operatorname{Var}_k ( \hat{P} )$ requires an ensemble of models, which is not always feasible.
Instead, note the decomposition 
$\operatorname{Var}_k ( \hat{P} ) = \mathbb{E} [ \lVert \hat{P} \rVert_k^2 ] - \lVert \mathbb{E} [ \hat{P} ] \rVert_k^2$
and observe that the distributional variance depends on the \textbf{predictive kernel entropy} $- \lVert \hat{P} \rVert_k^2$, which is estimated for single models.
The predictive kernel entropy also appears in the definition of kernel scores in Equation \eqref{eq:ks_def}.
This suggests that it may have a substantial influence on the generalization error.
In Section \ref{sec:applications}, we will discover that this influence is extremely high (Pearson correlation of approx. 0.95), but the sign of the correlation is task-specific.
Further, by using text embeddings, predictive kernel entropy is better than other baselines in predicting the performance of LLMs (c.f. Section \ref{sec:applications}).

\section{Unbiased and Consistent Estimators}
\label{sec:estimators}

If the prediction $\hat{P}$ is available in closed-form, the quantities in Theorem \ref{th:bvcd} can be computed according to conventional approaches \citep{gruber2023uncertainty}.
However, this is usually not the case for generative models in Deep Learning.
For example, Diffusion Models \citep{ho2020denoising} or closed-source LLMs \citep{openai2023gpt4} are also learning the training distribution, but they are often limited to generating samples.
In this section, we introduce estimators of the distributional variance and covariance for the case when only samples of the distributions are available.
This increases the practical applicability of Theorem \ref{th:bvcd} by a wide margin and allows investigating the most recent and largest generative models without constraints.
We assume a minimum of two samples from each distribution is given.
All estimators in the following require a two-stage sampling procedure \citep{sarndal2003model}: First, distributions are sampled in an outer loop, which can be seen as clusters.
In Section \ref{sec:applications}, this will be an ensemble of generative models.
Second, we sample of each distribution multiple times in an inner loop, which can be seen as within-cluster samples.
This will be the data generations of each model. \\
The procedure differs slightly between the variance and covariance case.
For simplicity, we also assume that all within-cluster sample sizes are the same.
All estimators can be adjusted if that is not the case and will still be unbiased and consistent.
Again, all proofs are presented in Appendix \ref{app:proofs}.

\subsection{Distributional Variance}

Assume we have a random variable $P$ with outcomes in $\mathcal{P}$ based on an unknown distribution $\mathbb{P}_P$ from which we can sample.
First, we sample distributions $P_1, \dots, P_n \overset{\text{iid}}{\sim} \mathbb{P}_P$.
Then, we sample $X_{i1}, \dots, X_{im} \overset{\text{iid}}{\sim} P_i$ for $i=1\dots n$.
The estimator we are about to propose is directly derived from the conventional variance estimator $\hat{\sigma}^2 \coloneqq \frac{1}{n-1} \sum_{i=1}^n \left\lVert P_i - \frac{1}{n} \sum_{s=1}^n P_s \right\rVert_k^2$.
Note that it holds $\hat{\sigma}^2 = \frac{1}{n} \sum_{i=1}^n \left\lVert P_i \right\rVert_k^2 - \frac{1}{n \left(n-1\right)} \sum_{i=1}^n \sum_{\substack{s=1 \\ s \neq i}}^n \Braket{ P_i | k | P_s}$, i.e. the estimator is the average of same-index pairs minus the average of the rest.
Our extended estimator then uses as plug-ins $\left\lVert P_i \right\rVert_k^2 \approx \frac{1}{m \left(m-1\right)} \sum_{j=1}^m \sum_{\substack{t=1 \\ t \neq j}}^m k \left( X_{ij}, X_{it} \right)$ and $\Braket{ P_i | k | P_s} \approx \frac{1}{m^2} \sum_{j=1}^m \sum_{\substack{t=1}}^m k \left( X_{ij}, X_{st} \right)$.
The complete estimator of the distributional variance $\operatorname{Var}_k \left( P \right)$ is defined by
\begin{equation}
\begin{split}
    \widehat{\operatorname{Var}}_k^{\left(n, m \right)} = & \underbrace{\frac{1}{n m \left(m-1\right)} \sum_{i=1}^n \sum_{\substack{j,t=1 \\ t \neq j}}^m k \left( X_{ij}, X_{it} \right)}_{\text{Average similarity within clusters}} \\
    & - \underbrace{\frac{1}{n \left(n \! - \! 1\right) m^2} \sum_{\substack{s,i=1 \\ s \neq i}}^n \sum_{j,t=1}^m k \left( X_{ij}, X_{st} \right)}_{\text{Average similarity between clusters}}.
\label{eq:var_est}
\end{split}
\end{equation}

\begin{figure*}
\centering
    \begin{subfigure}{.35\textwidth}
    \centering
    \resizebox{\columnwidth}{!}{%
    \begin{tikzpicture}
    \begin{axis}[
    ticks=none,
      axis x line=center,
      axis y line=center,
      xmin=-0.3,
      xmax=3.3,
      ymin=-0.3,
      ymax=1.7]
    \draw[red, line width=0.6mm] (0.5,0.3) -- (0.5,1.2);
    \draw[red, line width=0.6mm] (0.5,0.3) -- (0.9,0.7);
    \draw[red, line width=0.6mm] (0.5,1.2) -- (0.9,0.7);
    \draw[red, line width=0.6mm] (2.2,0.2) -- (2.5,1.2);
    \draw[red, line width=0.6mm] (1.8,0.7) -- (2.5,1.2);
    \draw[red, line width=0.6mm] (2.2,0.2) -- (1.8,0.7);

    \draw[cyan, line width=0.6mm, densely dotted] (0.5,0.3) -- (2.2,0.2);
    \draw[cyan, line width=0.6mm, densely dotted] (0.5,0.3) -- (1.8,0.7);
    \draw[cyan, line width=0.6mm, densely dotted] (0.5,0.3) -- (2.5,1.2);
    \draw[cyan, line width=0.6mm, densely dotted] (0.5,1.2) -- (2.2,0.2);
    \draw[cyan, line width=0.6mm, densely dotted] (0.5,1.2) -- (1.8,0.7);
    \draw[cyan, line width=0.6mm, densely dotted] (0.5,1.2) -- (2.5,1.2);
    \draw[cyan, line width=0.6mm, densely dotted] (0.9,0.7) -- (2.2,0.2);
    \draw[cyan, line width=0.6mm, densely dotted] (0.9,0.7) -- (1.8,0.7);
    \draw[cyan, line width=0.6mm, densely dotted] (0.9,0.7) -- (2.5,1.2);

    \filldraw[black] (0.5,0.3) circle (3.5pt) node[anchor=west] {};
    \filldraw[black] (0.5,1.2) circle (3.5pt) node[anchor=west] {};
    \filldraw[black] (0.9,0.7) circle (3.5pt) node[anchor=west] {};

    \filldraw[black] (2.2,0.2) circle (3.5pt) node[anchor=west] {};
    \filldraw[black] (2.5,1.2) circle (3.5pt) node[anchor=west] {};
    \filldraw[black] (1.8,0.7) circle (3.5pt) node[anchor=west] {};
    \node[] at (0.3,0.2) {\large $X_{13}$};
    \node[] at (0.67,0.7) {\large $X_{12}$};
    \node[] at (0.27,1.2) {\large $X_{11}$};
    \node[] at (2.44,0.15) {\large $X_{23}$};
    \node[] at (2.05,0.7) {\large $X_{22}$};
    \node[] at (2.75,1.2) {\large $X_{21}$};
    \node[] at (2.7,1.61) {\Large $\mathcal{X}$};
\end{axis}
    \end{tikzpicture}
    }
    \end{subfigure}%
    \begin{subfigure}{.4\textwidth}
    \centering
    \includegraphics[width=\columnwidth]{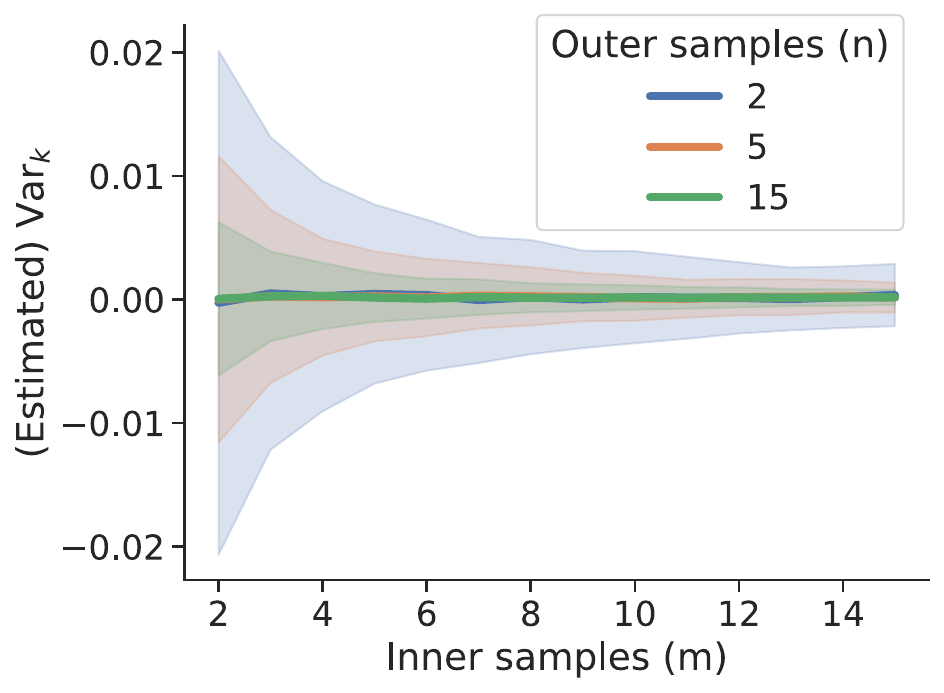}
    \end{subfigure}
\caption{
    \textbf{Left:} Illustration of the estimator $\widehat{\operatorname{Var}}_k^{\left(n,m\right)}$ in the sample space $\mathcal{X}$ for $n=2$ outer samples and $m=3$ inner samples. The estimator computes the average similarity within clusters (solid red lines) minus the average similarity between clusters (dotted blue lines). Shorter lines indicate higher similarity and larger kernel values.
    \textbf{Right:} Estimator standard deviation for various sample sizes. Even though the estimator does not converge in theory with the inner sample size $m$, it may still be influenced significantly by it for small sample sizes.
}
\label{fig:var_sim}
\end{figure*}

An illustration is given on the left in Figure \ref{fig:var_sim}.
The estimator is unbiased since $\mathbb{E} [ \widehat{\operatorname{Var}}_k^{\left(n, m \right)} ] = \operatorname{Var}_k \left( P \right)$.
Its runtime complexity is in $\mathcal{O} \left( m^2 n^2 \right)$. Estimators with lower complexity, like $\mathcal{O} \left( mn \right)$, exist but are not recommendable since they have a  worse performance and in most applications, generating the samples is far more costly than evaluating the estimator.
The variance of the estimator is in $\mathcal{O} \left( \frac{1}{n} \left( 1 + \frac{1}{m} \right) \right)$, which proves 
$\widehat{\operatorname{Var}}_k^{\left(n, m \right)} \longrightarrow \operatorname{Var}_k \left( P \right)$ in probability with growing $n$ but not $m$.
In words, the estimator is consistent with increasing outer samples but not inner samples.
This may suggest to neglect creating inner samples and keep $m$ small, but our analysis in Appendix \ref{app:var_estimator} shows that there exist sub-terms which converge equally fast in $m$ as in $n$.
In combination with the finite sample simulation in Figure \ref{fig:var_sim}, we recommend to use $m \geq n \geq 10$, if no prior information is available.

\subsection{Distributional Covariance and Correlation}

For the covariance case, assume we have random variables $P$ and $Q$ with outcomes in $\mathcal{P}$ based on an unknown joint distribution $\mathbb{P}_{PQ}$ from which we can sample.
We require samples $X_{i1}, \dots, X_{im} \overset{\text{iid}}{\sim} P_i$ and $Y_{i1}, \dots, Y_{im} \overset{\text{iid}}{\sim} Q_i$ with $\left(P_1, Q_1 \right), \dots, \left(P_n, Q_n \right) \overset{\text{iid}}{\sim} \mathbb{P}_{PQ}$.
Then, we propose the unbiased and consistent covariance estimator
\begin{equation}
\begin{split}
    \widehat{\operatorname{Cov}}_k^{\left(n, m \right)} \!\! \left( \mathbf{X}, \mathbf{Y} \right) & = \frac{1}{n m^2} \sum_{i=1}^n \sum_{j,t=1}^m k \left( X_{ij}, Y_{it} \right) \\
    & \!\!\!\!\! - \frac{1}{\left(n \! - \! 1\right)n m^2} \! \sum_{\substack{i,s=1 \\ s \neq i}}^n \sum_{j,t=1}^m \! k \left( X_{ij}, Y_{st} \right)
\label{eq:cov_est}    
\end{split}
\end{equation}
with $\mathbf{X} \coloneqq \left( X_{ij} \right)_{i=1\dots n, j=1 \dots m}$ and $\mathbf{Y} \coloneqq \left( Y_{ij} \right)_{i=1\dots n, j=1 \dots m}$.
It has the same runtime complexity and convergence rate as the variance estimator of Equation \eqref{eq:var_est} (c.f. Appendix \ref{app:cov_estimator}).
While the distributional covariance is directly implied by Theorem \ref{th:bvcd}, it is difficult to interpret since it is not bounded.
Consequently, we propose the distributional correlation estimator based on Equation \eqref{eq:cov_est} given by
\begin{equation}
    \widehat{\operatorname{Corr}}_k^{\left(n,m\right)} \!\! = \! \frac{\widehat{\operatorname{Cov}}_k^{\left(n, m \right)} \! \left( \mathbf{X}, \mathbf{Y} \right)}{\sqrt{\widehat{\operatorname{Cov}}_k^{\left(n, m \right)} \! \left( \mathbf{X}, \mathbf{X} \right) \widehat{\operatorname{Cov}}_k^{\left(n, m \right)} \! \left( \mathbf{Y}, \mathbf{Y} \right)}},
\end{equation}
which is in $\left[-1, 1 \right]$.
It is consistent since continuous transformations of consistent estimators are also consistent \citep{shao2003mathematical}, i.e. for $n \to \infty$ in probability
\begin{equation}
    \!\!\! \widehat{\operatorname{Corr}}_k^{\left(n,m\right)} \!\! \longrightarrow \operatorname{Corr}_k \! \left(P, Q \right) \! \coloneqq \! \frac{\operatorname{Cov}_k \! \left(P, Q \right)}{\sqrt{\operatorname{Var}_k \! \left( P \right) \operatorname{Var}_k \! \left( Q \right)}}.
\end{equation}
We use the covariance estimator for the variance terms since the variance estimator can be negative and the correlation estimator is only asymptotically unbiased no matter the choice.
Note that Pearson correlation is recovered by using the kernel $k_\delta$ of Example \ref{ex:brier}.

In the next section, we show that distributional correlation, as implied by Theorem \ref{th:bvcd}, is a natural tool to gain new insights into the fitting process of generative models.
For example, the correlations between epochs indicate how stable the convergence during training is.

\begin{figure*}[t]
\vskip 0.2in
\centering
    \begin{subfigure}{.41\textwidth}
    \centering
    \includegraphics[width=\columnwidth]{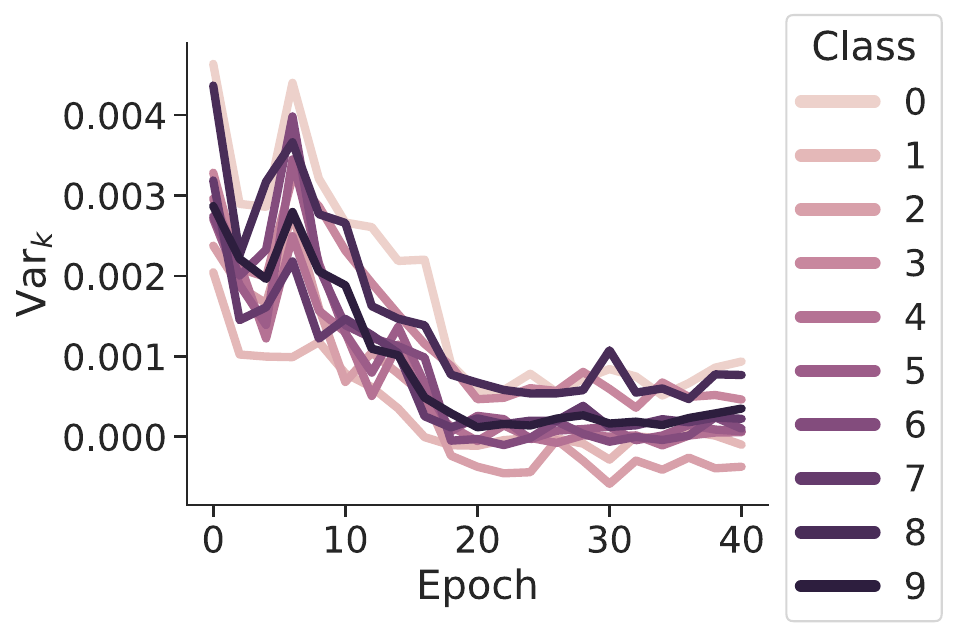}
    \end{subfigure}%
    \begin{subfigure}{.34\textwidth}
    \centering
    \includegraphics[width=\columnwidth]{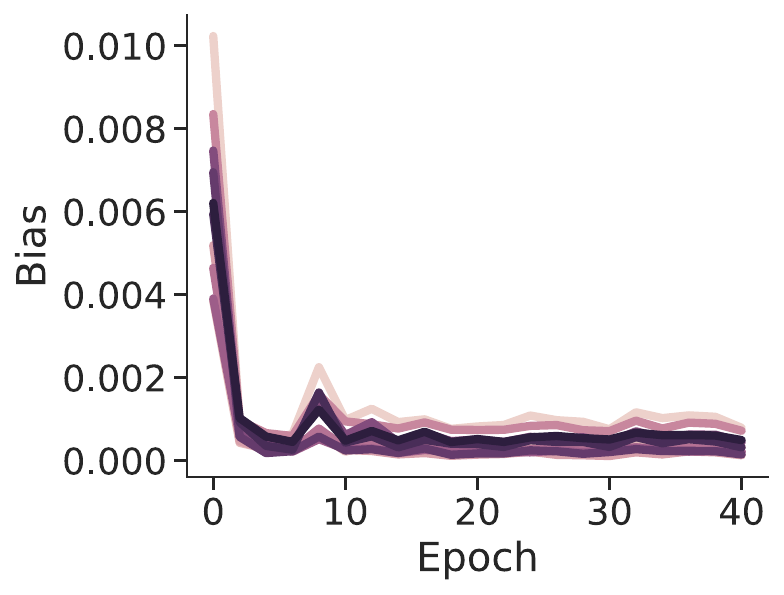}
    \end{subfigure}%
    \begin{subfigure}{.25\textwidth}
    \centering
    \includegraphics[width=\columnwidth]{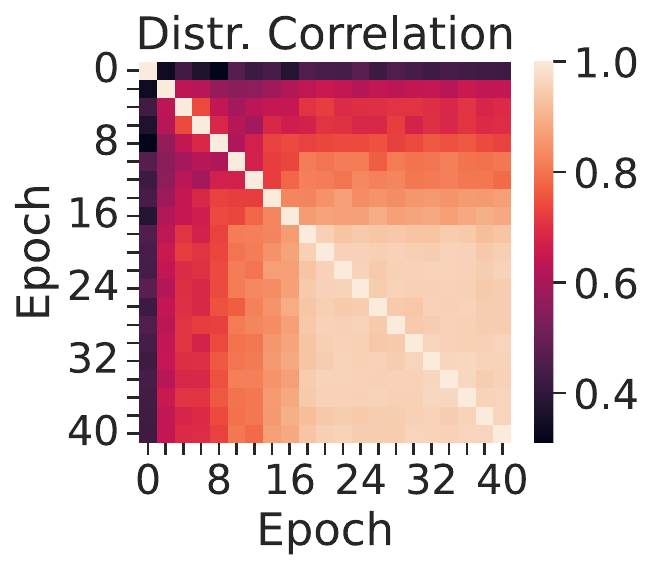}
    \end{subfigure}
\caption{
    \textbf{Left:} The variance starts high and is reduced throughout training. From 20 epochs onwards, the variance stays stable for all classes and no overfitting can be observed.
    \textbf{Mid: } The bias is reduced a lot quicker than the variance, reaching its minimum at 5 epochs, and converges after 10 epochs.
    \textbf{Right:} The distributional correlation between training epochs shows similar to the variance that convergence happens around epoch 20.
    Remarkably, the 'square' of very high correlations indicates that the model is stable in its convergence and does not iterate through equally good solutions.
}
\label{fig:metrics_per_epoch}
\vskip -0.2in
\end{figure*}

\begin{figure*}[t]
\vskip 0.2in
\centering
    \begin{subfigure}{.4\textwidth}
    \centering
    \includegraphics[width=\columnwidth]{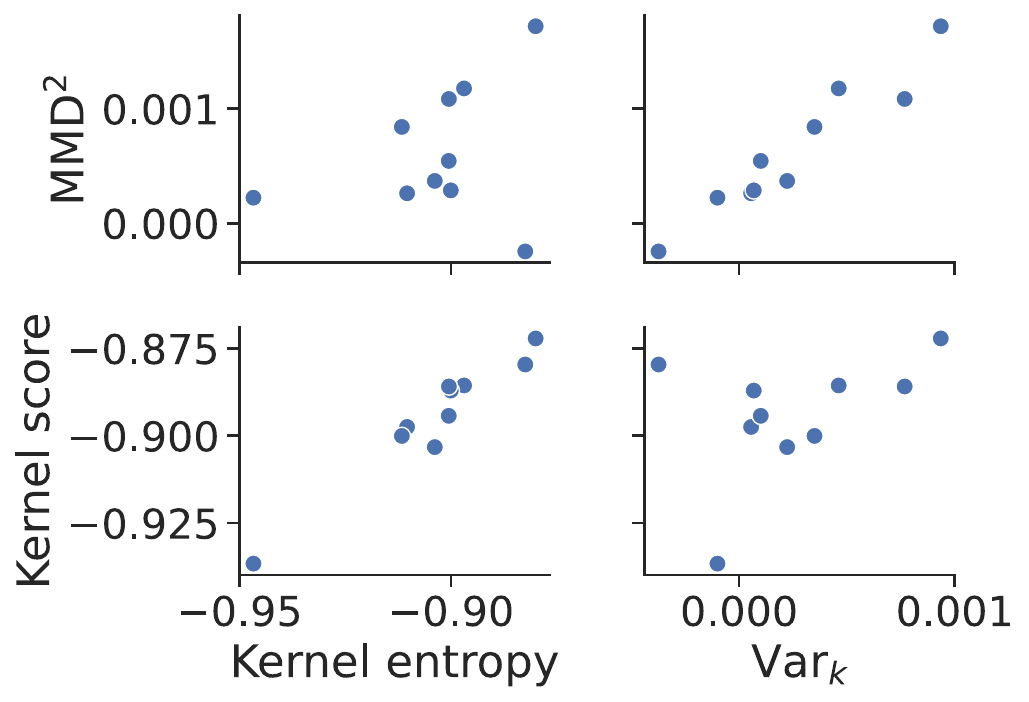}
    \end{subfigure}%
    \begin{subfigure}{.6\textwidth}
    \centering
    \includegraphics[width=\columnwidth]{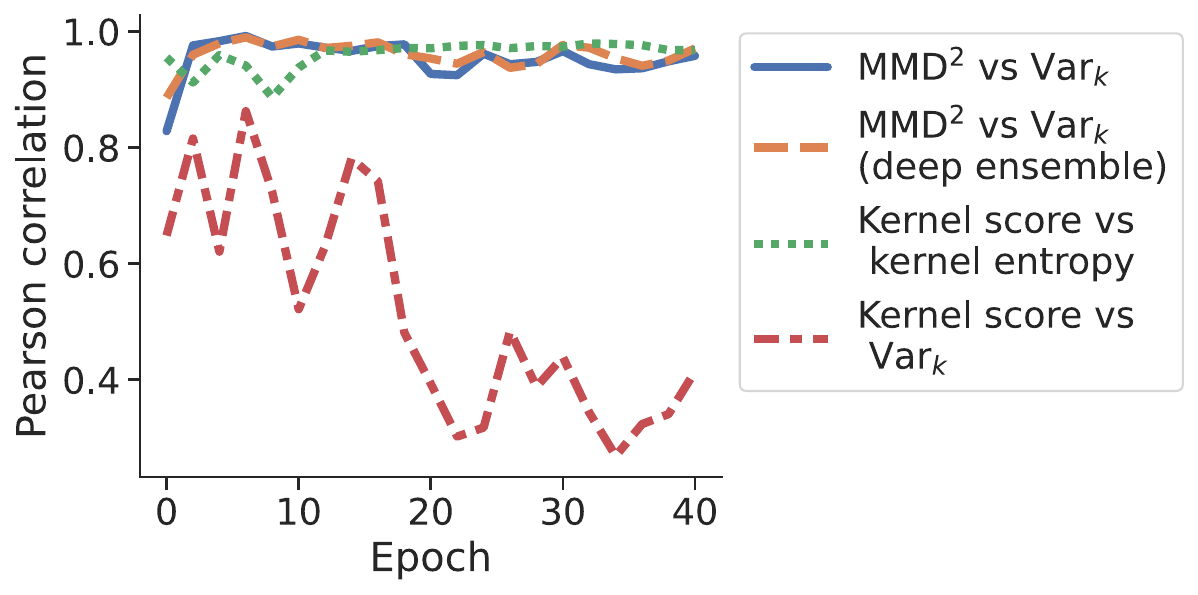}
    \end{subfigure}
\caption{
\textbf{Left:} Dependence between squared MMD and distributional variance.
The distributional variance correlates linearly with the squared MMD.
\textbf{Right:} Pearson correlation between squared MMD and distributional variance is very high ($\approx 0.95$) throughout training.
Approximation via deep ensembles does not deteriorate this relation.
Consequently, distributional variance and kernel entropy represent viable measures of uncertainty.
}
\label{fig:scatter_plots}
\vskip -0.2in
\end{figure*}

\section{Applications}
\label{sec:applications}

In this section, we apply the proposed statistical tools to assess generative models across a variety of different data generation tasks.
Specifically, we put emphasis on instance-level uncertainty estimation.
A meaningful measure of uncertainty is able to predict the loss for a given prediction.
We use the Pearson correlation coefficient to quantify how well an uncertainty measure predicts a continuous loss, and the area under receiver operator characteristic (AUROC) for a binary loss.
We first start in Section \ref{sec:applications_image} with diffusion models for image generation on the synthetic InfiMNIST dataset for a detailed examination of their generalization behavior.
In Section \ref{sec:applications_audio}, we evaluate an ensemble of Glow-TTS models for text-to-speech synthesis on the SpeechLJ dataset.
In all cases, kernel entropy shows strong performance as uncertainty measure.
Last, we use kernel entropy to outperform other baselines in uncertainty estimation for natural language generation in Section \ref{sec:applications_nlg}.
There, we evaluate single OPT models of different sizes on the question answering datasets CoQA and TriviaQA.
The source code of all experiments is openly available at \url{https://github.com/MLO-lab/BVCD_generative_models}.

\begin{figure}[t]
\vskip 0.2in
\centering
\includegraphics[width=0.34\textwidth]{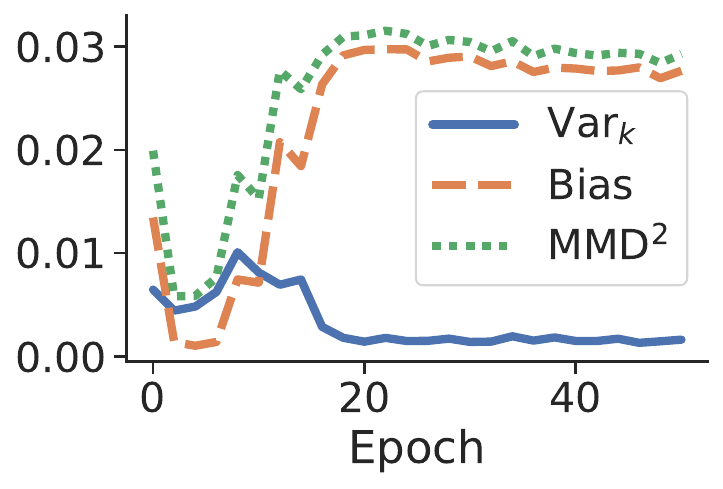}
\caption{
MMD$^2$, variance, and bias for class '0' throughout training with reduced training set of '0's.
After 5 epochs, mode collapse occurs, which is only expressed in the increased bias.
This indicates, that mode collapse is a contrary phenomenon to overfitting.
}
\label{fig:mode_collapse}
\vskip -0.2in
\end{figure}

\subsection{Image Generation}
\label{sec:applications_image}

For image generation, we use conditional diffusion models \citep{ho2020denoising, ho2021classifier} trained on MNIST-like datasets.
We use InfiMNIST to sample an infinite number of uniquely perturbed MNIST images \citep{loosli-canu-bottou-2006}.
By simulating the data generation process we can assess the ground truth generalization error of the diffusion model.
We sample $n=20$ distinct training sets from InfiMNIST each of size 60.000 (similar as MNIST).
We then train a model on each training set.
This is in correspondance to how generalization error, bias, and variance are evaluated for regression and classification tasks \citep{ueda1996generalization, gruber2023uncertainty}.
We use $m=20$ generated images per class and per model for all estimators.
The predictive kernel entropy is only evaluated on a single model to stay as closely as possible to practical constraints.
Our kernel choice is the commonly used RBF kernel $k_{\mathrm{rbf}} \left( x, y \right) = \exp ( - \gamma \left\lVert x - y \right\rVert^2_2 )$, where $x$ and $y$ are flattened images and $\gamma$ a normalization factor based on the number of pixels \citep{scholkopf1997support, scholkopf2002learning, liu2020learning}.
We first analyse the generalization behavior and then assess different approaches for uncertainty estimation.
In Figure \ref{fig:metrics_per_epoch}, we plot the distributional variance, bias, and distributional correlation throughout training for every second epoch.
As can be seen, the model converges quicker for the bias than the variance (around epochs 10 and 20).
Further, no overfitting occurs since both variance and bias stay small.
The correlation matrix also shows convergence around epoch 20, but, more interestingly, it shows a square of high correlations in the lower right corner.
This is an indication that the diffusion model training is stable in its convergence and does not iterate through different minima in the optimization landscape. \\
In Figure \ref{fig:scatter_plots}, we compare the relations between kernel score and MMD$^2$ to distributional variance and predictive kernel entropy for each class.
As can be seen, the predictive kernel entropy correlates strongly linearly with the generalization error (kernel score) but not so much with the generalization discrepancy (MMD$^2$), while the distributional variance correlates strongly linearly with the MMD$^2$ but not the kernel score.
This correlation can be observed throughout the whole training and gives a very high Pearson correlation coefficient of around $0.95$.
Importantly for practical settings, the correlation between MMD$^2$ and distributional variance does not deteriorate when we use a deep ensemble trained on a single dataset.\\
In summary, these results demonstrate that both distributional variance as well as predictive kernel entropy are viable measure of uncertainty to predict the correctness of generated instances, either in terms of kernel score or MMD$^2$.

We next set out to use our estimators for bias and variance to elucidate the phenomenon of mode collapse. 
When generative models are used to learn the distribution of a given training set, there are often groups of different sizes present.
In these cases, a common occurence is mode collapse towards the majority groups, i.e. the model catastrophically fails to model the minority groups.
To simulate this scenario in our setting, we repeat our evaluation but reduce the frequency of images of digit '0' to $\approx 1\%$ in each training set. 
The bias-variance curves of digit '0' across training (Figure \ref{fig:mode_collapse}) reveals the expected mode collapse, and, most importantly, demonstrates that it is only expressed in terms of the bias.
The variance term is further reduced throughout training as if no collapse occured.
This suggests that mode collapse may be seen as a contrary phenomenon to overfitting.
While in overfitting, the variance increases and the bias reduces, this is vice versa for mode collapse for prolonged training.
In Appendix \ref{app:experiments}, we conduct additional evaluations and confirm that our findings are robust with respect to common kernel choices.

\begin{figure*}[t]
\vskip 0.2in
\centering
    \begin{subfigure}{.38\textwidth}
    \centering
    \includegraphics[width=\columnwidth]{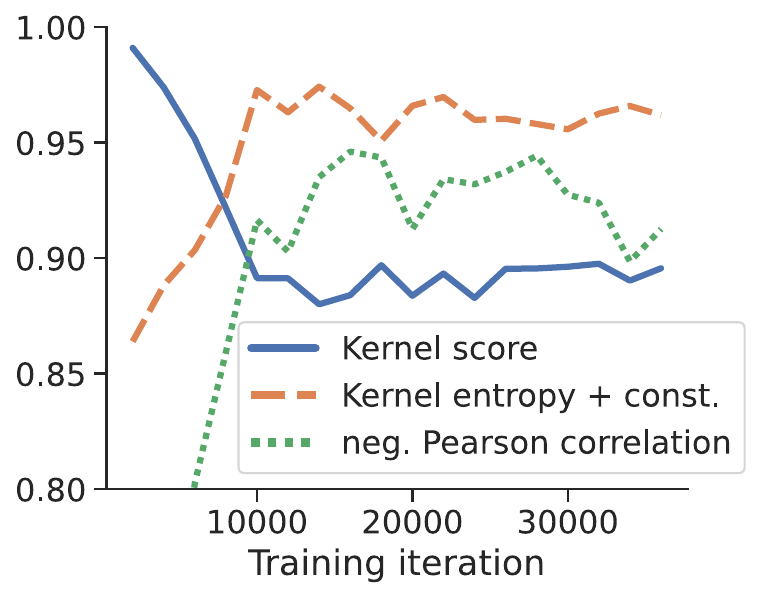}
    \end{subfigure}%
    \begin{subfigure}{.38\textwidth}
    \centering
    \includegraphics[width=\columnwidth]{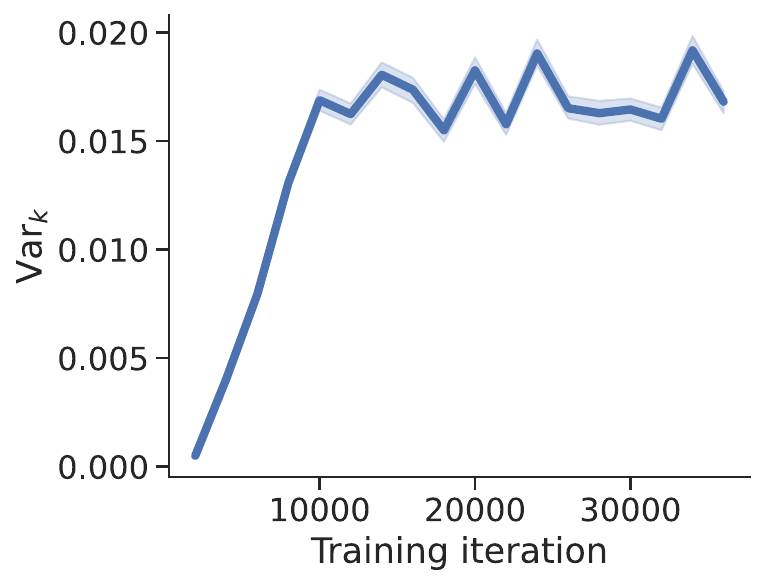}
    \end{subfigure}%
    \begin{subfigure}{.23\textwidth}
    \centering
    \includegraphics[width=\columnwidth]{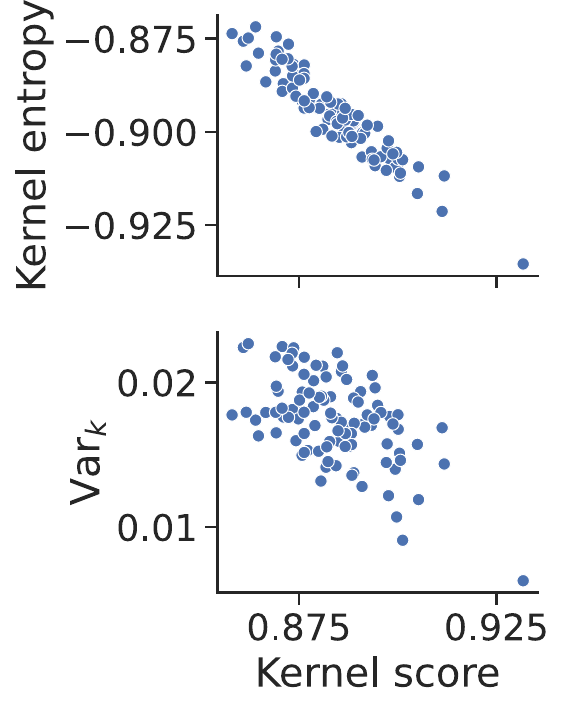}
    \end{subfigure}
\caption{
    \textbf{Left:} Kernel score, predictive kernel entropy, and their Pearson correlation throughout training for Glow-TTS.
    The entropy indicates that the model initially predicts too narrow distributions, which widen until convergence at around 10000 iterations.
    After convergence, the correlation between predictive entropy and kernel score is very high.
    \textbf{Mid:} The variance of an Deep Ensemble is also initially very small and converges at the same time as kernel score and entropy.
    \textbf{Right:} Comparing kernel score and kernel entropy as well as variance for 100 test instances at 16000 training iterations. Again, the correlation shows strong linearity for kernel entropy.
}
\label{fig:audio_per_epoch}
\vskip -0.2in
\end{figure*}

\begin{figure*}[t]
\vskip 0.2in
\centering
    \begin{subfigure}{.59\textwidth}
    \centering
    \includegraphics[width=\columnwidth]{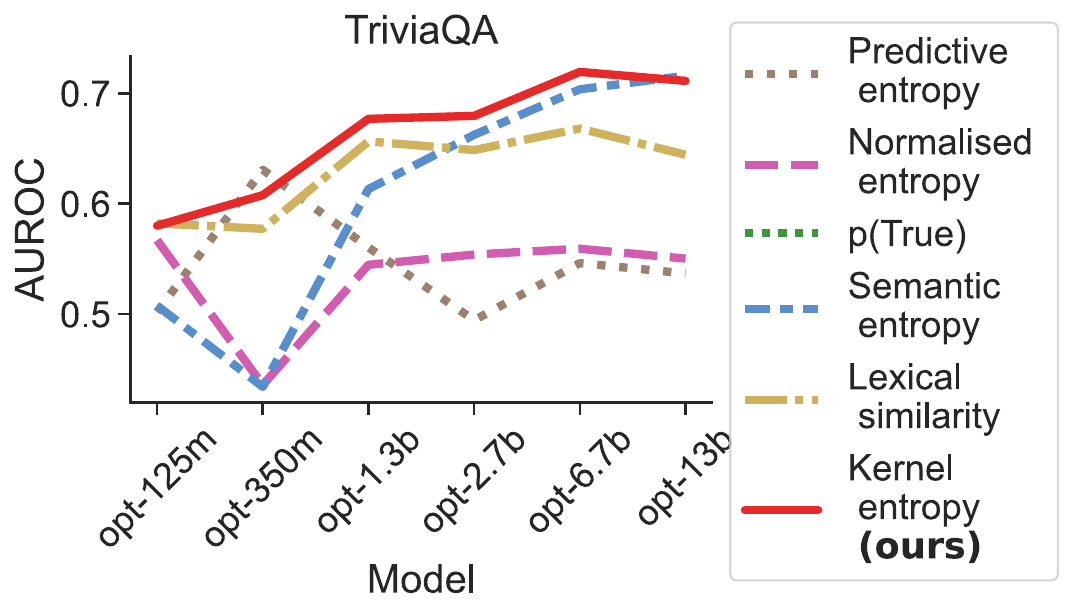}
    \end{subfigure}%
    \begin{subfigure}{.41\textwidth}
    \centering
    \includegraphics[width=\columnwidth]{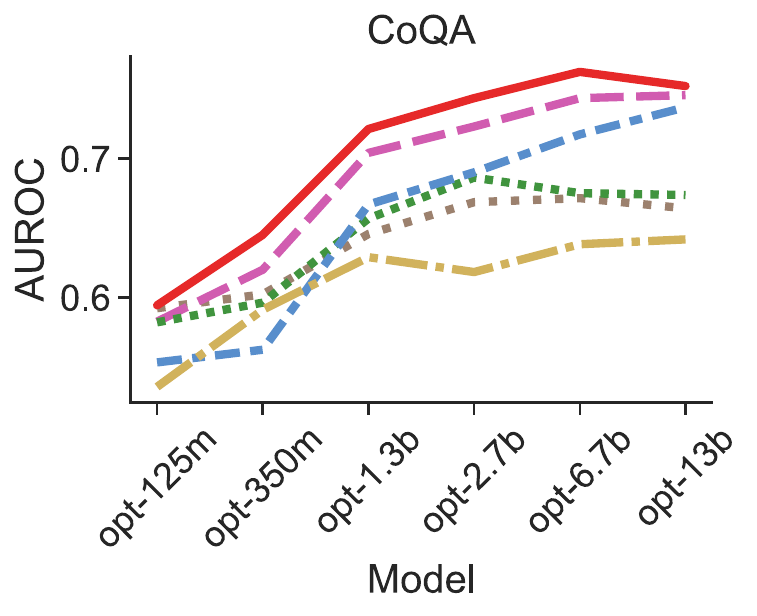}
    \end{subfigure}
\caption{
Area-Under-Curve of answer accuracy based on thresholds for different uncertainty measures.
The kernel entropy outperforms other baselines in predicting the correctness of generated answers across a wide range of differently sized models.
}
\label{fig:nlg_sota}
\vskip -0.2in
\end{figure*}

\subsection{Audio Generation}
\label{sec:applications_audio}

We next evaluate the fitting and generalization behavior of the generative flow model Glow-TTS \citep{kim2020glow} on the text-to-speech dataset LJSpeech \citep{ljspeech17} throughout training (Appendix \ref{app:experiments}).
We train a deep ensemble via $n=10$ different weight initializations on $90\%$ of the available data.
We evaluate the models every 2000 training iterations by generating $m=10$ speech waveforms for each of 100 test instances.
The evaluation includes the kernel score, (predictive) kernel entropy and the distributional variance. Note that ensemble members are only required to compute the distributional variance -- we compute kernel entropy for a single model, as before.
Here, we use the Laplacian kernel $k_{\mathrm{lap}} \left( x, y \right) = \exp ( - \gamma \left\lVert x - y \right\rVert_1 )$ \citep{scholkopf2002learning}, where $x$ and $y$ are waveforms represented by vectors and $\gamma$ a normalization constant based on the waveform length.
The results are depicted in Figure \ref{fig:audio_per_epoch}.
Our results reveal that similar as for image generation, overfitting does not occur for prolonged training and the Pearson correlation between kernel entropy and kernel score is very high after convergence.
But, the entropy is initially very small and increases until convergence.
This indicates that a successful training requires to widen the predicted distribution.
Consequently, the correlation between kernel entropy and kernel score is negative, since badly fitted instances have a more narrow predicted distribution.
We also conduct the evaluations with the RBF kernel, which gives similar but slightly more erratic curves than the Laplacian kernel in Appendix \ref{app:experiments}.

\subsection{Natural Language Generation}
\label{sec:applications_nlg}

An instance-level uncertainty measure is supposed to predict the correctness of an individual prediction and should therefore be highly correlated to the loss.
In all experiments so far, we observed a very high correlation between kernel entropy and kernel score.
This indicates that kernel entropy is an excellent measure of uncertainty.
In the following, we examine kernel entropy to predict the correctness of LLMs on question answering datasets.
Here, the setup differs from the previous experiments by two aspects.\\
First, we follow \citet{kuhn2022semantic} and do not use a kernel score but a binarized version of the $\operatorname{RougeL}$ similarity as loss.
For two sequences $s, t$, it is defined as $\operatorname{RougeL} \left(s, t \right) = \frac{2}{\operatorname{length} \left( s \right) + \operatorname{length} \left( t \right)} \operatorname{LCS} \left(s, t \right)$, where $\operatorname{LCS}$ is the length of the \textbf{l}ongest \textbf{c}ommon \textbf{s}equence between its two inputs \citep{lin2004automatic}.
\citet{kuhn2022semantic} propose to use the binary loss $L \left( answer, target \right) = \mathbf{1}_{\operatorname{RougeL} \left(answer, target \right) > 0.3}$.
Specifically, they claim that this particular loss matches human-based evaluation 89\% of the time for CoQA and 96\% of the time for TriviaQA.
This turns predicting the loss value into a binary classification problem.
Consequently, the AUROC is more meaningful than Pearson correlation for evaluating the performance of uncertainty measures \citep{kadavath2022language}. \\
Second, we do not directly use the generated answers as inputs for a kernel but, instead, their vector embeddings.
A well-trained vector embedder maps text into a semantically meaningful vector space in which a kernel then measures similarities \citep{camacho2018word}. \\
The investigated uncertainty baselines are predictive entropy, normalised (predictive) entropy, p(True), semantic entropy, and lexical similarity \citep{kuhn2022semantic}.
We consider uncertainty estimation for question answering predictions of the datasets CoQA \citep{reddy2019coqa} with 7983 test instances and TriviaQA \citep{joshi2017triviaqa} with 17943 test instances.
We use OPT models \citep{zhang2022opt} of all available sizes except the 30 billion parameter version, which is computationally prohibitive.
For our kernel entropy, we use the RBF kernel and text embeddings computed via a pretrained e5-small-v2 \citep{wang2022text}.
The results are depicted in Figure \ref{fig:nlg_sota}.
As can be seen, kernel entropy is the most robust approach and outperforms other baselines for uncertainty estimation in almost all cases.
We can achieve further improvements in our approach when we use alternative embedders (c.f. Appendix \ref{app:experiments}).
The cosine similarity, which is used in natural language processing \citep{steinbach2000comparison}, and other kernels show similar results as the RBF kernel in Appendix \ref{app:experiments}. \\
Our approach of combining an embedding model with a p.s.d. kernel outperforms all other uncertainty baselines.
However, a good embedding model might not always be accessible for every language setup.
In Appendix \ref{app:experiments}, we additionally evaluate kernel entropy based on a sequence kernel, which does not require an embedding model.
While the performance is worse compared to using an embedder, the sequence kernel is still competitive and clearly outperforms lexical similarity.

\section{Limitations}

Our work builds upon large bodies of literature, and, consequently, shares their limitations.
We give an overview of these in the following. \\
Our bias-variance decomposition assumes that the MMD and kernel score are meaningful measures of generalization performance.
The choice of kernel can be inspired by the large body of work on support vector machines \citep{scholkopf2002learning} and MMD-based hypothesis tests \citep{JMLR:v13:gretton12a}.
Further, we share similar limitations as other bias-variance decompositions in the literature \citep{ueda1996generalization, gruber2023uncertainty}.
Specifically, the intractability of simulating the exact training distribution is often limiting the bias-variance estimates in practice.
Deep ensembles can give reasonable approximations, which we use for the audio experiments. \\
Our language experiments are adopted from past literature on uncertainty estimation \citep{malinin2020uncertainty, kadavath2022language, kuhn2022semantic}.
Since this branch of research is mostly empirical, it is not possible to give guarantees of how each uncertainty baseline performs for new settings.
This limitation also holds for kernel entropy, and is underlined by its strong but negative correlations with the error for the audio experiments.
Recent theoretical work in the classification setting explored the link between uncertainty and accuracy via the class-aggregated calibration error (CACE) \citep{jiang2021assessing, kirsch2022note}.
However, this relationship still needs to be generalized because CACE is not defined for non-classification models. \\
Last, our estimators require i.i.d. samples to converge strongly with sample size \citep{capinski2004measure}.
This usually holds for samples from an individual model (i.e. when we estimate kernel entropy, kernel score, or MMD). However, it may not hold when estimating the variance, covariance, or bias of the model.

\section{Conclusion}

In this work we introduced the first bias-variance-covariance decomposition beyond the mean squared error for kernel scores.
We proposed estimators for the variance and covariance terms which only require samples of the predictive distributions.
This allows to evaluate all terms in the composition for arbitrary generative models and even in the closed-source setting.
We studied empirically the fitting behavior of common models for image and audio generations, and demonstrated that kernel entropy and variance are viable measures of uncertainty.
Finally, we showed that kernel entropy outperforms other baselines for predicting the correctness of LLMs in question answering tasks.

\section*{Impact Statement}

This paper presents work whose goal is to advance the field of machine learning theoretically and empirically.
There are many potential societal consequences of our work, none which we feel must be specifically highlighted here.
However, we emphasise that uncertainty estimation may improve safety and trustworthiness of generative models but without guarantees.
Practitioners should always verify any adaptations on an unseen test set and be wary of distribution shifts.

\section*{Acknowledgements}
Co-funded by the European Union (ERC, TAIPO, 101088594 to FB). Views and opinions expressed are however those of the authors only and do not necessarily reflect those of the European Union or the European Research Council. Neither the European Union nor the granting authority can be held responsible for them.

\bibliography{main}

\begin{thebibliography}{78}
\providecommand{\natexlab}[1]{#1}
\providecommand{\url}[1]{\texttt{#1}}
\expandafter\ifx\csname urlstyle\endcsname\relax
  \providecommand{\doi}[1]{doi: #1}\else
  \providecommand{\doi}{doi: \begingroup \urlstyle{rm}\Url}\fi

\bibitem[Baum et~al.(2023)Baum, Kanagawa, and Gretton]{baum2023kernel}
Baum, J., Kanagawa, H., and Gretton, A.
\newblock A kernel stein test of goodness of fit for sequential models.
\newblock In \emph{International Conference on Machine Learning}, pp.\  1936--1953. PMLR, 2023.

\bibitem[Bi{\'n}kowski et~al.(2018)Bi{\'n}kowski, Sutherland, Arbel, and Gretton]{binkowski2018demystifying}
Bi{\'n}kowski, M., Sutherland, D.~J., Arbel, M., and Gretton, A.
\newblock Demystifying {MMD} {GAN}s.
\newblock In \emph{International Conference on Learning Representations}, 2018.

\bibitem[Bishop \& Nasrabadi(2006)Bishop and Nasrabadi]{bishop2006pattern}
Bishop, C.~M. and Nasrabadi, N.~M.
\newblock \emph{Pattern recognition and machine learning}, volume~4.
\newblock Springer, 2006.

\bibitem[Breiman(2001)]{breiman2001random}
Breiman, L.
\newblock Random forests.
\newblock \emph{Machine learning}, 45\penalty0 (1):\penalty0 5--32, 2001.

\bibitem[Brier(1950)]{VERIFICATIONOFFORECASTSEXPRESSEDINTERMSOFPROBABILITY}
Brier, G.~W.
\newblock Verification of forecasts expressed in terms of probability.
\newblock \emph{Monthly Weather Review}, 78\penalty0 (1):\penalty0 1 -- 3, 1950.
\newblock \doi{10.1175/1520-0493(1950)078<0001:VOFEIT>2.0.CO;2}.
\newblock URL \url{https://journals.ametsoc.org/view/journals/mwre/78/1/1520-0493_1950_078_0001_vofeit_2_0_co_2.xml}.

\bibitem[Brown(2004)]{brown2004diversity}
Brown, G.
\newblock \emph{Diversity in neural network ensembles}.
\newblock PhD thesis, Citeseer, 2004.

\bibitem[Camacho-Collados \& Pilehvar(2018)Camacho-Collados and Pilehvar]{camacho2018word}
Camacho-Collados, J. and Pilehvar, M.~T.
\newblock From word to sense embeddings: A survey on vector representations of meaning.
\newblock \emph{Journal of Artificial Intelligence Research}, 63:\penalty0 743--788, 2018.

\bibitem[Capi{\'n}ski \& Kopp(2004)Capi{\'n}ski and Kopp]{capinski2004measure}
Capi{\'n}ski, M. and Kopp, P.~E.
\newblock \emph{Measure, integral and probability}, volume~14.
\newblock Springer, 2004.

\bibitem[Choi et~al.(2022)Choi, Lee, Shin, Kim, Kim, and Yoon]{choi2022perception}
Choi, J., Lee, J., Shin, C., Kim, S., Kim, H., and Yoon, S.
\newblock Perception prioritized training of diffusion models.
\newblock In \emph{Proceedings of the IEEE/CVF Conference on Computer Vision and Pattern Recognition}, pp.\  11472--11481, 2022.

\bibitem[Chwialkowski et~al.(2016)Chwialkowski, Strathmann, and Gretton]{chwialkowski2016kernel}
Chwialkowski, K., Strathmann, H., and Gretton, A.
\newblock A kernel test of goodness of fit.
\newblock In \emph{International conference on machine learning}, pp.\  2606--2615. PMLR, 2016.

\bibitem[Dawid(2007)]{dawid2007geometry}
Dawid, A.~P.
\newblock The geometry of proper scoring rules.
\newblock \emph{Annals of the Institute of Statistical Mathematics}, 59\penalty0 (1):\penalty0 77--93, 2007.

\bibitem[Dziugaite et~al.(2015)Dziugaite, Roy, and Ghahramani]{dziugaite2015training}
Dziugaite, G.~K., Roy, D.~M., and Ghahramani, Z.
\newblock Training generative neural networks via maximum mean discrepancy optimization.
\newblock In \emph{Proceedings of the Thirty-First Conference on Uncertainty in Artificial Intelligence}, pp.\  258--267, 2015.

\bibitem[Eaton(1981)]{eaton1981method}
Eaton, M.
\newblock A method for evaluating improper prior distributions.
\newblock Technical report, University of Minnesota, 1981.

\bibitem[Eaton et~al.(1996)Eaton, Giovagnoli, and Sebastiani]{eaton1996predictive}
Eaton, M.~L., Giovagnoli, A., and Sebastiani, P.
\newblock A predictive approach to the bayesian design problem with application to normal regression models.
\newblock \emph{Biometrika}, 83\penalty0 (1):\penalty0 111--125, 1996.

\bibitem[Eren \& {The Coqui TTS Team}(2021)Eren and {The Coqui TTS Team}]{Eren_Coqui_TTS_2021}
Eren, G. and {The Coqui TTS Team}.
\newblock {Coqui TTS}, January 2021.
\newblock URL \url{https://github.com/coqui-ai/TTS}.

\bibitem[Fomicheva et~al.(2020)Fomicheva, Sun, Yankovskaya, Blain, Guzm{\'a}n, Fishel, Aletras, Chaudhary, and Specia]{fomicheva2020unsupervised}
Fomicheva, M., Sun, S., Yankovskaya, L., Blain, F., Guzm{\'a}n, F., Fishel, M., Aletras, N., Chaudhary, V., and Specia, L.
\newblock Unsupervised quality estimation for neural machine translation.
\newblock \emph{Transactions of the Association for Computational Linguistics}, 8:\penalty0 539--555, 2020.

\bibitem[Friedman(2002)]{friedman2002stochastic}
Friedman, J.~H.
\newblock Stochastic gradient boosting.
\newblock \emph{Computational statistics \& data analysis}, 38\penalty0 (4):\penalty0 367--378, 2002.

\bibitem[Gneiting \& Raftery(2007)Gneiting and Raftery]{gneitingscores}
Gneiting, T. and Raftery, A.~E.
\newblock Strictly proper scoring rules, prediction, and estimation.
\newblock \emph{Journal of the American Statistical Association}, 102\penalty0 (477):\penalty0 359--378, 2007.
\newblock \doi{10.1198/016214506000001437}.
\newblock URL \url{https://doi.org/10.1198/016214506000001437}.

\bibitem[Gretton et~al.(2003)Gretton, Herbrich, and Smola]{gretton2003kernel}
Gretton, A., Herbrich, R., and Smola, A.~J.
\newblock The kernel mutual information.
\newblock In \emph{2003 IEEE International Conference on Acoustics, Speech, and Signal Processing, 2003. Proceedings.(ICASSP'03).}, volume~4, pp.\  IV--880. IEEE, 2003.

\bibitem[Gretton et~al.(2012{\natexlab{a}})Gretton, Borgwardt, Rasch, Sch{{\"o}}lkopf, and Smola]{JMLR:v13:gretton12a}
Gretton, A., Borgwardt, K.~M., Rasch, M.~J., Sch{{\"o}}lkopf, B., and Smola, A.
\newblock A kernel two-sample test.
\newblock \emph{Journal of Machine Learning Research}, 13\penalty0 (25):\penalty0 723--773, 2012{\natexlab{a}}.
\newblock URL \url{http://jmlr.org/papers/v13/gretton12a.html}.

\bibitem[Gretton et~al.(2012{\natexlab{b}})Gretton, Sejdinovic, Strathmann, Balakrishnan, Pontil, Fukumizu, and Sriperumbudur]{gretton2012optimal}
Gretton, A., Sejdinovic, D., Strathmann, H., Balakrishnan, S., Pontil, M., Fukumizu, K., and Sriperumbudur, B.~K.
\newblock Optimal kernel choice for large-scale two-sample tests.
\newblock \emph{Advances in neural information processing systems}, 25, 2012{\natexlab{b}}.

\bibitem[Gruber \& Buettner(2023)Gruber and Buettner]{gruber2023uncertainty}
Gruber, S.~G. and Buettner, F.
\newblock Uncertainty estimates of predictions via a general bias-variance decomposition.
\newblock In \emph{International Conference on Artificial Intelligence and Statistics}, pp.\  11331--11354. PMLR, 2023.

\bibitem[Hastie et~al.(2009)Hastie, Tibshirani, Friedman, and Friedman]{hastie2009elements}
Hastie, T., Tibshirani, R., Friedman, J.~H., and Friedman, J.~H.
\newblock \emph{The elements of statistical learning: data mining, inference, and prediction}, volume~2.
\newblock Springer, 2009.

\bibitem[Hekler et~al.(2023)Hekler, Brinker, and Buettner]{hekler2023test}
Hekler, A., Brinker, T.~J., and Buettner, F.
\newblock Test time augmentation meets post-hoc calibration: uncertainty quantification under real-world conditions.
\newblock In \emph{Proceedings of the AAAI Conference on Artificial Intelligence}, volume~37, pp.\  14856--14864, 2023.

\bibitem[Heusel et~al.(2017)Heusel, Ramsauer, Unterthiner, Nessler, and Hochreiter]{heusel2017gans}
Heusel, M., Ramsauer, H., Unterthiner, T., Nessler, B., and Hochreiter, S.
\newblock Gans trained by a two time-scale update rule converge to a local nash equilibrium.
\newblock \emph{Advances in neural information processing systems}, 30, 2017.

\bibitem[Ho \& Salimans(2021)Ho and Salimans]{ho2021classifier}
Ho, J. and Salimans, T.
\newblock Classifier-free diffusion guidance.
\newblock In \emph{NeurIPS 2021 Workshop on Deep Generative Models and Downstream Applications}, 2021.

\bibitem[Ho et~al.(2020)Ho, Jain, and Abbeel]{ho2020denoising}
Ho, J., Jain, A., and Abbeel, P.
\newblock Denoising diffusion probabilistic models.
\newblock \emph{Advances in neural information processing systems}, 33:\penalty0 6840--6851, 2020.

\bibitem[Ito \& Johnson(2017)Ito and Johnson]{ljspeech17}
Ito, K. and Johnson, L.
\newblock The lj speech dataset.
\newblock \url{https://keithito.com/LJ-Speech-Dataset/}, 2017.

\bibitem[Jiang et~al.(2021)Jiang, Nagarajan, Baek, and Kolter]{jiang2021assessing}
Jiang, Y., Nagarajan, V., Baek, C., and Kolter, J.~Z.
\newblock Assessing generalization of sgd via disagreement.
\newblock In \emph{International Conference on Learning Representations}, 2021.

\bibitem[Joshi et~al.(2017)Joshi, Choi, Weld, and Zettlemoyer]{joshi2017triviaqa}
Joshi, M., Choi, E., Weld, D.~S., and Zettlemoyer, L.
\newblock Triviaqa: A large scale distantly supervised challenge dataset for reading comprehension.
\newblock In \emph{Proceedings of the 55th Annual Meeting of the Association for Computational Linguistics (Volume 1: Long Papers)}, pp.\  1601--1611, 2017.

\bibitem[Kadavath et~al.(2022)Kadavath, Conerly, Askell, Henighan, Drain, Perez, Schiefer, Hatfield-Dodds, DasSarma, Tran-Johnson, et~al.]{kadavath2022language}
Kadavath, S., Conerly, T., Askell, A., Henighan, T., Drain, D., Perez, E., Schiefer, N., Hatfield-Dodds, Z., DasSarma, N., Tran-Johnson, E., et~al.
\newblock Language models (mostly) know what they know.
\newblock \emph{arXiv preprint arXiv:2207.05221}, 2022.

\bibitem[Karras et~al.(2020)Karras, Aittala, Hellsten, Laine, Lehtinen, and Aila]{karras2020training}
Karras, T., Aittala, M., Hellsten, J., Laine, S., Lehtinen, J., and Aila, T.
\newblock Training generative adversarial networks with limited data.
\newblock \emph{Advances in neural information processing systems}, 33:\penalty0 12104--12114, 2020.

\bibitem[Kasneci et~al.(2023)Kasneci, Se{\ss}ler, K{\"u}chemann, Bannert, Dementieva, Fischer, Gasser, Groh, G{\"u}nnemann, H{\"u}llermeier, et~al.]{kasneci2023chatgpt}
Kasneci, E., Se{\ss}ler, K., K{\"u}chemann, S., Bannert, M., Dementieva, D., Fischer, F., Gasser, U., Groh, G., G{\"u}nnemann, S., H{\"u}llermeier, E., et~al.
\newblock Chatgpt for good? on opportunities and challenges of large language models for education.
\newblock \emph{Learning and individual differences}, 103:\penalty0 102274, 2023.

\bibitem[Kim et~al.(2020)Kim, Kim, Kong, and Yoon]{kim2020glow}
Kim, J., Kim, S., Kong, J., and Yoon, S.
\newblock Glow-tts: A generative flow for text-to-speech via monotonic alignment search.
\newblock \emph{Advances in Neural Information Processing Systems}, 33:\penalty0 8067--8077, 2020.

\bibitem[Kir{\'a}ly \& Oberhauser(2019)Kir{\'a}ly and Oberhauser]{kiraly2019kernels}
Kir{\'a}ly, F.~J. and Oberhauser, H.
\newblock Kernels for sequentially ordered data.
\newblock \emph{Journal of Machine Learning Research}, 20\penalty0 (31):\penalty0 1--45, 2019.

\bibitem[Kirsch \& Gal(2022)Kirsch and Gal]{kirsch2022note}
Kirsch, A. and Gal, Y.
\newblock A note on" assessing generalization of sgd via disagreement".
\newblock \emph{Transactions on Machine Learning Research}, 2022.

\bibitem[K{\"u}bler et~al.(2020)K{\"u}bler, Jitkrittum, Sch{\"o}lkopf, and Muandet]{kubler2020learning}
K{\"u}bler, J., Jitkrittum, W., Sch{\"o}lkopf, B., and Muandet, K.
\newblock Learning kernel tests without data splitting.
\newblock \emph{Advances in Neural Information Processing Systems}, 33:\penalty0 6245--6255, 2020.

\bibitem[Kuhn et~al.(2023)Kuhn, Gal, and Farquhar]{kuhn2022semantic}
Kuhn, L., Gal, Y., and Farquhar, S.
\newblock Semantic uncertainty: Linguistic invariances for uncertainty estimation in natural language generation.
\newblock In \emph{The Eleventh International Conference on Learning Representations}, 2023.

\bibitem[Lakshminarayanan et~al.(2017)Lakshminarayanan, Pritzel, and Blundell]{lakshminarayanan2017simple}
Lakshminarayanan, B., Pritzel, A., and Blundell, C.
\newblock Simple and scalable predictive uncertainty estimation using deep ensembles.
\newblock \emph{Advances in neural information processing systems}, 30, 2017.

\bibitem[Li et~al.(2017)Li, Chang, Cheng, Yang, and P{\'o}czos]{li2017mmd}
Li, C.-L., Chang, W.-C., Cheng, Y., Yang, Y., and P{\'o}czos, B.
\newblock Mmd gan: Towards deeper understanding of moment matching network.
\newblock \emph{Advances in neural information processing systems}, 30, 2017.

\bibitem[Li et~al.(2015)Li, Swersky, and Zemel]{li2015generative}
Li, Y., Swersky, K., and Zemel, R.
\newblock Generative moment matching networks.
\newblock In \emph{International conference on machine learning}, pp.\  1718--1727. PMLR, 2015.

\bibitem[Li et~al.(2023)Li, Zhang, Zhang, Long, Xie, and Zhang]{li2023towards}
Li, Z., Zhang, X., Zhang, Y., Long, D., Xie, P., and Zhang, M.
\newblock Towards general text embeddings with multi-stage contrastive learning.
\newblock \emph{arXiv preprint arXiv:2308.03281}, 2023.

\bibitem[Lin \& Och(2004)Lin and Och]{lin2004automatic}
Lin, C.-Y. and Och, F.~J.
\newblock Automatic evaluation of machine translation quality using longest common subsequence and skip-bigram statistics.
\newblock In \emph{Proceedings of the 42nd Annual Meeting of the Association for Computational Linguistics (ACL-04)}, pp.\  605--612, 2004.

\bibitem[Liu et~al.(2020)Liu, Xu, Lu, Zhang, Gretton, and Sutherland]{liu2020learning}
Liu, F., Xu, W., Lu, J., Zhang, G., Gretton, A., and Sutherland, D.~J.
\newblock Learning deep kernels for non-parametric two-sample tests.
\newblock In \emph{International conference on machine learning}, pp.\  6316--6326. PMLR, 2020.

\bibitem[Liu \& Yao(1999{\natexlab{a}})Liu and Yao]{liu1999ensemble}
Liu, Y. and Yao, X.
\newblock Ensemble learning via negative correlation.
\newblock \emph{Neural networks}, 12\penalty0 (10):\penalty0 1399--1404, 1999{\natexlab{a}}.

\bibitem[Liu \& Yao(1999{\natexlab{b}})Liu and Yao]{liu1999simultaneous}
Liu, Y. and Yao, X.
\newblock Simultaneous training of negatively correlated neural networks in an ensemble.
\newblock \emph{IEEE Transactions on Systems, Man, and Cybernetics, Part B (Cybernetics)}, 29\penalty0 (6):\penalty0 716--725, 1999{\natexlab{b}}.

\bibitem[Liu et~al.(2015)Liu, Luo, Wang, and Tang]{liu2015deep}
Liu, Z., Luo, P., Wang, X., and Tang, X.
\newblock Deep learning face attributes in the wild.
\newblock In \emph{Proceedings of the IEEE international conference on computer vision}, pp.\  3730--3738, 2015.

\bibitem[Loosli et~al.(2007)Loosli, Canu, and Bottou]{loosli-canu-bottou-2006}
Loosli, G., Canu, S., and Bottou, L.
\newblock Training invariant support vector machines using selective sampling.
\newblock In Bottou, L., Chapelle, O., {DeCoste}, D., and Weston, J. (eds.), \emph{Large Scale Kernel Machines}, pp.\  301--320. MIT Press, Cambridge, MA., 2007.
\newblock URL \url{http://leon.bottou.org/papers/loosli-canu-bottou-2006}.

\bibitem[Malinin \& Gales(2020)Malinin and Gales]{malinin2020uncertainty}
Malinin, A. and Gales, M.
\newblock Uncertainty estimation in autoregressive structured prediction.
\newblock In \emph{International Conference on Learning Representations}, 2020.

\bibitem[Mesk{\'o} \& Topol(2023)Mesk{\'o} and Topol]{mesko2023imperative}
Mesk{\'o}, B. and Topol, E.~J.
\newblock The imperative for regulatory oversight of large language models (or generative ai) in healthcare.
\newblock \emph{npj Digital Medicine}, 6\penalty0 (1):\penalty0 120, 2023.

\bibitem[Murphy(2022)]{pml1Book}
Murphy, K.~P.
\newblock \emph{Probabilistic Machine Learning: An introduction}.
\newblock MIT Press, 2022.
\newblock URL \url{probml.ai}.

\bibitem[Ning et~al.(2019)Ning, He, Wu, Xing, and Zhang]{ning2019review}
Ning, Y., He, S., Wu, Z., Xing, C., and Zhang, L.-J.
\newblock A review of deep learning based speech synthesis.
\newblock \emph{Applied Sciences}, 9\penalty0 (19):\penalty0 4050, 2019.

\bibitem[OpenAI(2023)]{openai2023gpt4}
OpenAI.
\newblock Gpt-4 technical report, 2023.

\bibitem[Paszke et~al.(2019)Paszke, Gross, Massa, Lerer, Bradbury, Chanan, Killeen, Lin, Gimelshein, Antiga, Desmaison, Kopf, Yang, DeVito, Raison, Tejani, Chilamkurthy, Steiner, Fang, Bai, and Chintala]{NEURIPS2019_bdbca288}
Paszke, A., Gross, S., Massa, F., Lerer, A., Bradbury, J., Chanan, G., Killeen, T., Lin, Z., Gimelshein, N., Antiga, L., Desmaison, A., Kopf, A., Yang, E., DeVito, Z., Raison, M., Tejani, A., Chilamkurthy, S., Steiner, B., Fang, L., Bai, J., and Chintala, S.
\newblock Pytorch: An imperative style, high-performance deep learning library.
\newblock In Wallach, H., Larochelle, H., Beygelzimer, A., d\textquotesingle Alch\'{e}-Buc, F., Fox, E., and Garnett, R. (eds.), \emph{Advances in Neural Information Processing Systems 32}, 2019.

\bibitem[Paul et~al.(2021)Paul, Sanap, Shenoy, Kalyane, Kalia, and Tekade]{paul2021artificial}
Paul, D., Sanap, G., Shenoy, S., Kalyane, D., Kalia, K., and Tekade, R.~K.
\newblock Artificial intelligence in drug discovery and development.
\newblock \emph{Drug discovery today}, 26\penalty0 (1):\penalty0 80, 2021.

\bibitem[Radford et~al.(2015)Radford, Metz, and Chintala]{radford2015unsupervised}
Radford, A., Metz, L., and Chintala, S.
\newblock Unsupervised representation learning with deep convolutional generative adversarial networks.
\newblock \emph{arXiv preprint arXiv:1511.06434}, 2015.

\bibitem[Rame et~al.(2022)Rame, Kirchmeyer, Rahier, Rakotomamonjy, Gallinari, and Cord]{rame2022diverse}
Rame, A., Kirchmeyer, M., Rahier, T., Rakotomamonjy, A., Gallinari, P., and Cord, M.
\newblock Diverse weight averaging for out-of-distribution generalization.
\newblock \emph{Advances in Neural Information Processing Systems}, 35:\penalty0 10821--10836, 2022.

\bibitem[Ramesh et~al.(2021)Ramesh, Pavlov, Goh, Gray, Voss, Radford, Chen, and Sutskever]{ramesh2021zero}
Ramesh, A., Pavlov, M., Goh, G., Gray, S., Voss, C., Radford, A., Chen, M., and Sutskever, I.
\newblock Zero-shot text-to-image generation.
\newblock In \emph{International Conference on Machine Learning}, pp.\  8821--8831. PMLR, 2021.

\bibitem[Reddy et~al.(2019)Reddy, Chen, and Manning]{reddy2019coqa}
Reddy, S., Chen, D., and Manning, C.~D.
\newblock Coqa: A conversational question answering challenge.
\newblock \emph{Transactions of the Association for Computational Linguistics}, 7:\penalty0 249--266, 2019.

\bibitem[Ren et~al.(2016)Ren, Zhu, Li, and Luo]{ren2016conditional}
Ren, Y., Zhu, J., Li, J., and Luo, Y.
\newblock Conditional generative moment-matching networks.
\newblock \emph{Advances in Neural Information Processing Systems}, 29, 2016.

\bibitem[Ren et~al.(2021)Ren, Luo, and Zhu]{ren2021improving}
Ren, Y., Luo, Y., and Zhu, J.
\newblock Improving generative moment matching networks with distribution partition.
\newblock In \emph{Proceedings of the AAAI Conference on Artificial Intelligence}, volume~35, pp.\  9403--9410, 2021.

\bibitem[S{\"a}rndal et~al.(2003)S{\"a}rndal, Swensson, and Wretman]{sarndal2003model}
S{\"a}rndal, C.-E., Swensson, B., and Wretman, J.
\newblock \emph{Model assisted survey sampling}.
\newblock Springer Science \& Business Media, 2003.

\bibitem[Sch{\"o}lkopf(1997)]{scholkopf1997support}
Sch{\"o}lkopf, B.
\newblock \emph{Support vector learning}.
\newblock PhD thesis, Oldenbourg M{\"u}nchen, Germany, 1997.

\bibitem[Sch{\"o}lkopf \& Smola(2002)Sch{\"o}lkopf and Smola]{scholkopf2002learning}
Sch{\"o}lkopf, B. and Smola, A.~J.
\newblock \emph{Learning with kernels: support vector machines, regularization, optimization, and beyond}.
\newblock MIT press, 2002.

\bibitem[Schrab et~al.(2022)Schrab, Kim, Guedj, and Gretton]{schrab2022efficient}
Schrab, A., Kim, I., Guedj, B., and Gretton, A.
\newblock Efficient aggregated kernel tests using incomplete $ u $-statistics.
\newblock \emph{Advances in Neural Information Processing Systems}, 35:\penalty0 18793--18807, 2022.

\bibitem[Schrab et~al.(2023)Schrab, Kim, Albert, Laurent, Guedj, and Gretton]{schrab2023mmd}
Schrab, A., Kim, I., Albert, M., Laurent, B., Guedj, B., and Gretton, A.
\newblock Mmd aggregated two-sample test.
\newblock \emph{Journal of Machine Learning Research (JMLR)}, 24, 2023.

\bibitem[Shao(2003)]{shao2003mathematical}
Shao, J.
\newblock \emph{Mathematical statistics}.
\newblock Springer Science \& Business Media, 2003.

\bibitem[Shekhar et~al.(2022)Shekhar, Kim, and Ramdas]{shekhar2022permutation}
Shekhar, S., Kim, I., and Ramdas, A.
\newblock A permutation-free kernel two-sample test.
\newblock \emph{Advances in Neural Information Processing Systems}, 35:\penalty0 18168--18180, 2022.

\bibitem[Song et~al.(2020)Song, Tan, Qin, Lu, and Liu]{song2020mpnet}
Song, K., Tan, X., Qin, T., Lu, J., and Liu, T.-Y.
\newblock Mpnet: Masked and permuted pre-training for language understanding.
\newblock \emph{Advances in Neural Information Processing Systems}, 33:\penalty0 16857--16867, 2020.

\bibitem[Steinbach et~al.(2000)Steinbach, Karypis, and Kumar]{steinbach2000comparison}
Steinbach, M., Karypis, G., and Kumar, V.
\newblock A comparison of document clustering techniques.
\newblock Technical report, Department of Computer Science and Engineering, University of Minnesota, 2000.

\bibitem[Steinwart \& Ziegel(2021)Steinwart and Ziegel]{steinwart2021strictly}
Steinwart, I. and Ziegel, J.~F.
\newblock Strictly proper kernel scores and characteristic kernels on compact spaces.
\newblock \emph{Applied and Computational Harmonic Analysis}, 51:\penalty0 510--542, 2021.

\bibitem[Tseng et~al.(2023)Tseng, Li, Kim, Alsisan, Huang, and Kopf]{tseng2023consistent}
Tseng, H.-Y., Li, Q., Kim, C., Alsisan, S., Huang, J.-B., and Kopf, J.
\newblock Consistent view synthesis with pose-guided diffusion models.
\newblock In \emph{Proceedings of the IEEE/CVF Conference on Computer Vision and Pattern Recognition}, pp.\  16773--16783, 2023.

\bibitem[Ueda \& Nakano(1996)Ueda and Nakano]{ueda1996generalization}
Ueda, N. and Nakano, R.
\newblock Generalization error of ensemble estimators.
\newblock In \emph{Proceedings of International Conference on Neural Networks (ICNN'96)}, volume~1, pp.\  90--95. IEEE, 1996.

\bibitem[Wang et~al.(2022)Wang, Yang, Huang, Jiao, Yang, Jiang, Majumder, and Wei]{wang2022text}
Wang, L., Yang, N., Huang, X., Jiao, B., Yang, L., Jiang, D., Majumder, R., and Wei, F.
\newblock Text embeddings by weakly-supervised contrastive pre-training.
\newblock \emph{arXiv preprint arXiv:2212.03533}, 2022.

\bibitem[Wang et~al.(2020)Wang, Wei, Dong, Bao, Yang, and Zhou]{wang2020minilm}
Wang, W., Wei, F., Dong, L., Bao, H., Yang, N., and Zhou, M.
\newblock Minilm: Deep self-attention distillation for task-agnostic compression of pre-trained transformers.
\newblock \emph{Advances in Neural Information Processing Systems}, 33:\penalty0 5776--5788, 2020.

\bibitem[Wolf et~al.(2020)Wolf, Debut, Sanh, Chaumond, Delangue, Moi, Cistac, Rault, Louf, Funtowicz, et~al.]{Wolf_Transformers_State-of-the-Art_Natural_2020}
Wolf, T., Debut, L., Sanh, V., Chaumond, J., Delangue, C., Moi, A., Cistac, P., Rault, T., Louf, R., Funtowicz, M., et~al.
\newblock Transformers: State-of-the-art natural language processing.
\newblock In \emph{Proceedings of the 2020 conference on empirical methods in natural language processing: system demonstrations}, pp.\  38--45, 2020.

\bibitem[Wu \& Shang(2020)Wu and Shang]{wu2020managing}
Wu, J. and Shang, S.
\newblock Managing uncertainty in ai-enabled decision making and achieving sustainability.
\newblock \emph{Sustainability}, 12\penalty0 (21):\penalty0 8758, 2020.

\bibitem[Zhang et~al.(2022)Zhang, Roller, Goyal, Artetxe, Chen, Chen, Dewan, Diab, Li, Lin, et~al.]{zhang2022opt}
Zhang, S., Roller, S., Goyal, N., Artetxe, M., Chen, M., Chen, S., Dewan, C., Diab, M., Li, X., Lin, X.~V., et~al.
\newblock Opt: Open pre-trained transformer language models.
\newblock \emph{arXiv preprint arXiv:2205.01068}, 2022.

\end{thebibliography}
\bibliographystyle{icml2024}

\newpage
\appendix
\onecolumn

\section{Overview}
\label{app:overview}

In the following, we offer an extended discussion of kernel scores and MMD literature in Appendix \ref{app:ext_MMD}, further theoretical results in Appendix \ref{app:ext_theoretical_results}, more experimental details and additional experiments in Appendix \ref{app:experiments}, and we provide all missing proofs in Appendix \ref{app:proofs}.

\section{Related Work on Kernel Scores and Maximum Mean Discrepancy}
\label{app:ext_MMD}

The kernel score and the MMD$^2$ only differ by a target dependent constant \citep{steinwart2021strictly}.
Consequently, the gradient of the kernel score and the MMD$^2$ are equal, and, thus, optimizing kernel score or MMD$^2$ gives similar results. Importantly, \citet{binkowski2018demystifying} demonstrated that a specific setup of the MMD$^2$, referred to as kernel inception distance (KID), is a better metric for image generation than the commonly used Fréchet inception distance (FID) \citep{heusel2017gans}.
This directly motivates the use of the MMD$^2$ (and thereby the kernel score) as loss/metric to evaluate generative models.
The MMD$^2$ can be intuitively explained via the kernel trick (Schölkopf, 2002): A kernel can be decomposed into an inner product, which turns the MMD$^2$ into the squared euclidean distance within this inner product space (referred to as reproducing kernel Hilbert space, c.f. Equation \eqref{eq:mmd_rkhs}).
The specific form of the inner product space depends on the chosen kernel.
Originally, \citet{JMLR:v13:gretton12a} introduced the MMD for non-parametric hypothesis tests.
\citet{li2015generative} and \citet{dziugaite2015training} simultaneously demonstrated that the MMD can be used to train state-of-the-art (w.r.t. 2015) image generators.
These models are usually referred to as moment matching networks and successive improvements have been proposed \citep{ren2016conditional, li2017mmd, ren2021improving}.
Even though contemporary state-of-the-art image generation training optimizes a different loss, the MMD is still relevant for image generation model selection via the KID \citep{karras2020training, choi2022perception, tseng2023consistent}.

\section{Extended Theoretical Results}
\label{app:ext_theoretical_results}

In this section, we provide more minor theoretical results, which are very close to our main contribution in Theorem \ref{th:bvcd}.

\subsection{Decomposition in Reproducing Kernel Hilbert Spaces}
\label{app:rkhs_decomp}

The literature on MMD expanded to a significant size in the last decade \citep{JMLR:v13:gretton12a, gretton2012optimal, chwialkowski2016kernel, liu2020learning, kubler2020learning, shekhar2022permutation, schrab2022efficient, schrab2023mmd}.
The MMD is usually used in the context of a reproducing kernel Hilbert spaces (RKHS) \citep{scholkopf2002learning}. 
In the following, we express Theorem \ref{th:bvcd} according to a RKHS $\mathcal{H}$ to offer an alternative perspective on our result.
Assume the kernel $k \colon \mathcal{Y} \times \mathcal{Y} \to \mathbb{R}$ is associated with an RKHS $\mathcal{H}$ with inner product $\left\langle ., . \right\rangle_{\mathcal{H}}$ and norm $\left\lVert . \right\rVert_\mathcal{H}$ such that $k\left( x, y \right) = \left\langle k \left(x, . \right), k \left(y, . \right) \right\rangle_{\mathcal{H}}$.
The norm based on $k$ in the distribution space relates to the RKHS norm via $\left\lVert Q \right\rVert_k = \left\lVert \mu_Q \right\rVert_{\mathcal{H}}$ with mean embedding $\mu_Q \coloneqq \mathbb{E} \left[ k \left( Y, . \right) \right] \in \mathcal{H}$ for a $Y \sim Q \in \mathcal{P}$.
Consequently, given a prediction $\hat{P}$ we have
\begin{equation}
    \underbrace{\mathbb{E} \left[ S_k \left(\hat{P}, Y \right) \right]}_{\text{Generalization Error}} = \underbrace{- \left\lVert \mu_Q \right\rVert^2_{\mathcal{H}}}_{\text{Noise}} + \underbrace{\left\lVert \mathbb{E} \left[ \mu_{\hat{P}} \right] - \mu_Q \right\rVert_{\mathcal{H}}^2}_{\text{Bias}} + \underbrace{\mathbb{E} \left[ \left\lVert \mu_{\hat{P}} - \mathbb{E} \left[ \mu_{\hat{P}} \right] \right\rVert_{\mathcal{H}}^2 \right]}_{\text{Variance}}.
\label{eq:mmd_bvd}
\end{equation}
The covariance decomposition can be expressed similarly since $\Braket{ P | k | Q} = \left\langle \mu_P, \mu_Q \right\rangle_{\mathcal{H}}$.
Note that the bias and variance terms in Theorem \ref{th:bvcd} and Equation \eqref{eq:mmd_bvd} are equal.
As a further note, we can also express the kernel score and the MMD for $P,Q \in \mathcal{P}$ and $y \in \mathcal{Y}$ in terms of the RKHS since it holds
\begin{equation}
    S_k \left( P, y \right) = \left\lVert \mu_P \right\rVert_{\mathcal{H}}^2 - 2 \left\langle \mu_P, k \left(y, . \right) \right\rangle_{\mathcal{H}}
\end{equation}
and
\begin{equation}
    \operatorname{MMD}^2_k \left( P, Q \right) = \left\lVert \mu_P - \mu_Q \right\rVert_{\mathcal{H}}^2
\label{eq:mmd_rkhs}
\end{equation}
(c.f. \cite{steinwart2021strictly} for a more detailed discussion).

\subsection{Covariance decomposition without Identical Distribution Assumption}

Let $k$ be a p.s.d. kernel and $\hat{P}$ a predicted distribution for a target $Y \sim Q$ similar as in Theorem \ref{th:bvcd}.

If we have an ensemble prediction $\hat{P}^{\left( n \right)} \coloneqq \frac{1}{n} \sum_{i=1}^n \hat{P}_i$ with members $\hat{P}_1, \dots, \hat{P}_n$, then
\begin{equation}
    \operatorname{Var}_k \left( \hat{P}^{\left( n \right)} \right) = \frac{1}{n^2} \sum_{i=1}^n \operatorname{Var}_k \left( \hat{P}_i \right) + \frac{1}{n^2} \sum_{i=1}^n \sum_{\substack{j=1 \\ j \neq i}}^n \operatorname{Cov}_k \left( \hat{P}_i, \hat{P}_j \right).
\end{equation}

This results is part of the proof in Equation \eqref{eq:bvcd_proof} of Theorem \ref{th:bvcd}.

\section{Extended Experiments}
\label{app:experiments}

In this section, we give more details on the experiments and show further results.

\subsection{Experimental Details}

We give some additional details on the experimental setup.
First, we formalize computing the predictive kernel entropy in Algorithm \ref{alg:kent_est}.
Second, we describe in more detail the image, audio, and natural language generation experiments.

\begin{algorithm}[tb]
   \caption{Estimating predictive kernel entropy}
   \label{alg:example}
\begin{algorithmic}
   \STATE {\bfseries Required:} Generated outputs $a_1, \dots, a_n$ for a given input, p.s.d. kernel $k$, (optional) embedder $\phi$
   \IF{$\phi$ is given}
   \FOR{$i=1$ {\bfseries to} $n$}
   \STATE $a_i \gets \phi \left( a_i \right)$
   \ENDFOR
   \ENDIF
   \STATE {\bfseries Return:} $- \frac{1}{n \left(n-1 \right)} \sum_{i=1}^n \sum_{j \neq i}^n k \left( a_i, a_j \right)$
\label{alg:kent_est}
\end{algorithmic}
\end{algorithm}

\subsubsection{Image Generation}
\label{app:experiments:img}

We use the following procedure for the simulation in Figure \ref{fig:var_sim}.
First, we sample 32 distinct training sets from InfiMNIST of size 60.000.
Then, we train a conditional diffusion model on each training set for 20 epochs.
We adopt implementation and hyperparameters from open source PyTorch code of a conditional diffusion model trained on normal MNIST \citep{NEURIPS2019_bdbca288}.
This includes an initial learning rate of 1e-4, a batch size of 256, 400 diffusion steps, and a feature dimension of 128.
We then generate 100 images of class '0' of each model after training.
Finally, to get the approximate standard deviation of each tick and each line in Figure \ref{fig:var_sim}, we estimate the distributional variance 1000 times on randomly drawn samples without replacement of all generated images.
The normalization constant in the RBF kernel is set to $\gamma = \frac{1}{728}$ since each image has 728 pixels.
We also repeated the whole procedure for other classes with similar results.

The results in Figure \ref{fig:metrics_per_epoch} and \ref{fig:scatter_plots} are produced in a similar manner.
Only difference here is that we train on only 20 training sets for 40 epochs and generate only 20 images per class.
We chose these numbers based on the insights gained by the previous simulation experiment.
The correlation matrix is the average of all class-wise correlation matrices.
The values of the corresponding average covariance matrix of epochs 20 to 40 is given by (all values rounded)

\setcounter{MaxMatrixCols}{11}
\begin{equation}
\left(\operatorname{Cov}_k \left( P_i, P_j \right)\right)_{i,j=20 \dots 40} \approx 10^3 \cdot
\begin{pmatrix}
    5.1 & 4.9 & 4.9 & 4.9 & 4.9 & 4.9 & 4.8 & 4.9 & 4.9 & 4.8 & 4.9 \\
    4.9 & 5.1 & 4.9 & 4.8 & 4.9 & 4.9 & 4.9 & 4.9 & 4.9 & 4.9 & 4.9 \\
    4.9 & 4.9 & 5.1 & 4.8 & 4.9 & 4.9 & 4.8 & 4.9 & 4.9 & 4.8 & 4.9 \\
    4.9 & 4.8 & 4.8 & 5.1 & 4.9 & 4.8 & 4.9 & 4.9 & 4.9 & 4.9 & 4.9 \\
    4.9 & 4.9 & 4.9 & 4.9 & 5.2 & 4.9 & 4.9 & 4.9 & 4.9 & 4.9 & 4.9 \\
    4.9 & 4.9 & 4.9 & 4.8 & 4.9 & 5.2 & 4.9 & 4.9 & 4.9 & 4.9 & 5.0 \\
    4.8 & 4.9 & 4.8 & 4.9 & 4.9 & 4.9 & 5.1 & 4.9 & 4.9 & 4.9 & 4.9 \\
    4.9 & 4.9 & 4.9 & 4.9 & 4.9 & 4.9 & 4.9 & 5.1 & 4.9 & 4.9 & 4.9 \\
    4.9 & 4.9 & 4.9 & 4.9 & 4.9 & 4.9 & 4.9 & 4.9 & 5.1 & 4.9 & 5.0 \\
    4.8 & 4.9 & 4.8 & 4.9 & 4.9 & 4.9 & 4.9 & 4.9 & 4.9 & 5.2 & 5.0 \\
    4.9 & 4.9 & 4.9 & 4.9 & 4.9 & 5.0 & 4.9 & 4.9 & 5.0 & 5.0 & 5.2
\end{pmatrix}.
\label{eq:cov_numbers}
\end{equation}
This covariance matrix confirms that $\operatorname{Var}_k \left( P_i \right)$ and $\operatorname{Cov}_k \left( P_i, P_j \right)$ are approximately equal for all $i,j \in \left\{20, \dots, 40 \right\}$ with $i \neq j$.
We made use of this assumption in Example \ref{ex:cov_numeric}. \\
All training was done on Nvidia RTX5000 GPUs.

\subsubsection{Audio Generation}

For the audio experiments, we use an implementation of Glow-TTS given in the TTS library \citep{Eren_Coqui_TTS_2021}.
The LJSpeech dataset consists of 13.100 instances of text-speech pairs.
Each speech is of a single woman reading out loud the corresponding text.
We use a random 90\% of the data for training and a batch size of 32.
On this single training set, we train 12 randomly initialized models for 100 epochs with an initial learning rate of 1e-2.
The evaluation happens every 2.000 gradient descent iterations for each model.
For a single model in a single evaluation step, we generate 10 waveforms (speeches) for each of 100 test instances.
The normalization constant in the Laplacian and RBF kernel is set to  $\gamma = \frac{1}{\lambda_{\mathrm{iter},\mathrm{inst}}}$, where $\lambda_{\mathrm{iter},\mathrm{inst}}$ is the longest generated waveform in each evaluation step $\mathrm{iter}$ and each test instance $\mathrm{inst}$.
The generated audio instances are of various length, so we pad them with zeros to match their length. \\
Further, all training was done on Nvidia RTX5000 GPUs.
But, noteworthy to this experiment, storing the model iterations and generated waveforms required up to 400 GB of hard disk storage.

\subsubsection{Natural Language Generation}

For the natural language experiments, we adopted the experimental setup of \citet{kuhn2022semantic}.
We used their provided code implementations including the hyperparameters.
This includes a temperature of $T=0.5$ for generating the answers used for uncertainty estimation.
Similarly, we use 10 answer generations for each prompt for TriviaQA and 20 answer generations for CoQA.
All natural language models are pretrained and downloaded from HuggingFace \citep{Wolf_Transformers_State-of-the-Art_Natural_2020}.
We used a single Nvidia A6000 GPU for the natural language experiments.

\begin{figure*}
\centering
    \begin{subfigure}{.3\textwidth}
    \centering
    \includegraphics[width=\columnwidth]{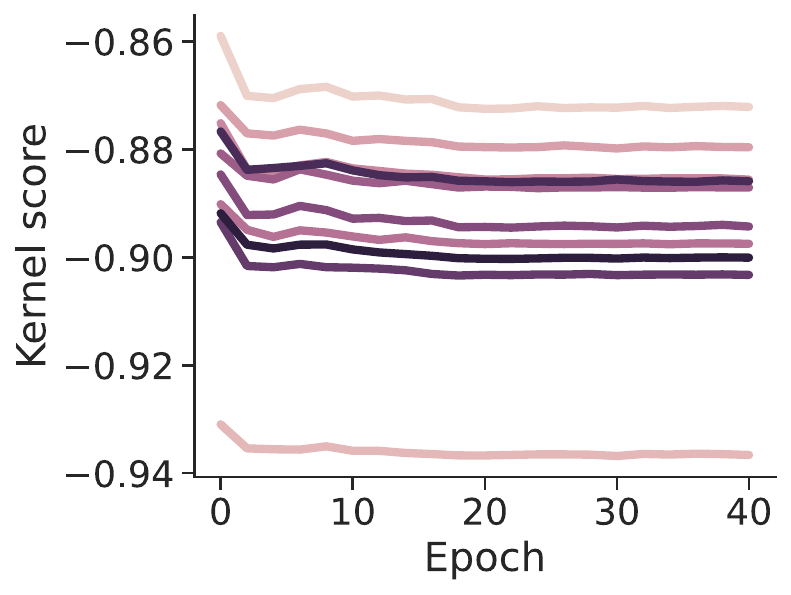}
    \end{subfigure}%
    \begin{subfigure}{.4\textwidth}
    \centering
    \includegraphics[width=\columnwidth]{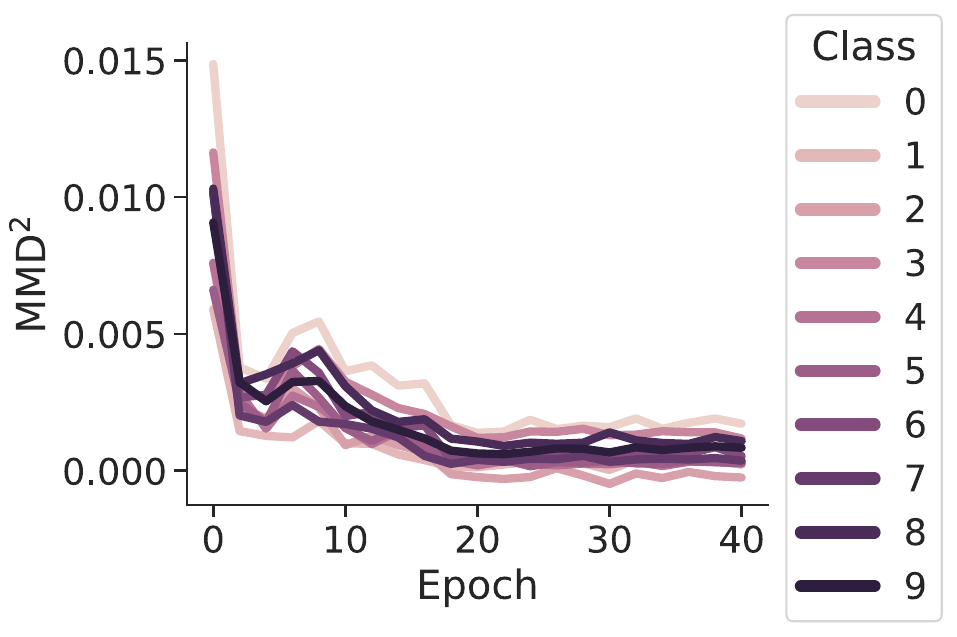}
    \end{subfigure}%
    \begin{subfigure}{.3\textwidth}
    \centering
    \includegraphics[width=\columnwidth]{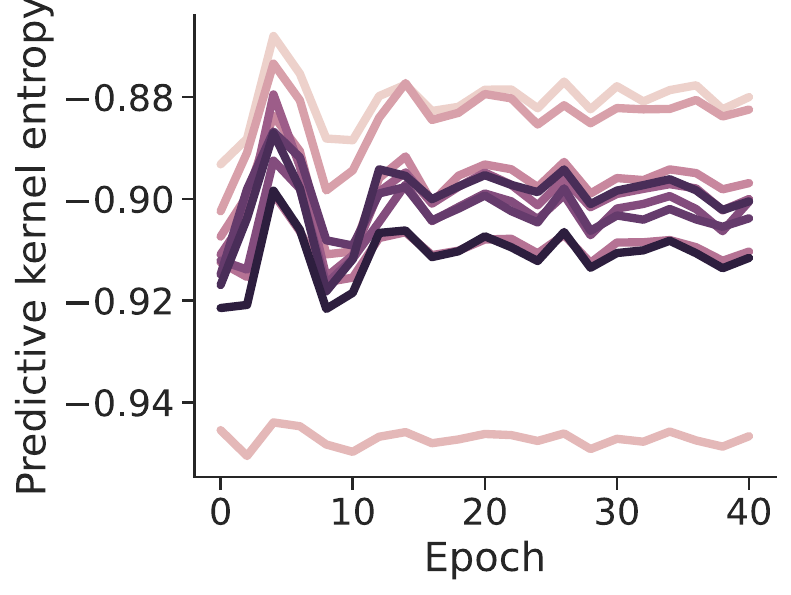}
    \end{subfigure}
\caption{
    \textbf{Left:} Kernel score throughout training. The kernel score cannot be compared meaningfully between classes since each optimum depends on a constant specific to each target class.
    \textbf{Mid:} MMD$^2$ throughout training. Here, it is easier to compare the errors since the MMD$^2$ is zero for the optimal prediction.
    \textbf{Right:} The predictive kernel entropy of each class fluctuates throughout training but stays constant upon convergence.
}
\label{fig:kscore_mmd_per_epoch_ext}
\end{figure*}

\begin{figure*}
\centering
    \begin{subfigure}{.4\textwidth}
    \centering
    \includegraphics[width=\columnwidth]{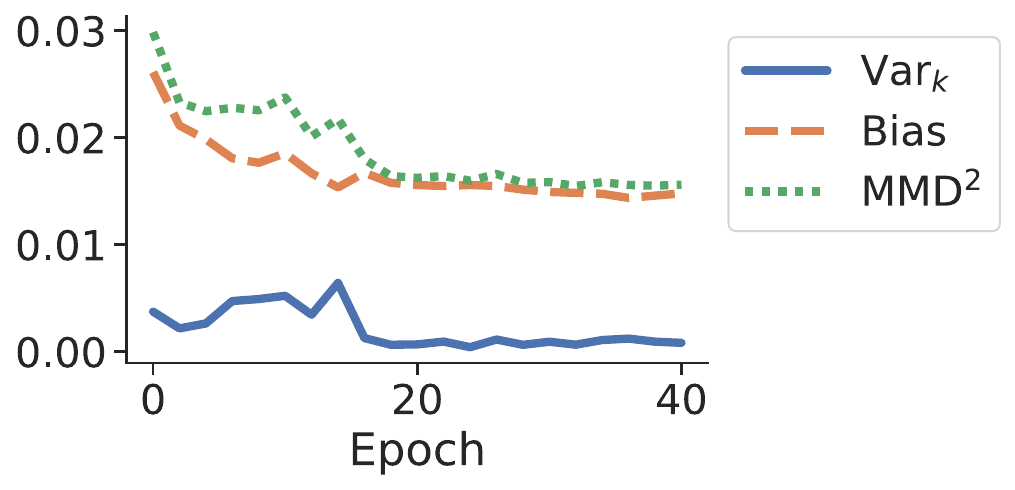}
    \end{subfigure}%
    \begin{subfigure}{.4\textwidth}
    \centering
    \includegraphics[width=\columnwidth]{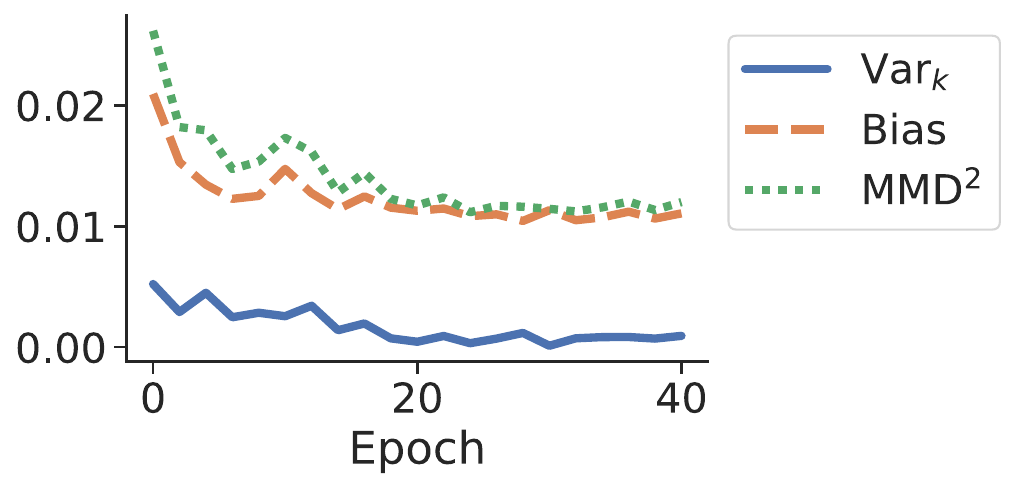}
    \end{subfigure}
\caption{
Generalization curves of the digit, which we reduce in frequency during training of DDPMs on InfiMNIST.
The reduced digit has only approx. 60 training instances while the other digits have approx. 6000 each.
\textbf{Left:} We reduce and evaluate exclusively digit '2'. \textbf{Right:} We reduce and evaluate exclusively digit '3'. We cannot observe a mode collapse as with digit '0' in Figure \ref{fig:mode_collapse}. The bias is still elevated compared to normal digit frequencies (c.f. Figure \ref{fig:metrics_per_epoch}).
}
\label{fig:img_reduced_freq}
\end{figure*}


\subsection{Additional Results}

In the following, we give some additional results for image, audio, and language generation.

\subsubsection{Image Generation}

\begin{figure*}
\centering
    \begin{subfigure}{.3\textwidth}
    \centering
    \includegraphics[width=\columnwidth]{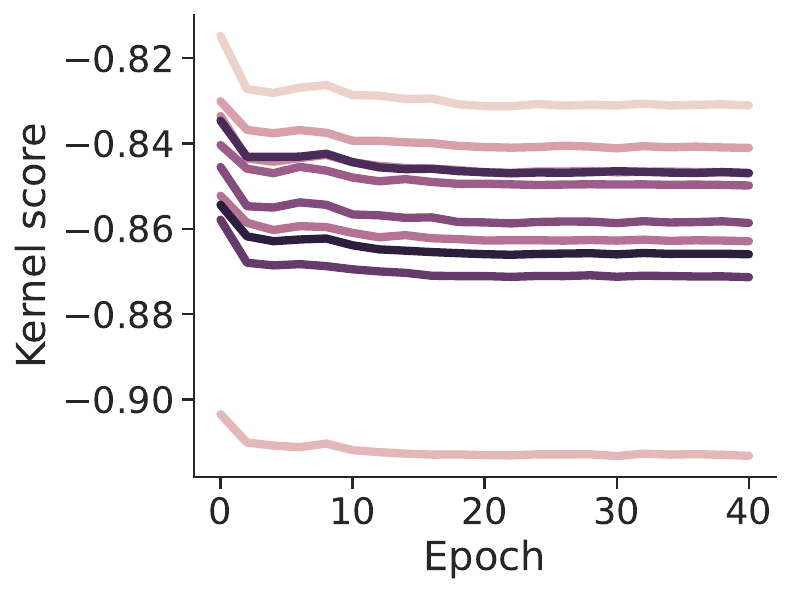}
    \end{subfigure}%
    \begin{subfigure}{.4\textwidth}
    \centering
    \includegraphics[width=\columnwidth]{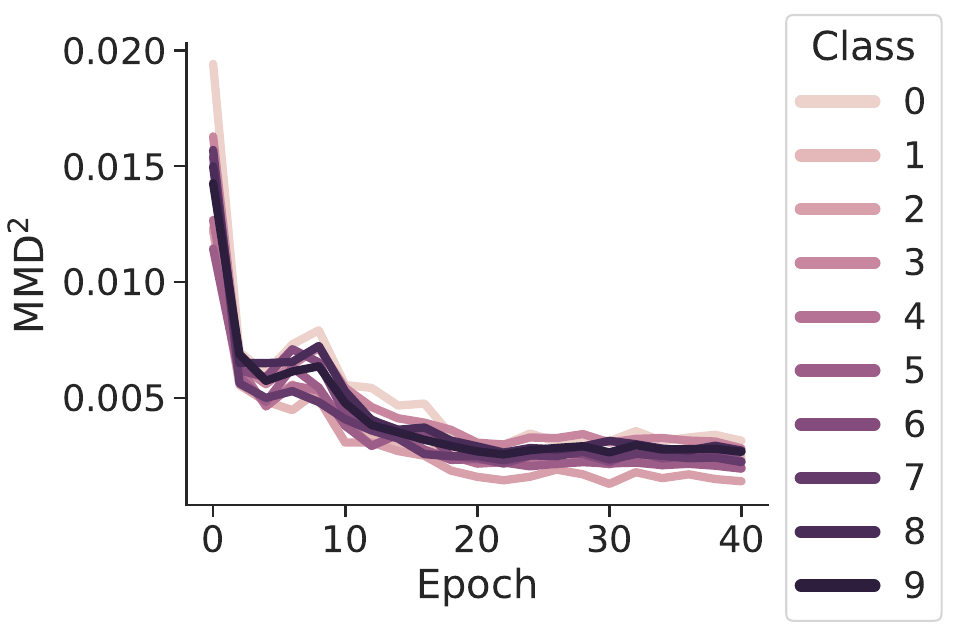}
    \end{subfigure}%
    \begin{subfigure}{.3\textwidth}
    \centering
    \includegraphics[width=\columnwidth]{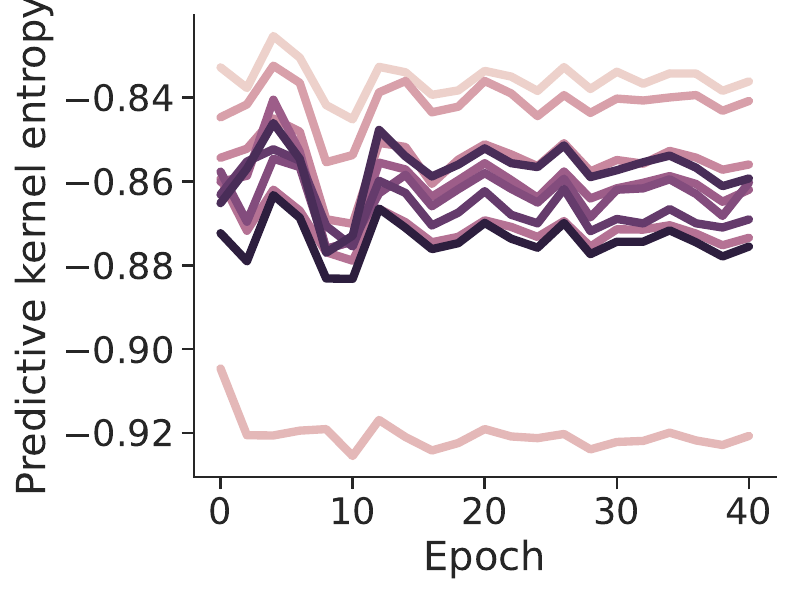}
    \end{subfigure} \\
    \begin{subfigure}{.41\textwidth}
    \centering
    \includegraphics[width=\columnwidth]{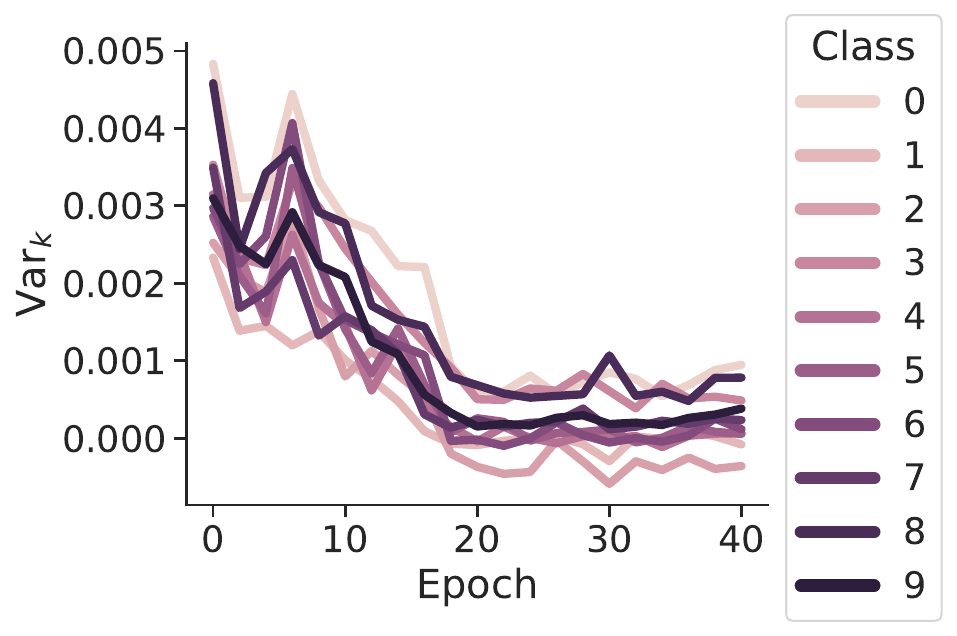}
    \end{subfigure}%
    \begin{subfigure}{.34\textwidth}
    \centering
    \includegraphics[width=\columnwidth]{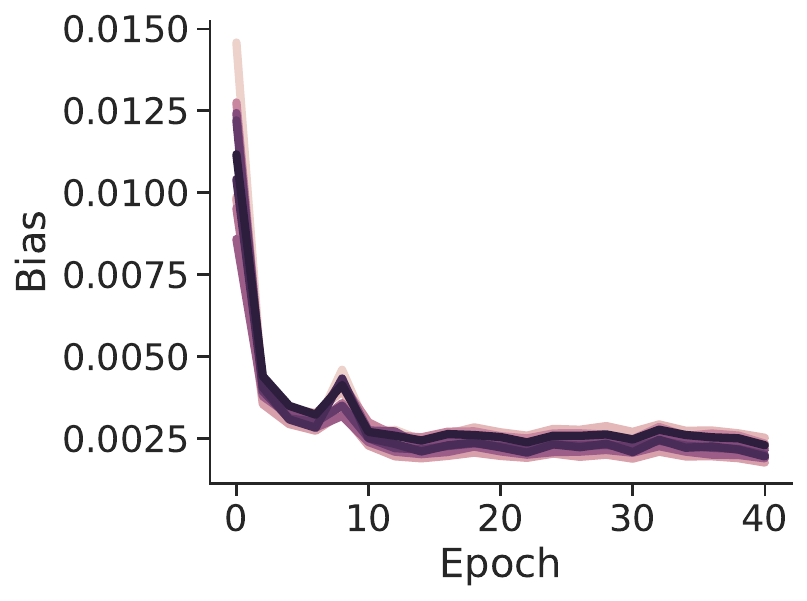}
    \end{subfigure}%
    \begin{subfigure}{.25\textwidth}
    \centering
    \includegraphics[width=\columnwidth]{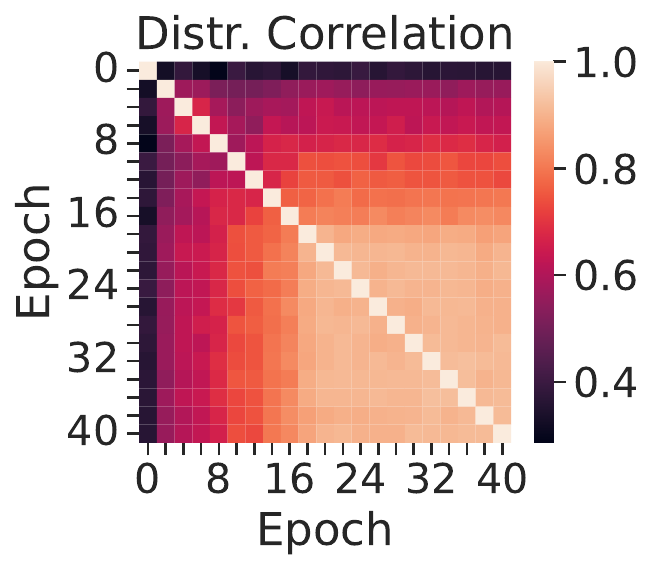}
    \end{subfigure} \\
    \begin{subfigure}{.4\textwidth}
    \centering
    \includegraphics[width=\columnwidth]{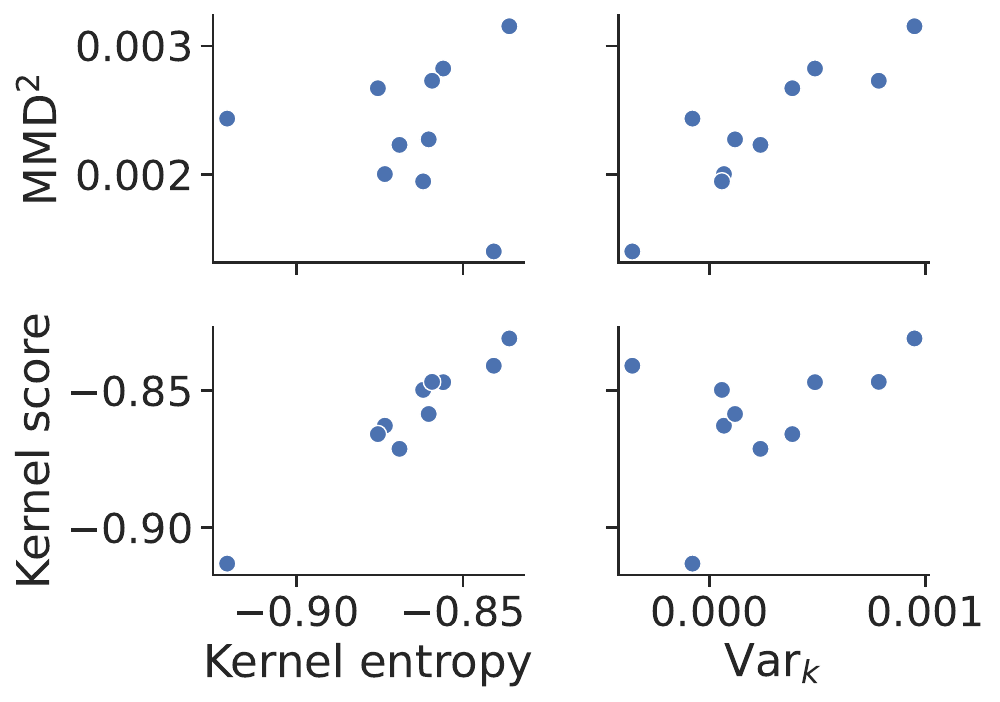}
    \end{subfigure}%
    \begin{subfigure}{.6\textwidth}
    \centering
    \includegraphics[width=\columnwidth]{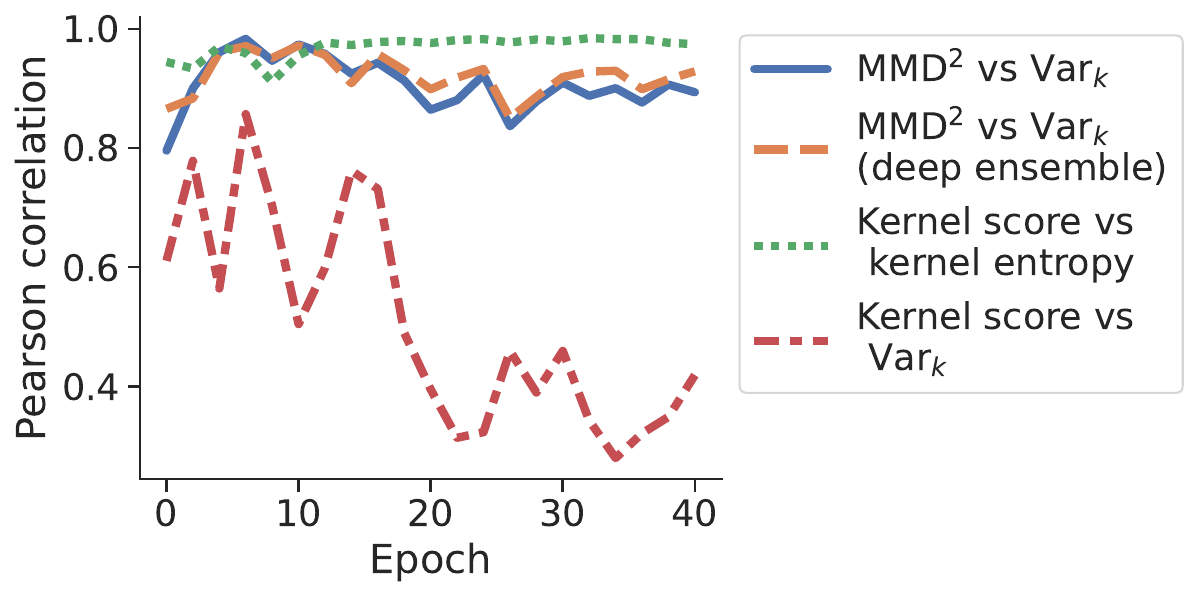}
    \end{subfigure}

\caption{
Evaluations in Figures \ref{fig:metrics_per_epoch}, \ref{fig:scatter_plots}, and \ref{fig:kscore_mmd_per_epoch_ext} repeated with Laplacian kernel. The reported findings based on the RBF kernel also hold for the Laplacian kernel and only minor differences can be spotted.
}
\label{fig:img_gen_lap_kernels}
\end{figure*}

\begin{figure*}
\centering
    \begin{subfigure}{.3\textwidth}
    \centering
    \includegraphics[width=\columnwidth]{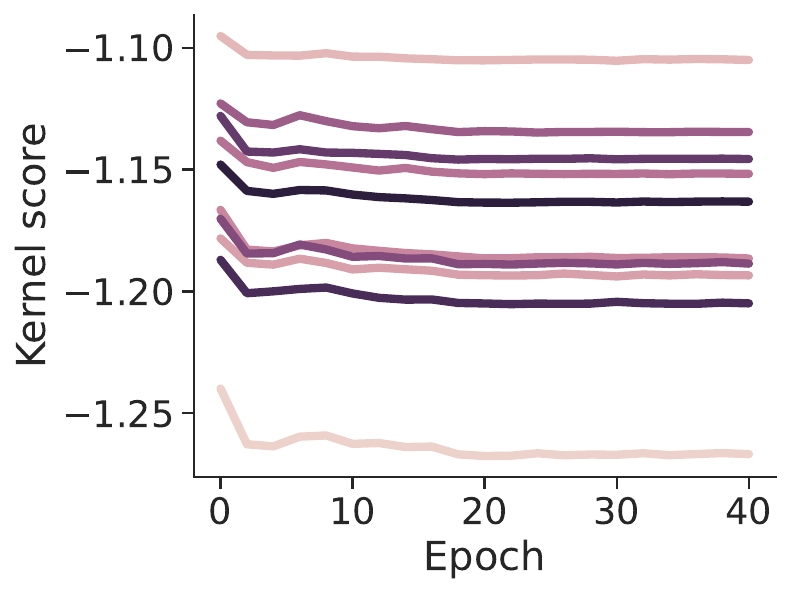}
    \end{subfigure}%
    \begin{subfigure}{.4\textwidth}
    \centering
    \includegraphics[width=\columnwidth]{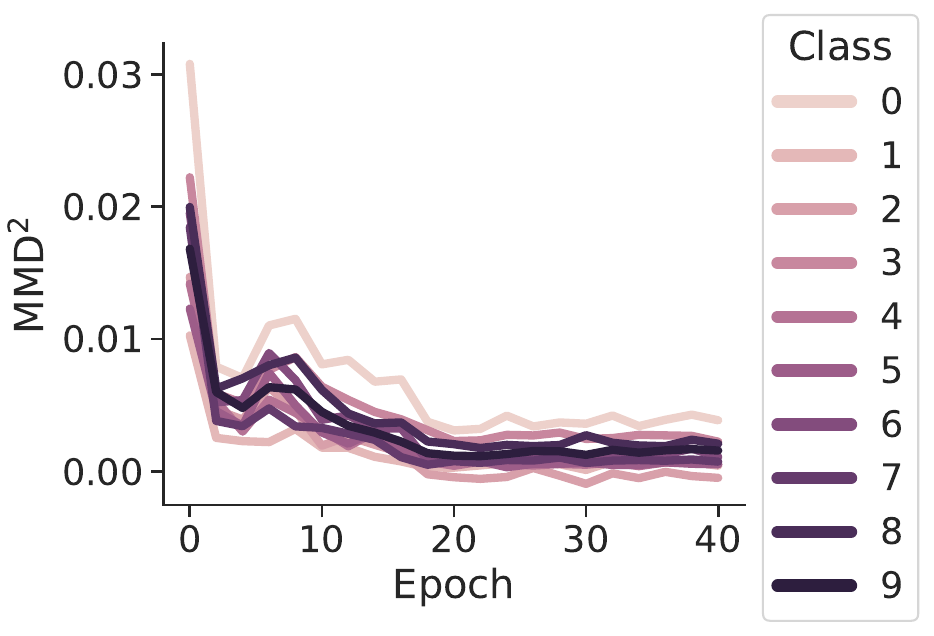}
    \end{subfigure}%
    \begin{subfigure}{.3\textwidth}
    \centering
    \includegraphics[width=\columnwidth]{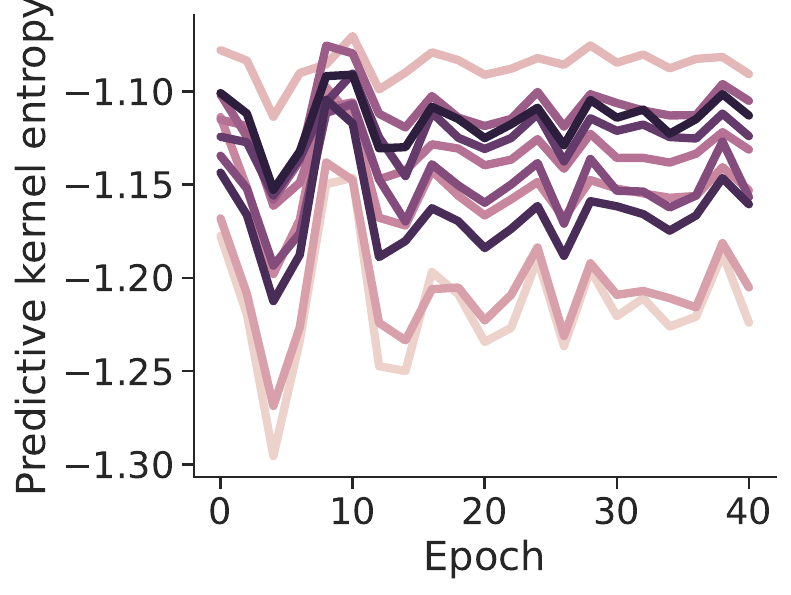}
    \end{subfigure} \\
    \begin{subfigure}{.41\textwidth}
    \centering
    \includegraphics[width=\columnwidth]{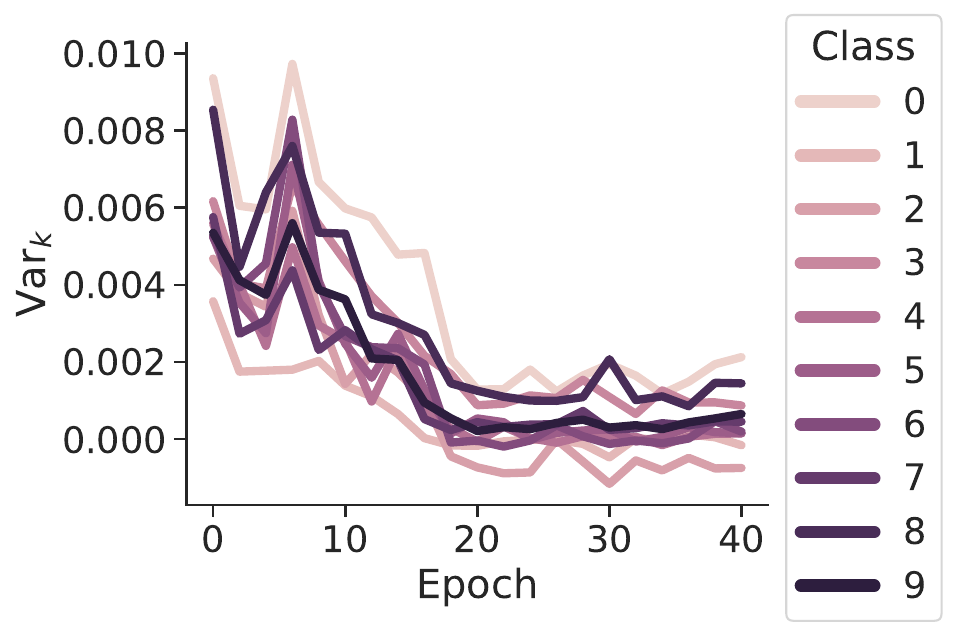}
    \end{subfigure}%
    \begin{subfigure}{.34\textwidth}
    \centering
    \includegraphics[width=\columnwidth]{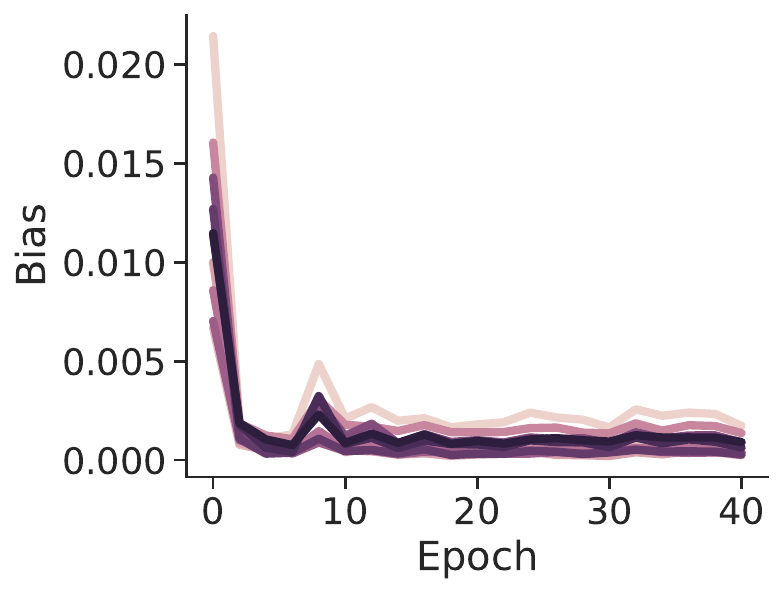}
    \end{subfigure}%
    \begin{subfigure}{.25\textwidth}
    \centering
    \includegraphics[width=\columnwidth]{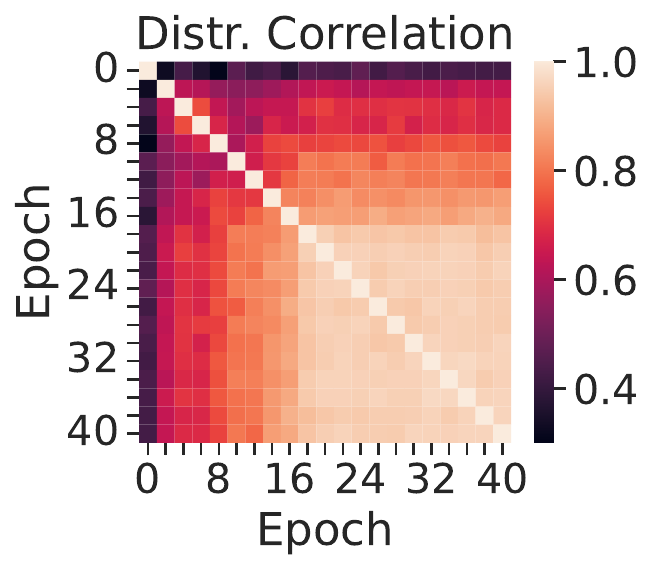}
    \end{subfigure} \\
    \begin{subfigure}{.4\textwidth}
    \centering
    \includegraphics[width=\columnwidth]{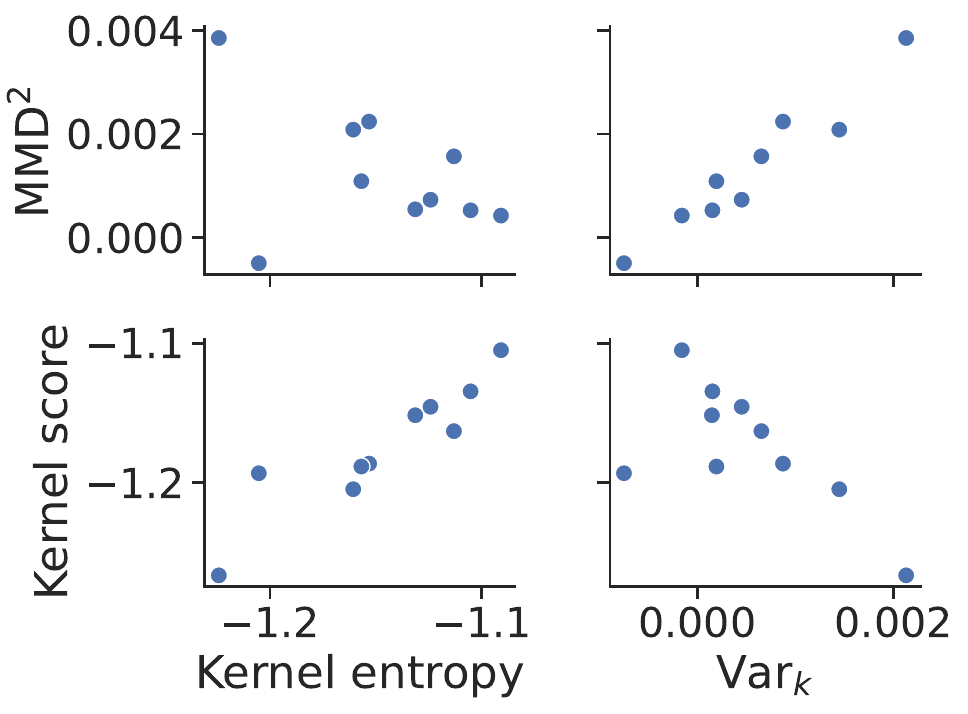}
    \end{subfigure}%
    \begin{subfigure}{.6\textwidth}
    \centering
    \includegraphics[width=\columnwidth]{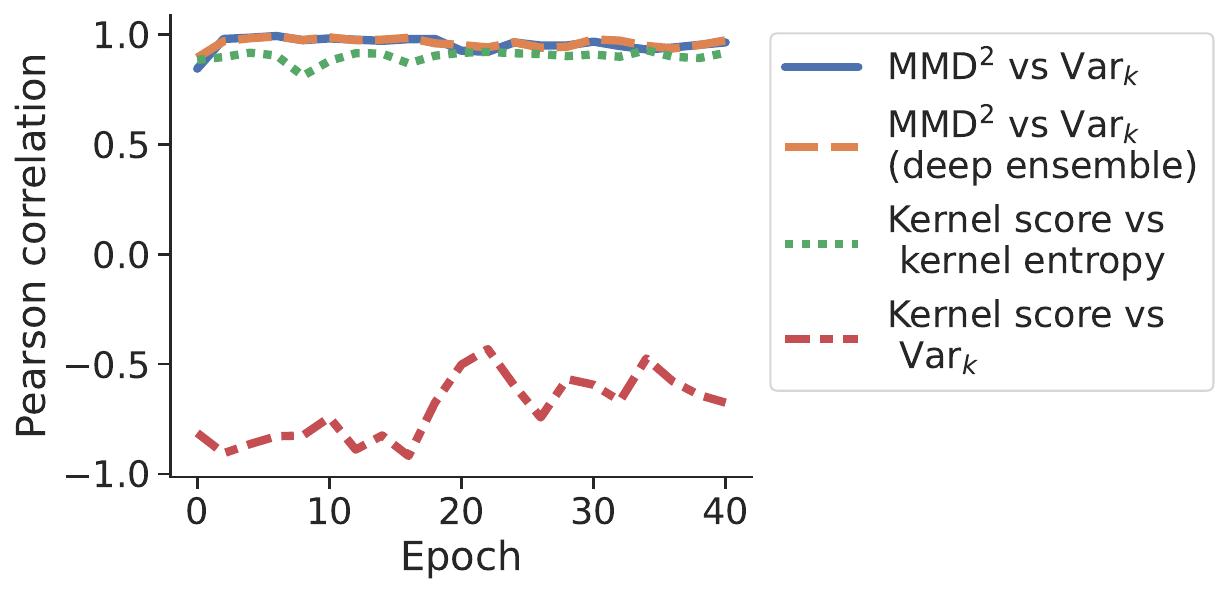}
    \end{subfigure}

\caption{
Evaluations in Figures \ref{fig:metrics_per_epoch}, \ref{fig:scatter_plots}, and \ref{fig:kscore_mmd_per_epoch_ext} repeated with polynomial kernel. The reported findings based on the RBF kernel also hold for the polynomial kernel and only minor differences can be spotted. One such difference is that the distributional variance shows a strong negative correlation with the kernel score, which cannot be seen for the other kernels.
}
\label{fig:img_gen_pol_kernels}
\end{figure*}

\begin{figure*}
\centering
    \begin{subfigure}{.35\textwidth}
    \centering
    \includegraphics[width=\columnwidth]{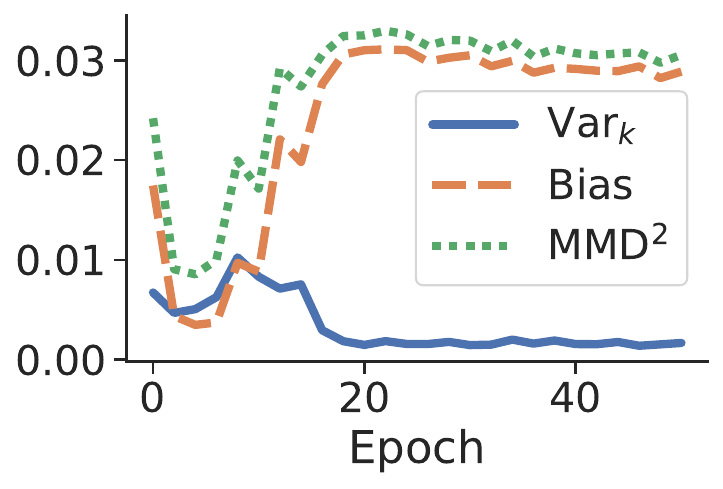}
    \caption{Laplacian kernel.}
    \end{subfigure}%
    \begin{subfigure}{.35\textwidth}
    \centering
    \includegraphics[width=\columnwidth]{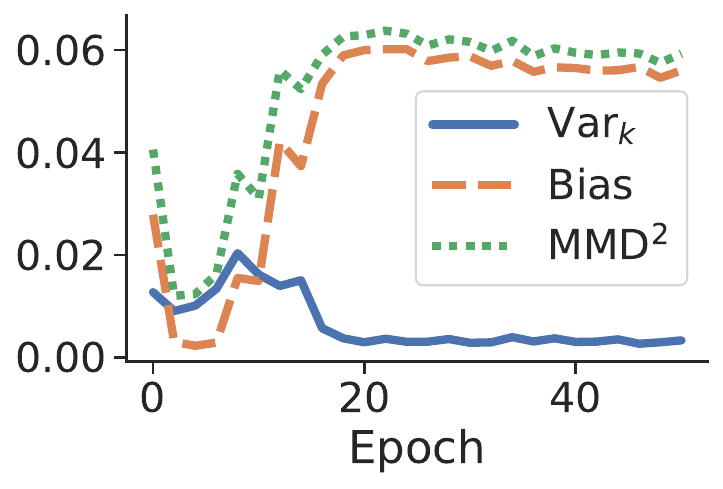}
    \caption{Polynomial kernel.}
    \end{subfigure}
\caption{
    Evaluation of Figure \ref{fig:mode_collapse} repeated with other kernels. The mode collapse result is observable independent of kernel choice.
}
\label{fig:mode_collapse_lap_pol}
\end{figure*}

We start with the image experiments.
In Figure \ref{fig:kscore_mmd_per_epoch_ext}, we show the corresponding generalization error (kernel score), MMD$^2$, and predictive kernel entropy values throughout training of the same setup as in Figure \ref{fig:metrics_per_epoch}.
As can be seen, MMD$^2$ is more interpretative for comparing different classes.
But, the MMD can also not be evaluated in a lot of practical cases including the audio and natural language settings in this work.
Further, the kernel entropy does not show a trend throughout training contrary to the audio setting seen in Figure \ref{fig:audio_per_epoch}.
To test the robustness of our findings with respect to the kernel choice, we repeat all evaluations with the Laplacian kernel with $\gamma = \frac{1}{28^2}$ in Figure \ref{fig:img_gen_lap_kernels}, and the polynomial kernel $k_{\operatorname{pol}} \left(x, y \right) = \left(\frac{\left\langle x, y \right\rangle + 1}{28^2} \right)^3$ in Figure \ref{fig:img_gen_pol_kernels}.
We also repeat the mode collapse evaluation of Figure \ref{fig:mode_collapse} with these kernels in Figure \ref{fig:mode_collapse_lap_pol}.
As can be seen, all reported findings still hold with only minor differences.
This suggests that the choice of kernel is fairly robust towards spotting major trends during model training.

Further, we perform similar experiments with the RBF kernel as in Figure \ref{fig:mode_collapse}, where we observed a mode collapse.
In Figure \ref{fig:img_reduced_freq}, we repeat the evaluation but use digit '2' and digit '3' instead of digit '0'.
As can be seen via our bias-variance decomposition, mode collapse does not occur, but the bias term is still larger than in Figure \ref{fig:metrics_per_epoch}.

\begin{figure*}
\centering
    \begin{subfigure}{.3\textwidth}
    \centering
    \includegraphics[width=\columnwidth]{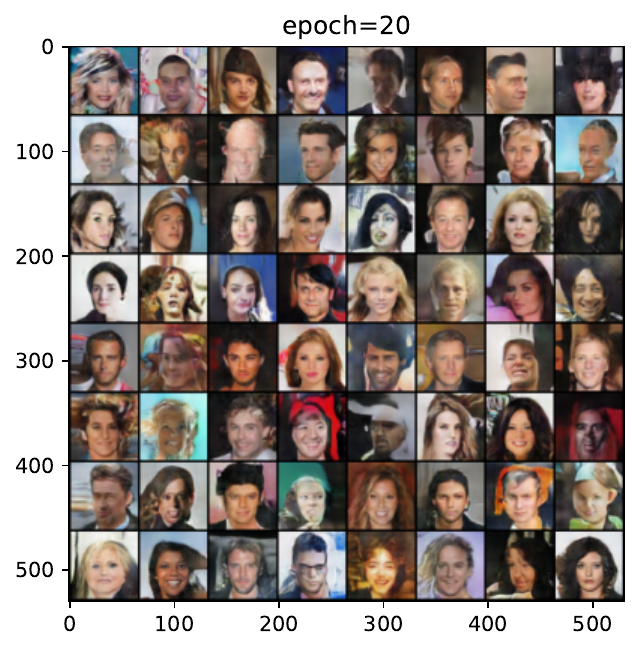}
    \caption{Generated images at Epoch 20 (each row is a different model).}
    \end{subfigure}%
    \begin{subfigure}{.5\textwidth}
    \centering
    \includegraphics[width=\columnwidth]{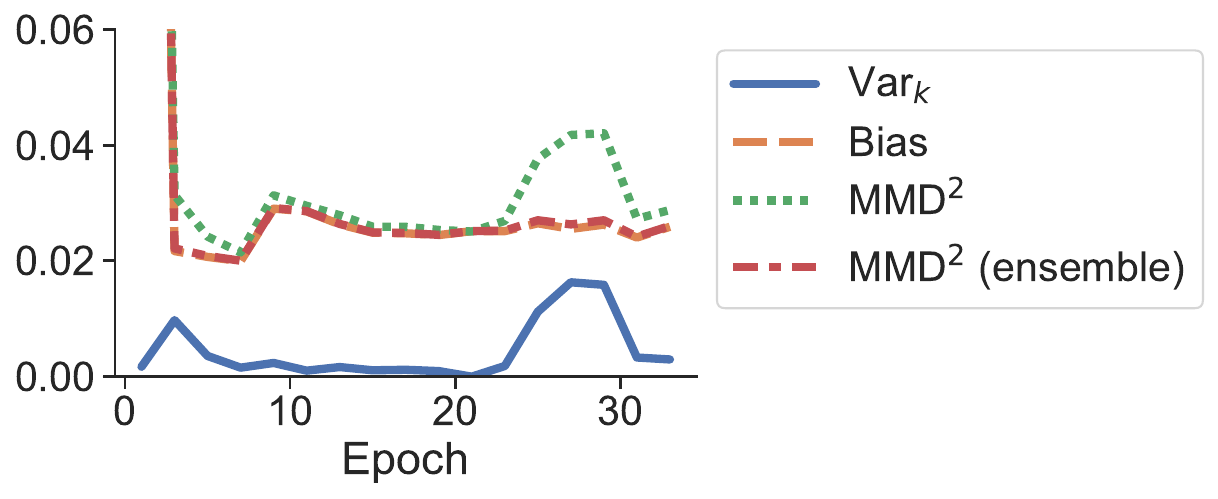}
    \caption{Evaluations throughout training.}
    \end{subfigure}
\caption{
We train an ensemble of 12 DCGAN models on the CelebA dataset and evaluate the MMD$^2$, bias, and variance. The bias and variance are approximated via the ensemble and do not necessarily reflect the ground truth bias and variance. The variance curve indicates the section of improvement in error by using an ensemble instead of a single model as suggested by Theorem \ref{th:bvcd}.
}
\label{fig:celebA}
\end{figure*}

To confirm that our evaluations also give meaningful results for larger setups, we train the DCGAN architecture \citep{radford2015unsupervised} for image generation on the CelebA dataset \citep{liu2015deep}, which consists of colored 64x64 images.
We train an ensemble of 12 models to approximate the bias-variance decomposition.
For each evaluation step, we sample 20 images per model, and we use the Laplacian kernel.
We evaluate the MMD$^2$ of the individual models and of the ensemble of all models, and, as can be seen in Figure \ref{fig:celebA}, the variance indicates the regions in which the ensemble outperforms the individual models in error.
This is predicted by Theorem \ref{th:bvcd} since the ensemble size reduces the variance term in the generalisation error.

This underlines that our decomposition is a useful tool to investigate and analyze the fitting behaviour of generative models.

\subsubsection{Audio Generation}

\begin{figure*}
\centering
    \begin{subfigure}{.4\textwidth}
    \centering
    \includegraphics[width=\columnwidth]{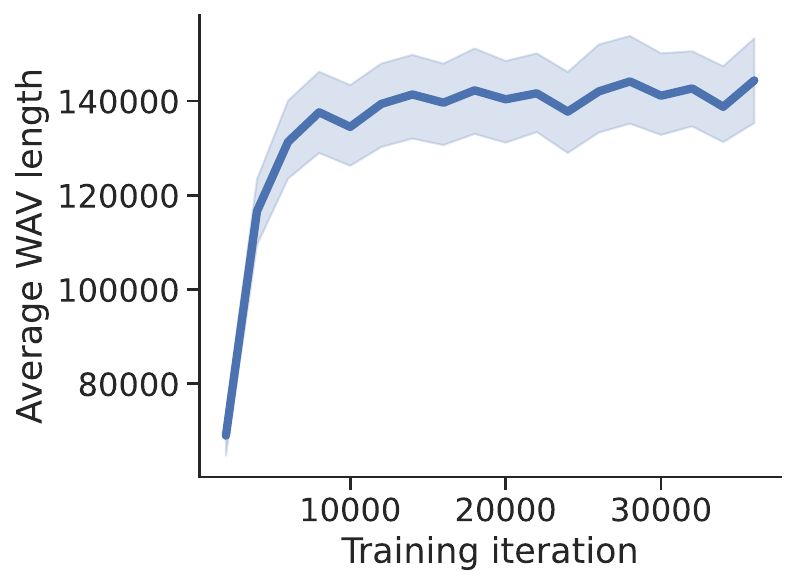}
    \end{subfigure}
\caption{
Average WAV length of generated audio instances by Glow-TTS. Initially, the model generates only very short instances, which explains the low initial kernel entropy.
}
\label{fig:tts_wav_len}
\end{figure*}

We repeat the evaluations of the audio generations in Section \ref{sec:applications_audio}, where we used the Laplacian kernel, with the RBF kernel.
It is known that the RBF kernel does not scale well to higher dimensions \citep{binkowski2018demystifying}.
We represent the generated audio instances via vectors of various lengths, often exceeding 100,000 dimensions (c.f. Figure \ref{fig:tts_wav_len}).

\begin{figure*}
\centering
    \begin{subfigure}{.33\textwidth}
    \centering
    \includegraphics[width=\columnwidth]{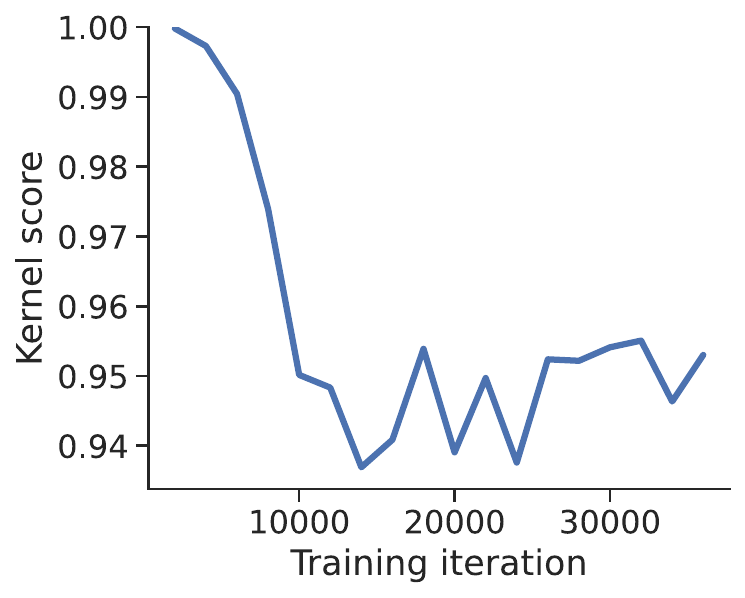}
    \end{subfigure}%
    \begin{subfigure}{.33\textwidth}
    \centering
    \includegraphics[width=\columnwidth]{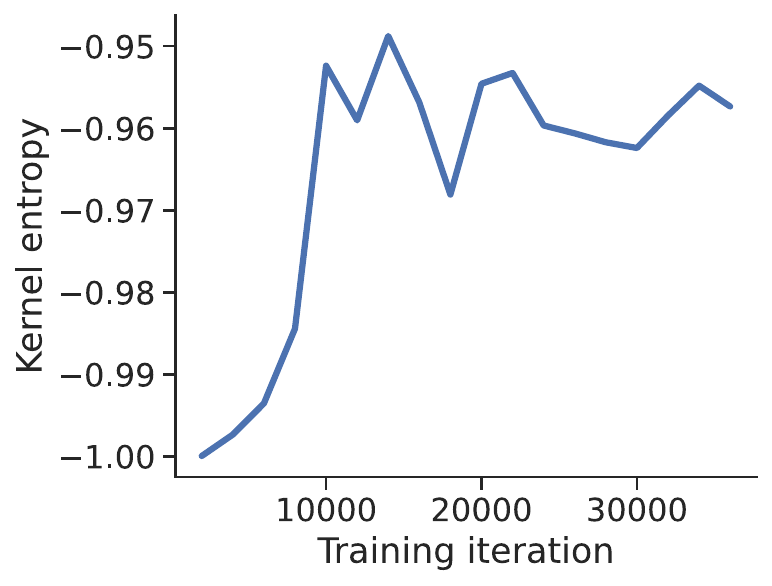}
    \end{subfigure}
    \begin{subfigure}{.33\textwidth}
    \centering
    \includegraphics[width=\columnwidth]{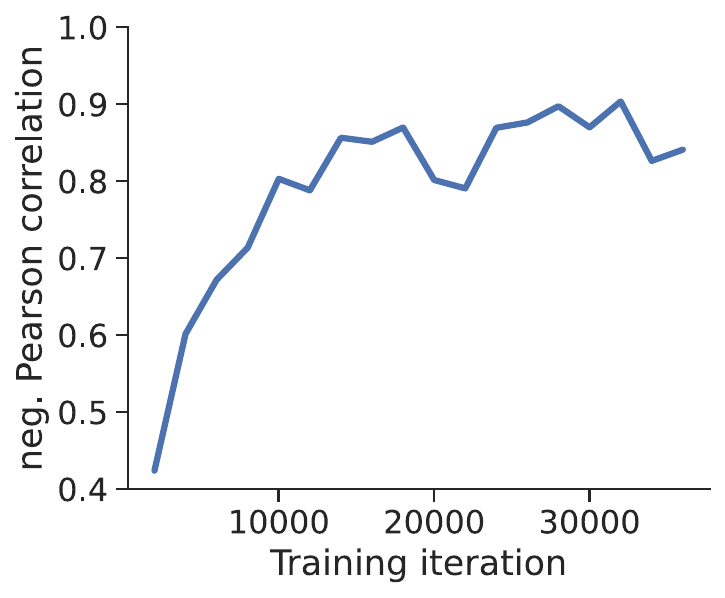}
    \end{subfigure}%
\caption{
Audio generation results for kernel score, kernel entropy, and negative Pearson correlation between them with the RBF kernel.
The trends are similar as in Figure \ref{fig:audio_per_epoch} but more erratic and the absolute correlation is slightly less.
}
\label{fig:tts_rbf}
\end{figure*}

Consequently, we expect the RBF kernel to behave more erratic than the Laplacian kernel in the main paper.
The results are shown in Figure \ref{fig:tts_rbf}.

\subsubsection{Natural Language Generation}

\begin{figure*}[t]
\vskip 0.2in
\centering
    \begin{subfigure}{.49\textwidth}
    \centering
    \includegraphics[width=\columnwidth]{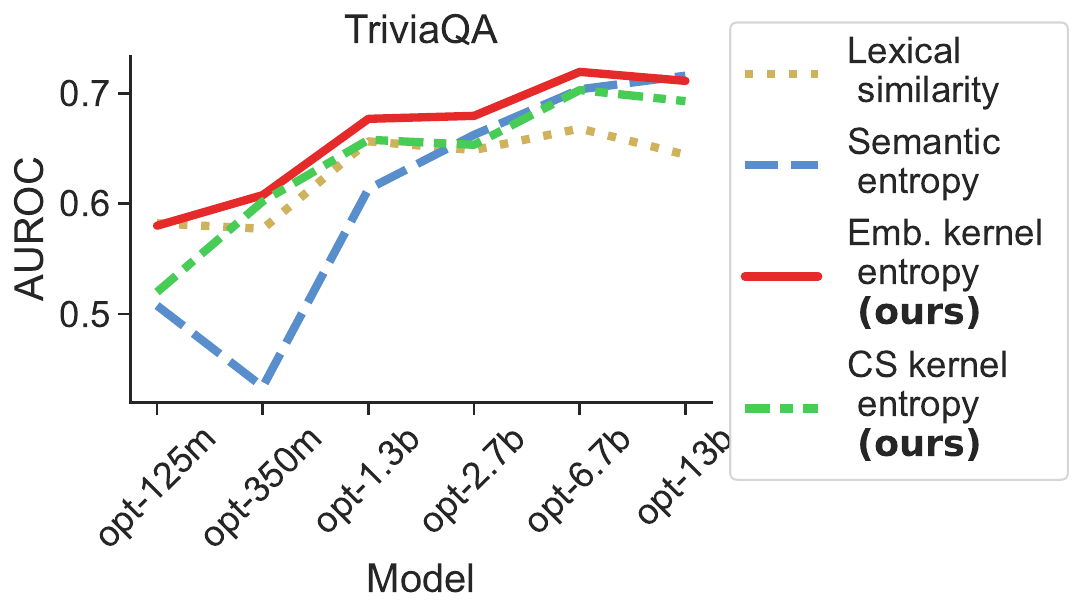}
    \end{subfigure}%
    \begin{subfigure}{.49\textwidth}
    \centering
    \includegraphics[width=\columnwidth]{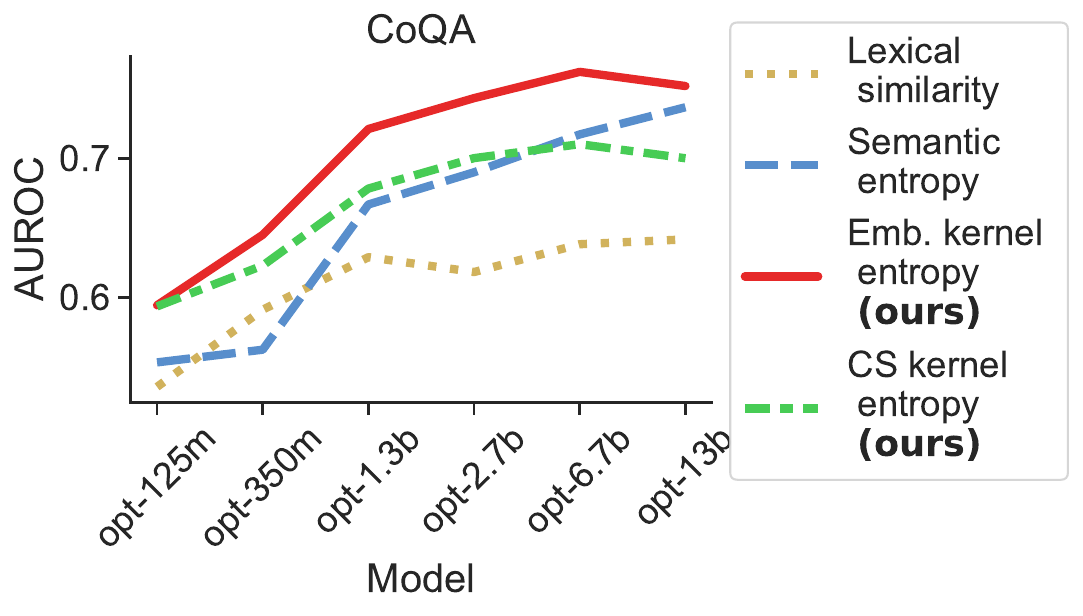}
    \end{subfigure}
\caption{
Area-Under-Curve (AUROC) of answer accuracy based on thresholds for best-performing uncertainty measures (embedding-based kernel entropy, semantic entropy) and model-free closed-source uncertainty measures (cs kernel entropy, lexical similarity).
Even when no embedding model is available, the kernel entropy is a very strong baseline in predicting the correctness of generated answers across a wide range of differently sized models.
In our evaluations, cs kernel entropy strongly outperforms lexical similarity.}
\label{fig:nlg_csk}
\vskip -0.2in
\end{figure*}

\begin{table}[t]
\caption{
AUROC between answer correctness and uncertainty estimates based on CS kernel entropy (ours) and lexical similarity. These are the only uncertainty measures in our experiments which do not require a pretrained semantic model and are applicable to any closed-source LLM.}
\label{sample-table}
\vskip 0.15in
\begin{center}
\begin{small}
\begin{sc}
\begin{tabular}{lcccr}
\toprule
Dataset & Model & CS Kernel Ent. & Lex. Sim. \\
\midrule
\multirow{6}{4em}{TriviaQA}
& Opt-125m & 0.520 & \textbf{0.582} \\
& Opt-350m & \textbf{0.602} & 0.577 \\
& Opt-1.3b & \textbf{0.657} & 0.656 \\
& Opt-2.7b & \textbf{0.653} & 0.648 \\
& Opt-6.7b & \textbf{0.702} & 0.668 \\
& Opt-13b & \textbf{0.692} & 0.644 \\
\hline
\multirow{6}{4em}{CoQA}
& Opt-125m & \textbf{0.593} & 0.535 \\
& Opt-350m & \textbf{0.623} & 0.591 \\
& Opt-1.3b & \textbf{0.678} & 0.629 \\
& Opt-2.7b & \textbf{0.700} & 0.618 \\
& Opt-6.7b & \textbf{0.710} & 0.638 \\
& Opt-13b & \textbf{0.700} & 0.641 \\
\bottomrule
\end{tabular}
\end{sc}
\end{small}
\end{center}
\vskip -0.1in
\label{tab:csk}
\end{table}

\begin{figure*}
\centering
    \begin{subfigure}{.5\textwidth}
    \centering
    \includegraphics[width=\columnwidth]{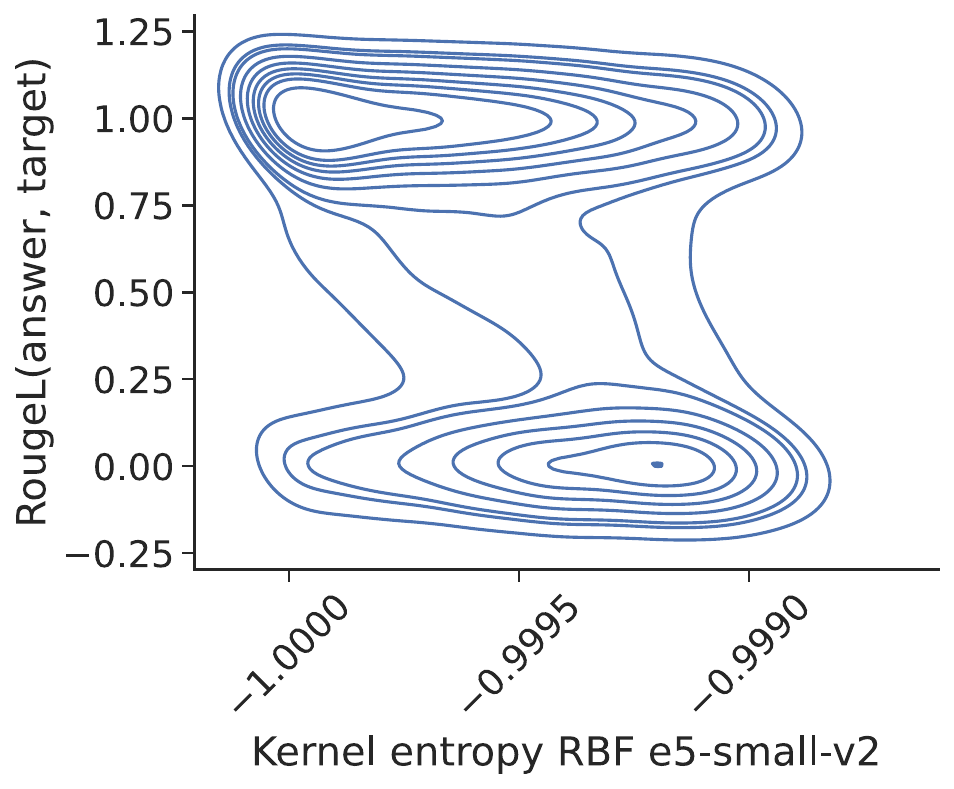}
    \end{subfigure}
\caption{
Kernel density estimation of kernel entropy and RougeL between answer and target for opt-1.3b model on CoQA.
A large RougeL corresponds to a high accuracy and a low error.
Consequently, low kernel entropy indicates a high likelihood of answer correctness.
}
\label{fig:nlg_kent_rougel}
\end{figure*}

Next, we continue with the natural language experiments. \\
In Figure \ref{fig:nlg_kent_rougel}, we confirm that the correlation between the predictive kernel entropy and the RougeL (which is supposed to be maximized) has the same sign as in the image experiments in Figure \ref{fig:scatter_plots}.

\begin{figure*}
\centering
    \begin{subfigure}{.5\textwidth}
    \centering
    \includegraphics[width=\columnwidth]{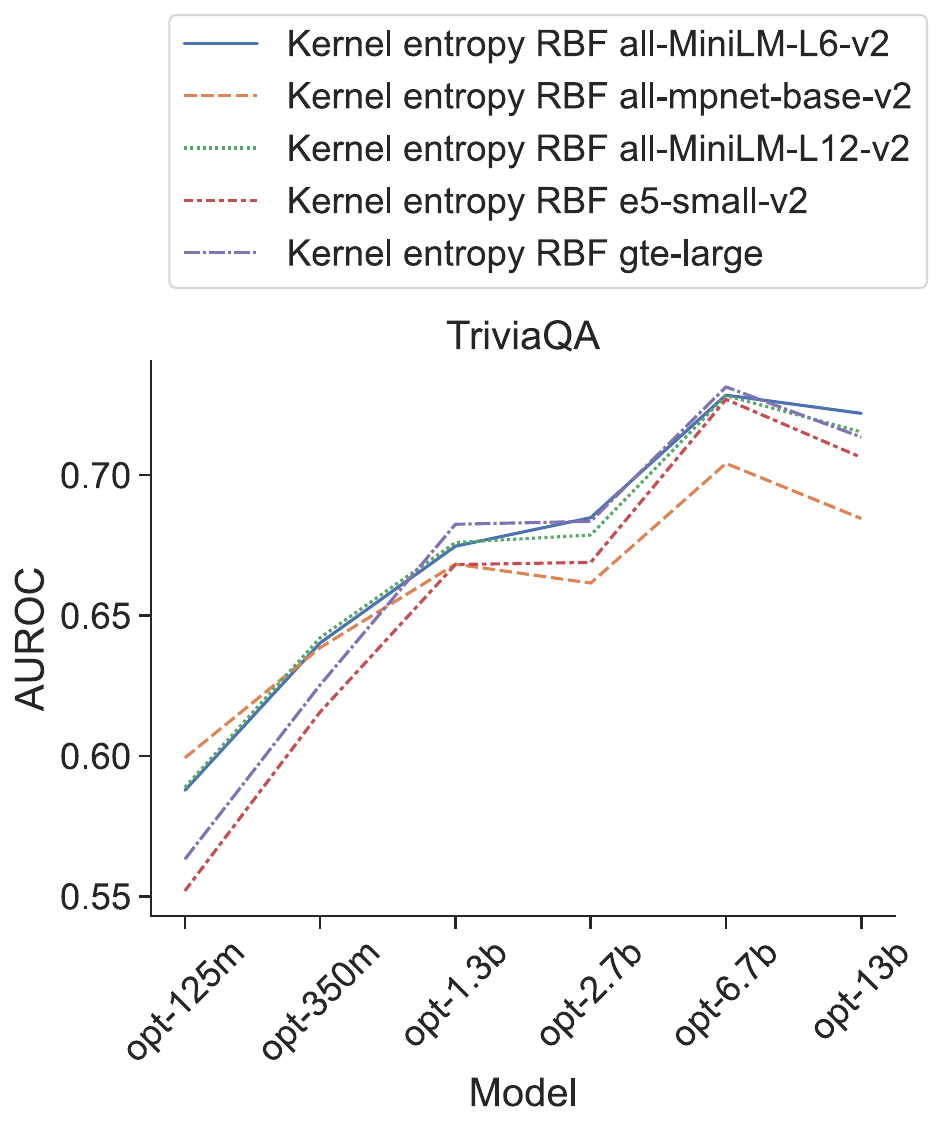}
    \end{subfigure}%
    \begin{subfigure}{.5\textwidth}
    \centering
    \includegraphics[width=\columnwidth]{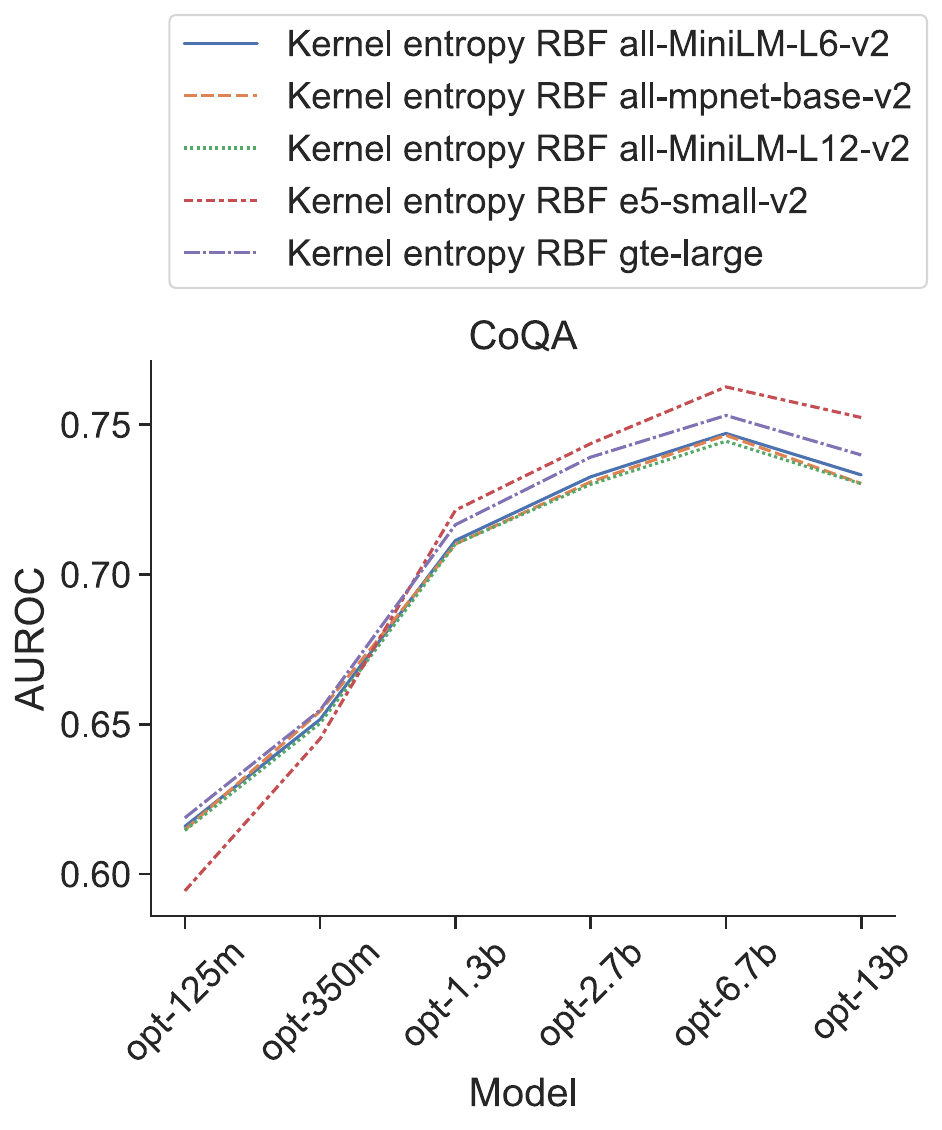}
    \end{subfigure}
\caption{
Comparing uncertainty estimates of kernel entropy for different embedders and RBF kernel.
In both cases, the differences are not substantial.
But, all-MiniLM-L6-v2 and all-MiniLM-L12-v2 perform comparably better on TriviaQA, likely due to them being trained on the TriviaQA training set.
}
\label{fig:nlg_embs_ext}
\end{figure*}

We also evaluate different embedders.
For this, we compare different ones, which have been pretrained on a variety of different training sets and with different embedding dimensions available on HuggingFace.
These are e5-small-v2 (384 dimensions; used in the main paper) \citep{wang2022text}, gte-large (1024 dimensions) \citep{li2023towards}, all-mpnet-base-v2 (768 dimensions) \citep{song2020mpnet}, as well as all-MiniLM-l6-v2 and all-MiniLM-L12-v2 (both 384 dimensions) \citep{wang2020minilm}.
The models all-mpnet-base-v2, all-MiniLM-l6-v2, and all-MiniLM-L12-v2 included the training set of TriviaQA in their training data, while e5-small-v2 and gte-large did not.
Consequently, these three models are an unfair comparison for TriviaQA to other baselines not using its training set.
In Figure \ref{fig:nlg_embs_ext}, we compare the ability of each embedder in combination with the RBF kernel to predict the answer accuracy for the same settings as in Figure \ref{fig:nlg_sota} in the main paper.
As we can see for CoQA, all embedders provide approximately similar performance.
This indicates that kernel entropy is a robust approach as long as the embedder is meaningful.
The results for TriviaQA are similar, but all-MiniLM-L6-v2 and all-MiniLM-L12-v2 perform comparably better.
We hypothesis that this is due to them being trained on the training set of TriviaQA.
This suggests that we can achieve even better uncertainty estimates by using a task specific training set.

\begin{figure*}
\centering
    \begin{subfigure}{.5\textwidth}
    \centering
    \includegraphics[width=\columnwidth]{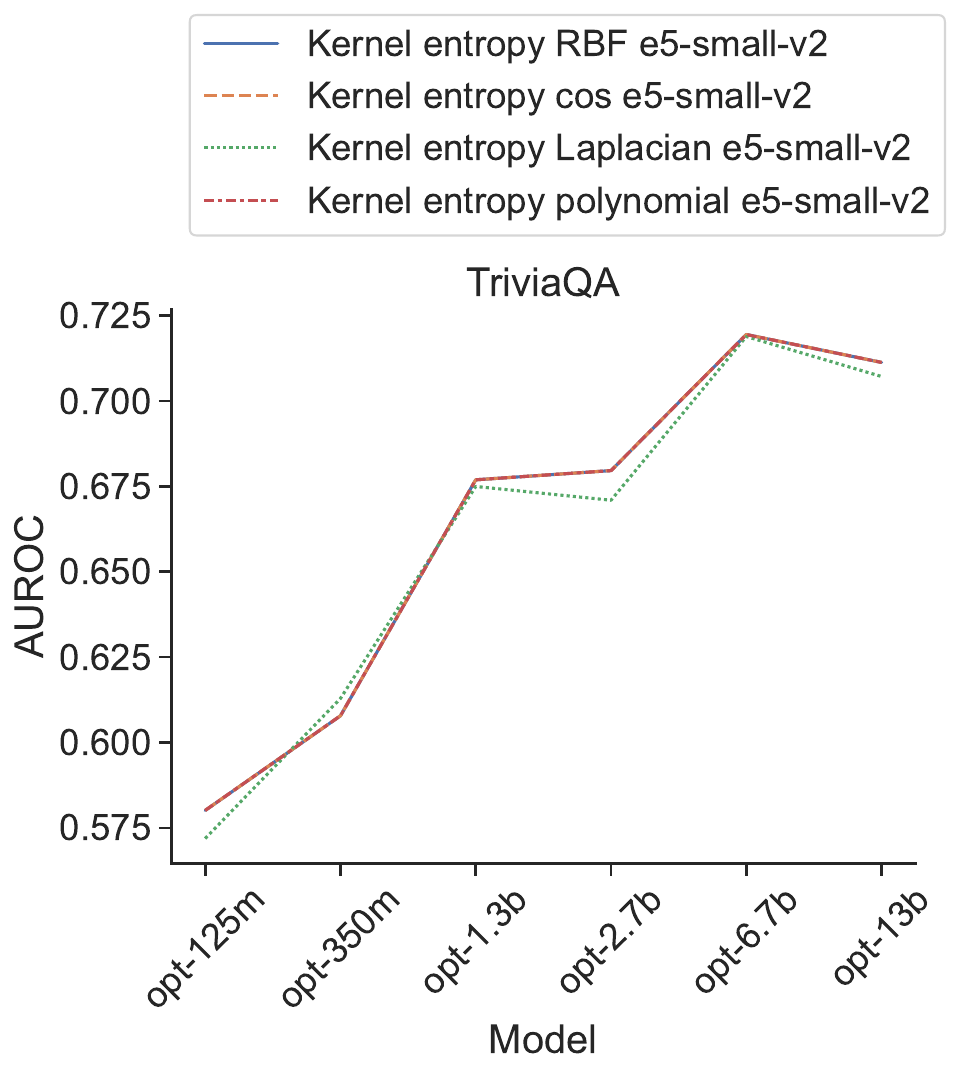}
    \end{subfigure}%
    \begin{subfigure}{.5\textwidth}
    \centering
    \includegraphics[width=\columnwidth]{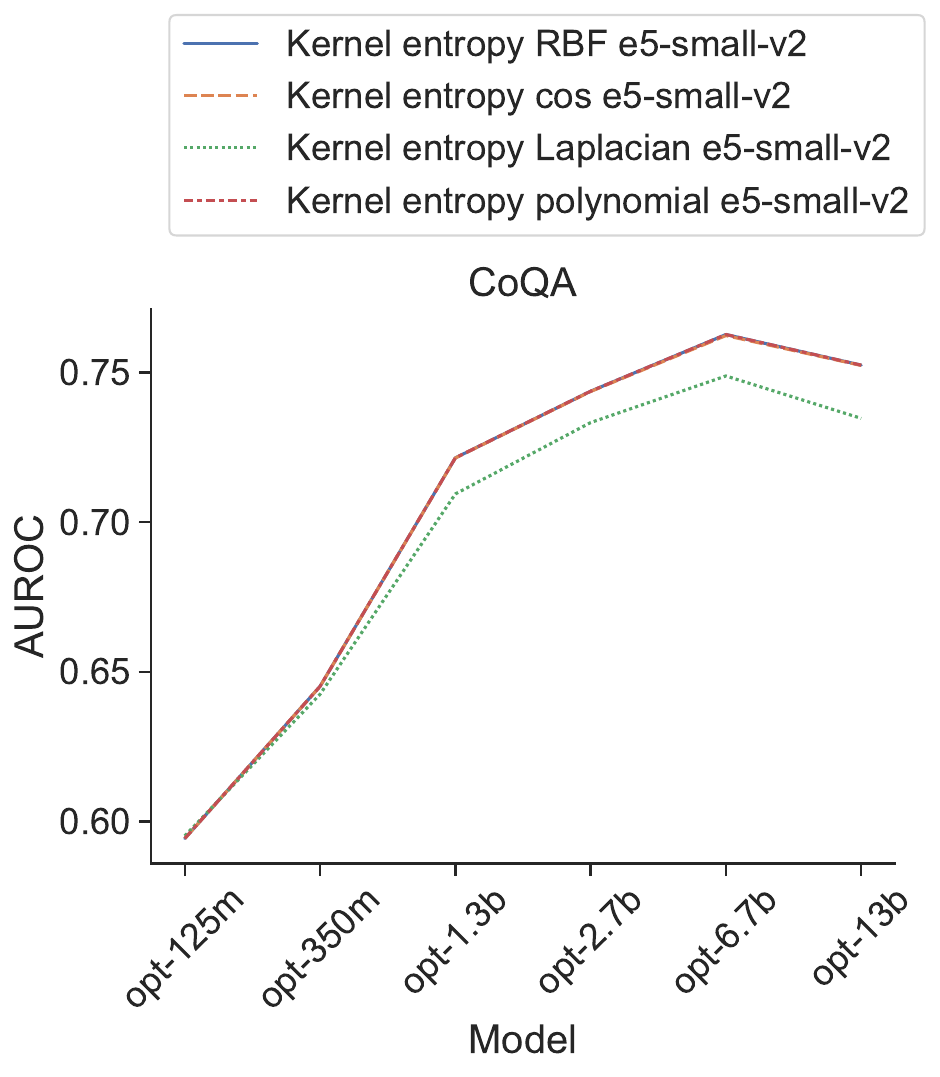}
    \end{subfigure}
\caption{
Comparing uncertainty estimates of kernel entropy for cosine similarity and RBF, polynomial, and Laplacian kernel.
Even though the embedder e5-small-v2 is trained via cosine similarity, the performance differences are marginal.
}
\label{fig:nlg_rbfvscos}
\end{figure*}

We also compare the impact of different kernel choices in Figure \ref{fig:nlg_rbfvscos}.
The differences are marginal for cosine similarity, and RBF and polynomial kernel.
Contrary, the Laplacian kernel performs often worse.
The lack of a difference between RBF or polynomial kernel and cosine similarity is surprising considering that e5-small-v2 has been trained using the cosine similarity \citep{wang2022text}.
Our results suggests that the embedder has substantially more influence than the kernel, and that resources should be spent on optimizing the former and not the latter.

\begin{figure*}
\centering
    \begin{subfigure}{.5\textwidth}
    \centering
    \includegraphics[width=\columnwidth]{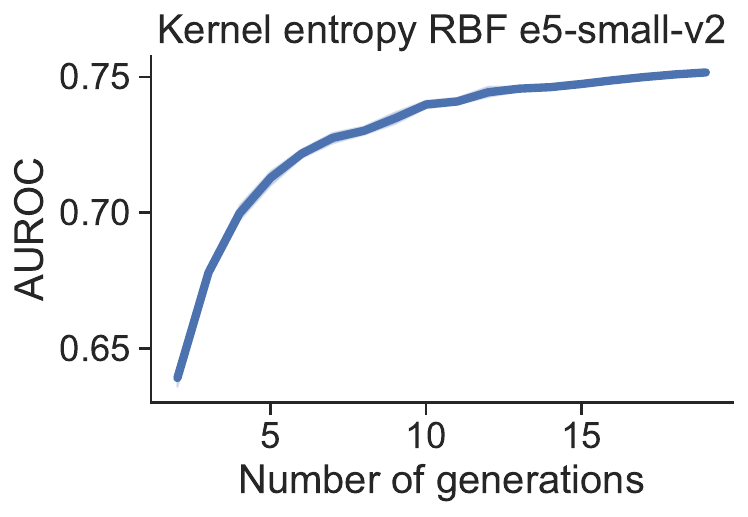}
    \end{subfigure}
\caption{
Number of generated answers to estimate kernel entropy compared to the AUROC for Opt-13b on CoQA.
More generations monotonically improve kernel entropy as an uncertainty estimate.
}
\label{fig:nlg_kent_samples}
\end{figure*}

Further, we compare different number of generated answers to estimate the kernel entropy in the case of Opt-13b on CoQA in Figure \ref{fig:nlg_kent_samples}.
As can be seen, more samples are continuously better.
Consequently, we expect to get even better results in Figure \ref{fig:nlg_sota} for larger numbers of generations.

Last, we also evaluate kernel entropy with a sequence kernel, which does not require an embedding model.
This makes kernel entropy as an uncertainty estimate applicable to scenarios, where no embedding model exists.
We choose the contiguous subsequence (cs) kernel defined by
\begin{equation}
    k_{\mathrm{CS}} \left( x, y \right) = \frac{k_t \left( x, y \right)}{\sqrt{k_t \left( x, x \right)} \sqrt{k_t \left( y, y \right)}}
\end{equation}
for a hyperparameter $t \in \mathbb{N}$ with
\begin{equation}
    k_t \left( x_{(1:l)}, y_{(1:l^\prime)} \right) = \sum_{i=1}^{l-t+1} \sum_{j=1}^{l^\prime-t+1} \mathbf{1}_{x_{(i:i+t-1)} = y_{(j:j+t-1)}}
\end{equation}
where $x_{(i:i+t-1)} = (x_i, x_{i+1}, \dots, x_{i+t-1})^\intercal$ and $\mathbf{1}_{x = y} = 1$ if $x=y$ else $0$ \citep{baum2023kernel}.
The kernel $k_t$ is p.s.d. (c.f. \citet{kiraly2019kernels}) and counts the number of contiguous subsequences of length $t$, which appear in $x$ and $y$.
The p.s.d. kernel $k_{\mathrm{CS}}$ is the normalised version bounded between 0 and 1.
Since our evaluations do not include a validation set, we pick $t=2$ based on \citep{baum2023kernel}.
In general, better AUROC performance may be achieved by optimizing kernel hyperparameters via a validation set. In Figure \ref{fig:nlg_csk} and Table \ref{tab:csk} we compare the AUROC performance of the cs kernel with other uncertainty estimates from the main paper.
As can be seen, the cs kernel entropy is a competitive baseline and strongly outperforms lexical similarity.
We specifically highlight the comparison with the lexical similarity, since it is the only other approach within our evaluations which is applicable to any setting with generated sequences.
This includes the closed-source setting and settings, where the model has no understanding of natural language (the baseline p(True) requires Q\&A queries, which strongly depend on the capabilities of the underlying model).

\newpage
$ $

\section{Missing Proofs}
\label{app:proofs}

In this section, we give all missing proofs for Theorem \ref{th:bvcd} and for the statements in Section \ref{sec:estimators}.
We first start with the proof for the bias-variance-covariance decomposition in Section \ref{app:bvcd}.
Then in Section \ref{app:var_estimator}, we solve the expectation and the variance of the distributional variance estimator proposed in Equation \eqref{eq:var_est}.
Last, we do the same for the distributional covariance estimator of Equation \eqref{eq:cov_est} in Section \ref{app:cov_estimator}.

\subsection{Bias-Variance-Covariance Decomposition}
\label{app:bvcd}

In Theorem \ref{th:bvcd}, the covariance decomposition is introduced after stating the more simpler bias-variance decomposition.
Here, we prove both in one go.
Assume we have target $Y \sim Q \in \mathcal{P}$ and ensemble prediction $\hat{P}^{(n)} \coloneqq \frac{1}{n} \sum_{i=1}^n \hat{P}_i$ of $n \in \mathbb{N}$ identically distributed predictions $\hat{P}_1, \dots, \hat{P}_n$ with outcomes in $\mathcal{P}$.
Further, assume that all expectations in the following are finite.
We will use $\hat{P} \coloneqq \hat{P}_1$ and $\hat{P}^\prime \coloneqq \hat{P}_2$.
Then, the decomposition can be constructed via

\begin{equation}
\begin{split}
    & \mathbb{E} \left[ - S_k \left( \hat{P}^{(n)}, Y \right) \right] \\
    & = \mathbb{E} \left[ \left\lVert \hat{P}^{(n)} \right\rVert_k^2 - 2 \Braket{ \hat{P}^{(n)} | k | \delta_Y} \right] \\
    \overset{\text{(i)}}&{=} \mathbb{E} \left[ \left\lVert \hat{P}^{(n)} \right\rVert_k^2 - 2 \Braket{ \hat{P}^{(n)} | k | Q} \right] \\
    & = \mathbb{E} \left[ \left\lVert \hat{P}^{(n)} \right\rVert_k^2 \right] - 2 \Braket{ \mathbb{E} \left[ \hat{P}^{(n)} \right] | k | Q} \\
    & = \mathbb{E} \left[ \left\lVert \hat{P}^{(n)} \right\rVert_k^2 \right] - 2 \Braket{ \mathbb{E} \left[ \hat{P}^{(n)} \right] | k | Q} + 2 \left\lVert \mathbb{E} \left[ \hat{P}^{(n)} \right] \right\rVert_k^2 - 2 \mathbb{E} \left[ \Braket{ \mathbb{E} \left[ \hat{P}^{(n)} \right] | k | \hat{P}^{(n)}} \right] \\
    & = \mathbb{E} \left[ \left\lVert \hat{P}^{(n)} - \mathbb{E} \left[ \hat{P}^{(n)} \right] \right\rVert_k^2 \right] - 2 \Braket{ \mathbb{E} \left[ \hat{P}^{(n)} \right] | k | Q} + \left\lVert \mathbb{E} \left[ \hat{P}^{(n)} \right] \right\rVert_k^2 \\
    & = \mathbb{E} \left[ \left\lVert \hat{P}^{(n)} - \mathbb{E} \left[ \hat{P}^{(n)} \right] \right\rVert_k^2 \right] - 2 \Braket{ \mathbb{E} \left[ \hat{P}^{(n)} \right] | k | Q} + \left\lVert \mathbb{E} \left[ \hat{P}^{(n)} \right] \right\rVert_k^2 + \left\lVert Q \right\rVert_k^2 - \left\lVert Q \right\rVert_k^2 \\
    & =  - \left\lVert Q \right\rVert_k^2 + \left\lVert \mathbb{E} \left[ \hat{P}^{(n)} \right] - Q \right\rVert_k^2 + \mathbb{E} \left[ \left\lVert \hat{P}^{(n)} - \mathbb{E} \left[ \hat{P}^{(n)} \right] \right\rVert_k^2 \right] \\
    \overset{\text{(ii)}}&{=} - \left\lVert Q \right\rVert_k^2 + \left\lVert \mathbb{E} \left[ \hat{P} \right] - Q \right\rVert_k^2 + \mathbb{E} \left[ \left\lVert \hat{P}^{(n)} - \mathbb{E} \left[ \hat{P} \right] \right\rVert_k^2 \right] \\
    & = - \left\lVert Q \right\rVert_k^2 + \left\lVert \mathbb{E} \left[ \hat{P} \right] - Q \right\rVert_k^2 + \mathbb{E} \left[ \left\lVert \frac{1}{n} \sum_{i=1}^n \hat{P}_i - \mathbb{E} \left[ \hat{P} \right] \right\rVert_k^2 \right] \\
    & = - \left\lVert Q \right\rVert_k^2 + \left\lVert \mathbb{E} \left[ \hat{P} \right] - Q \right\rVert_k^2 + \frac{1}{n^2} \sum_{i,j=1}^n \mathbb{E} \left[ \Braket{ \hat{P}_i - \mathbb{E} \left[ \hat{P} \right] | k | \hat{P}_j - \mathbb{E} \left[ \hat{P}  \right] } \right] \\
    & = - \left\lVert Q \right\rVert_k^2 + \left\lVert \mathbb{E} \left[ \hat{P} \right] - Q \right\rVert_k^2 + \frac{1}{n^2} \sum_{i,j=1}^n \operatorname{Cov}_k \left( \hat{P}_i, \hat{P}_j \right) \\
    & = - \left\lVert Q \right\rVert_k^2 + \frac{1}{n^2} \sum_{i=1}^n \operatorname{Var}_k \left( \hat{P}_i \right) + \frac{1}{n^2} \sum_{\substack{i,j=1 \\ i \neq j}}^n \operatorname{Cov}_k \left( \hat{P}_i, \hat{P}_j \right) + \left\lVert \mathbb{E} \left[ \hat{P} \right] - Q \right\rVert_k^2 \\
    \overset{\text{(ii)}}&{=} - \left\lVert Q \right\rVert_k^2 + \left\lVert \mathbb{E} \left[ \hat{P} \right] - Q \right\rVert_k^2 + \frac{1}{n^2} n \operatorname{Var}_k \left( \hat{P} \right) + \frac{1}{n^2} n \left(n-1\right) \operatorname{Cov}_k \left( \hat{P}, \hat{P}^\prime \right) \\
    & = \underbrace{- \left\lVert Q \right\rVert_k^2}_{\text{Noise}} + \underbrace{\left\lVert \mathbb{E} \left[ \hat{P} \right] - Q \right\rVert_k^2}_{\text{Bias}} + \underbrace{\frac{1}{n} \operatorname{Var}_k \left( \hat{P} \right)}_{\text{Variance}} + \underbrace{\frac{n-1}{n} \operatorname{Cov}_k \left( \hat{P}, \hat{P}^\prime \right)}_{\text{Covariance}} \\
\label{eq:bvcd_proof}
\end{split}
\end{equation}

(i) $Y$ and $\hat{P}^{(n)}$ independently distributed \\
(ii) $\hat{P}_1, \dots, \hat{P}_n$ identically distributed \\

\subsection{Distributional Variance Estimator}
\label{app:var_estimator}

In general, the notation of \citet{eaton1981method} allows us to write for random variables $P$ and $Q$ with outcomes in a distribution space, and for independent $X \sim P$ and $Y \sim Q$ that
\begin{equation}
    \mathbb{E} \left[ k \left( X, Y \right) \mid P, Q \right] = \int k \left( x, y \right) \mathrm{d} P \otimes Q \left( x, y \right) = \int \int k \left( x, y \right) \mathrm{d} P \left( x \right) \mathrm{d} Q \left( y \right) = \Braket{P | k | Q},
\end{equation}
where we used Tonelli's Theorem to split the integral.

Assume the variables are defined as in Section \ref{sec:estimators}.
As a reminder: $P, P_1, \dots, P_n$ i.i.d. and $X_{i1}, \dots, X_{im} \sim P_i$ i.i.d. for a given $i = 1, \dots, n$.
Note we have for $j \neq t$ that $X_{ij}$ and $X_{it}$ are independent given $P_i$ for all $i = 1, \dots, n$.
From this follows

\begin{equation}
\begin{split}
    \mathbb{E} \left[ k \left( X_{ij}, X_{it} \right) \right]
    & = \mathbb{E} \left[ \mathbb{E} \left[ k \left( X_{ij}, X_{it} \right) \mid P_i \right] \right] \\
    \overset{\text{iid}}&{=} \mathbb{E} \left[ \mathbb{E} \left[ k \left( X_{i1}, X_{i2} \right) \mid P_i \right] \right] \\
    \overset{\text{iid}}&{=} \mathbb{E} \left[ \Braket{ P_i | k | P_i } \right] \\
    \overset{\text{iid}}&{=} \mathbb{E} \left[ \Braket{ P | k | P } \right]
\label{eq:var_est1}
\end{split}
\end{equation}

as well as for $i \neq s$

\begin{equation}
\begin{split}
    \mathbb{E} \left[ k \left( X_{ij}, X_{st} \right) \right]
    & = \mathbb{E} \left[ \mathbb{E} \left[ k \left( X_{ij}, X_{st} \right) \mid P_i, P_s \right] \right] \\
    \overset{\text{iid}}&{=} \mathbb{E} \left[ \mathbb{E} \left[ k \left( X_{i1}, X_{s1} \right) \mid P_i, P_s \right] \right] \\
    \overset{\text{iid}}&{=} \mathbb{E} \left[ \Braket{ P_i | k | P_s } \right] \\
    \overset{\text{iid}}&{=} \Braket{ \mathbb{E} \left[ P \right] | k | \mathbb{E} \left[ P \right] }.
\label{eq:var_est2}
\end{split}
\end{equation}

\subsubsection{Expectation of the Estimator}

Using these equations gives

\begin{equation}
\begin{split}
    & \mathbb{E} \left[ \widehat{\operatorname{Var}}_k^{\left(n, m \right)} \right] \\
    & = \mathbb{E} \left[ \frac{1}{n m} \sum_{i=1}^n \sum_{j=1}^m \left( \frac{1}{m-1} \sum_{\substack{t=1 \\ t \neq j}}^m k \left( X_{ij}, X_{it} \right) - \frac{1}{\left(n - 1\right) m} \sum_{\substack{s=1 \\ s \neq i}}^n \sum_{t=1}^m k \left( X_{ij}, X_{st} \right) \right) \right] \\
    & = \frac{1}{n m} \sum_{i=1}^n \sum_{j=1}^m \left( \frac{1}{m-1} \sum_{\substack{t=1 \\ t \neq j}}^m \mathbb{E} \left[ k \left( X_{ij}, X_{it} \right) \right] - \frac{1}{\left(n - 1\right) m} \sum_{\substack{s=1 \\ s \neq i}}^n \sum_{t=1}^m \mathbb{E} \left[ k \left( X_{ij}, X_{st} \right) \right] \right) \\
    \overset{\text{Eq. \eqref{eq:var_est1} \& \eqref{eq:var_est2}}}&{=} \frac{1}{n m} \sum_{i=1}^n \sum_{j=1}^m \left( \frac{1}{m-1} \sum_{\substack{t=1 \\ t \neq j}}^m \mathbb{E} \left[ \Braket{ P | k | P } \right] - \frac{1}{\left(n - 1\right) m} \sum_{\substack{s=1 \\ s \neq i}}^n \sum_{t=1}^m \Braket{ \mathbb{E} \left[ P \right] | k | \mathbb{E} \left[ P \right] } \right) \\
    & = \mathbb{E} \left[ \Braket{ P | k | P } \right] - \Braket{ \mathbb{E} \left[ P \right] | k | \mathbb{E} \left[ P \right] } \\
    & = \mathbb{E} \left[ \Braket{ P - \mathbb{E} \left[ P \right] | k | P - \mathbb{E} \left[ P \right] } \right] \\
    & = \operatorname{Var}_k \left( P \right).
\end{split}
\end{equation}



\subsubsection{Variance of the Estimator}

Note that for a U-statistic $\hat{U}_n = \frac{1}{n \left(n-1\right)} \sum_{i=1}^n \sum_{\substack{j=1 \\ j \neq i}}^n h \left( X_i, X_j \right)$ based on i.i.d. samples $X_1, \dots, X_n$ and symmetric function $h$, the estimator variance is given by 
\begin{equation}
    \mathbb{V} \left( \hat{U}_n \right) = \frac{2}{n \left( n-1 \right)} \mathbb{V} \left( h \left(X_1, X_2 \right) \right) + \frac{4 \left( n-2 \right)}{n \left( n-1 \right)} \mathbb{V} \left( \mathbb{E} \left[ h \left( X_1, X_2 \right) \mid X_2 \right] \right)
\label{eq:var_ustat}
\end{equation}
\citep{shao2003mathematical}.

We will use the law of total variance (TV) several times to create independence between the summands, since for dependent random variables $X$ and $Y$ with scalars $a,b$ we have $\mathbb{V} \left(aX + bY \right) = a^2 \mathbb{V} \left( X \right) + b^2 \mathbb{V} \left( Y \right) + 2ab \operatorname{Cov} \left( X,Y \right)$ \citep{shao2003mathematical}.

From this also follows that

\begin{equation}
\begin{split}
    & \mathbb{V} \left( \widehat{\operatorname{Var}}_k^{\left(n, m \right)} \right) \\
    & = \mathbb{V} \left( \frac{1}{n m} \sum_{i=1}^n \sum_{j=1}^m \left( \frac{1}{m-1} \sum_{\substack{t=1 \\ t \neq j}}^m k \left( X_{ij}, X_{it} \right) - \frac{1}{\left(n - 1\right) m} \sum_{\substack{s=1 \\ s \neq i}}^n \sum_{t=1}^m k \left( X_{ij}, X_{st} \right) \right) \right) \\
    & = \mathbb{V} \left( \frac{1}{n m \left(m-1\right)} \sum_{i=1}^n \sum_{j=1}^m \sum_{\substack{t=1 \\ t \neq j}}^m k \left( X_{ij}, X_{it} \right) \right) + \mathbb{V} \left( \frac{1}{n \left(n - 1\right) m^2} \sum_{i=1}^n \sum_{j=1}^m \sum_{\substack{s=1 \\ s \neq i}}^n \sum_{t=1}^m k \left( X_{ij}, X_{st} \right) \right) \\
    & \quad \quad - 2 \operatorname{Cov} \left( \frac{1}{n m \left(m-1\right)} \sum_{i=1}^n \sum_{j=1}^m \sum_{\substack{t=1 \\ t \neq j}}^m k \left( X_{ij}, X_{it} \right),  \frac{1}{n \left(n - 1\right) m^2} \sum_{i=1}^n \sum_{j=1}^m \sum_{\substack{s=1 \\ s \neq i}}^n \sum_{t=1}^m k \left( X_{ij}, X_{st} \right) \right) \\
\label{eq:var_var6}
\end{split}
\end{equation}

We will analyse each term successively and then combine the results further down in \eqref{eq:var_var_result}.

\begin{equation}
\begin{split}
    & \mathbb{V} \left( \frac{1}{n m \left(m-1\right)} \sum_{i=1}^n \sum_{j=1}^m \sum_{\substack{t=1 \\ t \neq j}}^m k \left( X_{ij}, X_{it} \right) \right) \\
    \overset{\text{TV}}&{=} \mathbb{V} \left( \mathbb{E} \left[ \frac{1}{n m \left(m-1\right)} \sum_{i=1}^n \sum_{j=1}^m \sum_{\substack{t=1 \\ t \neq j}}^m k \left( X_{ij}, X_{it} \right) \mid P_1 \dots P_n \right] \right) \\
    & \quad \quad + \mathbb{E} \left[ \mathbb{V} \left( \frac{1}{n m \left(m-1\right)} \sum_{i=1}^n \sum_{j=1}^m \sum_{\substack{t=1 \\ t \neq j}}^m k \left( X_{ij}, X_{it} \right) \mid P_1 \dots P_n \right) \right] \\
    \overset{\text{iid}}&{=} \mathbb{V} \left( \frac{1}{n m \left(m-1\right)} \sum_{i=1}^n \sum_{j=1}^m \sum_{\substack{t=1 \\ t \neq j}}^m \mathbb{E} \left[ k \left( X_{ij}, X_{it} \right) \mid P_i \right] \right) \\
    & \quad \quad + \mathbb{E} \left[ \frac{1}{n^2} \sum_{i=1}^n \mathbb{V} \left( \frac{1}{ m \left(m-1\right)} \sum_{j=1}^m \sum_{\substack{t=1 \\ t \neq j}}^m k \left( X_{ij}, X_{it} \right) \mid P_i \right) \right] \\
    \overset{\text{iid}}&{=} \mathbb{V} \left( \frac{1}{n} \sum_{i=1}^n \mathbb{E} \left[ k \left( X_{i1}, X_{i2} \right) \mid P_i \right] \right) + \frac{1}{n} \mathbb{E} \left[ \mathbb{V} \left( \frac{1}{ m \left(m-1\right)} \sum_{j=1}^m \sum_{\substack{t=1 \\ t \neq j}}^m k \left( X_{1j}, X_{1t} \right) \mid P_1 \right) \right] \\
    \overset{\text{iid}}&{=} \frac{1}{n} \mathbb{V} \left( \Braket{P |k| P} \right) + \frac{1}{n} \mathbb{E} \left[ \mathbb{V} \left( \frac{1}{ m \left(m-1\right)} \sum_{j=1}^m \sum_{\substack{t=1 \\ t \neq j}}^m k \left( X_{1j}, X_{1t} \right) \mid P_1 \right) \right] \\
    \overset{\text{(i)}}&{=} \frac{1}{n} \mathbb{V} \left( \Braket{P |k| P} \right) + \frac{4 \left( m - 2 \right)}{nm \left(m - 1 \right)} \zeta_1 + \frac{2}{nm \left(m-1\right)} \zeta_2 \\
\label{eq:var_var1}
\end{split}
\end{equation}

(i) with $\zeta_1 = \mathbb{E} \left[ \mathbb{V} \left( \mathbb{E} \left[ k \left( X_{11}, X_{12} \right) \mid X_{12}, P_1 \right] \mid P_1 \right) \right]$ and $\zeta_2 = \mathbb{E} \left[ \mathbb{V} \left( k \left( X_{11}, X_{12} \right) \mid P_1 \right) \right]$ based on \eqref{eq:var_ustat}.

\begin{equation}
\begin{split}
    & \mathbb{V} \left( \frac{1}{n \left(n - 1\right) m^2} \sum_{i=1}^n \sum_{j=1}^m \sum_{\substack{s=1 \\ s \neq i}}^n \sum_{t=1}^m k \left( X_{ij}, X_{st} \right) \right) \\
    \overset{\text{TV}}&{=} \underbrace{\mathbb{V} \left( \mathbb{E} \left[ \frac{1}{n \left(n - 1\right) m^2} \sum_{i=1}^n \sum_{j=1}^m \sum_{\substack{s=1 \\ s \neq i}}^n \sum_{t=1}^m k \left( X_{ij}, X_{st} \right) \mid P_1 \dots P_n \right] \right)}_{(\mathrm{I}) \coloneqq} \\
    & \quad \quad + \underbrace{\mathbb{E} \left[ \mathbb{V} \left( \frac{1}{n \left(n - 1\right) m^2} \sum_{i=1}^n \sum_{j=1}^m \sum_{\substack{s=1 \\ s \neq i}}^n \sum_{t=1}^m k \left( X_{ij}, X_{st} \right) \mid P_1 \dots P_n \right) \right]}_{(\mathrm{II}) \coloneqq}. \\
\end{split}
\end{equation}

Due to the length of the expression, we first solve (I) and then (II).

\begin{equation}
\begin{split}
    (\mathrm{I}) \overset{\text{iid}}&{=} \mathbb{V} \left( \frac{1}{n \left(n - 1\right) m^2} \sum_{i=1}^n \sum_{j=1}^m \sum_{\substack{s=1 \\ s \neq i}}^n \sum_{t=1}^m \mathbb{E} \left[ k \left( X_{ij}, X_{st} \right) \mid P_i, P_s \right] \right) \\
    \overset{\text{iid}}&{=} \mathbb{V} \left( \frac{1}{n \left(n - 1\right)} \sum_{i=1}^n \sum_{\substack{s=1 \\ s \neq i}}^n \Braket{P_i | k | P_s} \right) \\
    \overset{\text{(i)}}&{=} \frac{4 \left(n-2\right)}{n \left(n-1\right)} \zeta_3 + \frac{2}{n \left(n-1\right)} \zeta_4 \\
\label{eq:var_var2}
\end{split}
\end{equation}

(i) with $\zeta_3 = \mathbb{V} \left( \mathbb{E} \left[ \Braket{P_1 | k | P_2} \mid P_1 \right] \right)$ and $\zeta_4 = \mathbb{V} \left( \Braket{P_1 | k | P_2} \right)$ based on \eqref{eq:var_ustat}.

For the next term note that $\frac{1}{m^2} \sum_{j=1}^m \sum_{t=1}^m k \left( X_{ij}, X_{st} \right)$ and $\frac{1}{m^2} \sum_{j=1}^m \sum_{t=1}^m k \left( X_{aj}, X_{bt} \right)$ are independent given $P_i, P_s, P_a, P_b$ for $i \neq s \neq a \neq b$ from which follows

\begin{equation}
    \operatorname{Cov} \left( \frac{1}{m^2} \sum_{j=1}^m \sum_{t=1}^m k \left( X_{ij}, X_{st} \right), \frac{1}{m^2} \sum_{c=1}^m \sum_{d=1}^m k \left( X_{ac}, X_{bd} \right) \right) = 0.
\end{equation}

Consequently, we have

\begin{equation}
\begin{split}
    (\mathrm{II}) \overset{\text{}}&{=} \mathbb{E} \left[ \frac{1}{n^2 \left(n - 1\right)^2} \sum_{i=1}^n \sum_{\substack{s=1 \\ s \neq i}}^n \mathbb{V} \left( \frac{1}{m^2} \sum_{j=1}^m \sum_{t=1}^m k \left( X_{ij}, X_{st} \right) \mid P_i, P_s \right) \right] \\
    & \quad \quad + \mathbb{E} \left[ \frac{1}{n^2 \left(n - 1\right)^2} \sum_{i=1}^n \sum_{\substack{s=1 \\ s \neq i}}^n \sum_{\substack{b=1 \\ b \neq i \\ b \neq s}}^n \operatorname{Cov} \left( \frac{1}{m^2} \sum_{j=1}^m \sum_{t=1}^m k \left( X_{ij}, X_{st} \right), \frac{1}{m^2} \sum_{c=1}^m \sum_{d=1}^m k \left( X_{ic}, X_{bd} \right) \mid P_i, P_s, P_b \right) \right] \\
    & \quad \quad + \mathbb{E} \left[ \frac{1}{n^2 \left(n - 1\right)^2} \sum_{i=1}^n \sum_{\substack{s=1 \\ s \neq i}}^n \sum_{\substack{a=1 \\ a \neq s \\ a \neq i}}^n \operatorname{Cov} \left( \frac{1}{m^2} \sum_{j=1}^m \sum_{t=1}^m k \left( X_{ij}, X_{st} \right), \frac{1}{m^2} \sum_{c=1}^m \sum_{d=1}^m k \left( X_{ac}, X_{sd} \right) \mid P_i, P_s, P_a \right) \right] \\
    \overset{\text{sym}}&{=} \underbrace{\mathbb{E} \left[ \frac{1}{n^2 \left(n - 1\right)^2} \sum_{i=1}^n \sum_{\substack{s=1 \\ s \neq i}}^n \mathbb{V} \left( \frac{1}{m^2} \sum_{c=1}^m \sum_{d=1}^m k \left( X_{ic}, X_{sd} \right) \mid P_i, P_s \right) \right]}_{(\mathrm{IIa}) \coloneqq} \\
    & \quad \quad + \underbrace{\mathbb{E} \left[ \frac{2}{n^2 \left(n - 1\right)^2} \sum_{i=1}^n \sum_{\substack{s=1 \\ s \neq i}}^n \sum_{\substack{b=1 \\ b \neq i \\ b \neq s}}^n \operatorname{Cov} \left( \frac{1}{m^2} \sum_{j=1}^m \sum_{t=1}^m k \left( X_{ij}, X_{st} \right), \frac{1}{m^2} \sum_{c=1}^m \sum_{d=1}^m k \left( X_{ic}, X_{bd} \right) \mid P_i, P_s, P_b \right) \right]}_{(\mathrm{IIb}) \coloneqq}. \\
\end{split}
\end{equation}

Due to the length of the expressions, we again first look at (IIa) and then (IIb).

\begin{equation}
\begin{split}
    (\mathrm{IIa}) \overset{\text{iid}}&{=} \mathbb{E} \left[ \frac{1}{n^2 \left(n - 1\right)^2} \sum_{i=1}^n \sum_{\substack{s=1 \\ s \neq i}}^n \mathbb{V} \left( \frac{1}{m^2} \sum_{j=1}^m \sum_{t=1}^m k \left( X_{ij}, X_{st} \right) \mid P_i, P_s \right) \right] \\
    \overset{\text{iid}}&{=} \frac{1}{n \left(n - 1\right)} \mathbb{E} \left[ \mathbb{V} \left( \frac{1}{m^2} \sum_{j=1}^m \sum_{t=1}^m k \left( X_{1j}, X_{2t} \right) \mid P_1, P_2 \right) \right] \\
    \overset{}&{=} \frac{1}{n \left(n - 1\right)} \mathbb{E} \left[ \operatorname{Cov} \left( \frac{1}{m^2} \sum_{j=1}^m \sum_{t=1}^m k \left( X_{1j}, X_{2t} \right), \frac{1}{m^2} \sum_{i=1}^m \sum_{s=1}^m k \left( X_{1i}, X_{2s} \right) \mid P_1, P_2 \right) \right] \\
    \overset{}&{=} \frac{1}{n \left(n - 1\right) m^4} \sum_{j=1}^m \sum_{t=1}^m \sum_{i=1}^m \sum_{s=1}^m \mathbb{E} \left[ \operatorname{Cov} \left( k \left( X_{1j}, X_{2t} \right), k \left( X_{1i}, X_{2s} \right) \mid P_1, P_2 \right) \right] \\
    \overset{\text{iid}}&{=} \frac{1}{n \left(n - 1\right) m^4} \sum_{j=1}^m \sum_{t=1}^m \mathbb{E} \left[ \operatorname{Cov} \left( k \left( X_{1j}, X_{2t} \right), k \left( X_{1j}, X_{2t} \right) \mid P_1, P_2 \right) \right] \\
    & \quad \quad + \frac{1}{n \left(n - 1\right) m^4} \sum_{j=1}^m \sum_{t=1}^m \sum_{\substack{i=1 \\ i \neq j}}^m \mathbb{E} \left[ \operatorname{Cov} \left( k \left( X_{1j}, X_{2t} \right), k \left( X_{1i}, X_{2t} \right) \mid P_1, P_2 \right) \right] \\
    & \quad \quad + \frac{1}{n \left(n - 1\right) m^4} \sum_{j=1}^m \sum_{t=1}^m \sum_{\substack{s=1 \\ s \neq t}}^m \mathbb{E} \left[ \operatorname{Cov} \left( k \left( X_{1j}, X_{2t} \right), k \left( X_{1j}, X_{2s} \right) \mid P_1, P_2 \right) \right] \\
    \overset{\text{iid}}&{=} \frac{1}{n \left(n - 1\right) m^2} \mathbb{E} \left[ \operatorname{Cov} \left( k \left( X_{11}, X_{21} \right), k \left( X_{11}, X_{21} \right) \mid P_1, P_2 \right) \right] \\
    & \quad \quad + \frac{m-1}{n \left(n - 1\right) m^2} \mathbb{E} \left[ \operatorname{Cov} \left( k \left( X_{11}, X_{21} \right), k \left( X_{12}, X_{21} \right) \mid P_1, P_2 \right) \right] \\
    & \quad \quad + \frac{m-1}{n \left(n - 1\right) m^2} \mathbb{E} \left[ \operatorname{Cov} \left( k \left( X_{11}, X_{21} \right), k \left( X_{11}, X_{22} \right) \mid P_1, P_2 \right) \right] \\
    \overset{\text{(i)}}&{=} \frac{1}{n \left(n - 1\right) m^2} \underbrace{\mathbb{E} \left[ \mathbb{V} \left( k \left( X_{11}, X_{21} \right) \mid P_1, P_2 \right) \right]}_{\zeta_6 \coloneqq} \\
    & \quad \quad + \frac{2 \left(m-1\right)}{n \left(n - 1\right) m^2} \underbrace{\mathbb{E} \left[ \operatorname{Cov} \left( k \left( X_{11}, X_{21} \right), k \left( X_{12}, X_{21} \right) \mid P_1, P_2 \right) \right]}_{\zeta_5 \coloneqq} \\
\label{eq:var_var3}
\end{split}
\end{equation}

(i) follows from symmetry of $k$ and assumption of identical distributions.

Further, we have $\operatorname{Cov} \left( k \left( X_{ij}, X_{st} \right), k \left( X_{ic}, X_{bd} \right) \mid P_i, P_s, P_b \right) = 0$ whenever $c \neq j$ (since $s \neq b$ and independence assumption), giving

\begin{equation}
\begin{split}
    (\mathrm{IIb}) & = \mathbb{E} \left[ \frac{2}{n^2 \left(n - 1\right)^2} \sum_{i=1}^n \sum_{\substack{s=1 \\ s \neq i}}^n \sum_{\substack{b=1 \\ b \neq i \\ b \neq s}}^n \operatorname{Cov} \left( \frac{1}{m^2} \sum_{j=1}^m \sum_{t=1}^m k \left( X_{ij}, X_{st} \right), \frac{1}{m^2} \sum_{c=1}^m \sum_{d=1}^m k \left( X_{ic}, X_{bd} \right) \mid P_i, P_s, P_b \right) \right] \\
    & = \mathbb{E} \left[ \frac{2}{n^2 \left(n - 1\right)^2 m^4} \sum_{i=1}^n \sum_{\substack{s=1 \\ s \neq i}}^n \sum_{\substack{b=1 \\ b \neq i \\ b \neq s}}^n \sum_{j=1}^m \sum_{t=1}^m \sum_{d=1}^m \operatorname{Cov} \left( k \left( X_{ij}, X_{st} \right), k \left( X_{ij}, X_{bd} \right) \mid P_i, P_s, P_b \right) \right] \\
    \overset{\text{iid}}&{=} \mathbb{E} \left[ \frac{2}{n^2 \left(n - 1\right)^2 m^4} \sum_{i=1}^n \sum_{\substack{s=1 \\ s \neq i}}^n \sum_{\substack{b=1 \\ b \neq i \\ b \neq s}}^n \sum_{j=1}^m \sum_{t=1}^m \sum_{d=1}^m \operatorname{Cov} \left( k \left( X_{11}, X_{21} \right), k \left( X_{11}, X_{31} \right) \mid P_1, P_2, P_3 \right) \right] \\
    \overset{\text{}}&{=} \frac{2 \left( n-2 \right)}{n \left(n - 1\right) m} \underbrace{\mathbb{E} \left[ \operatorname{Cov} \left( k \left( X_{11}, X_{21} \right), k \left( X_{11}, X_{31} \right) \mid P_1, P_2, P_3 \right) \right]}_{\zeta_9 \coloneqq} \\
\label{eq:var_var4}
\end{split}
\end{equation}

The only term left is
\begin{equation}
\begin{split}
    & \operatorname{Cov} \left( \frac{1}{n m \left(m-1\right)} \sum_{i=1}^n \sum_{j=1}^m \sum_{\substack{t=1 \\ t \neq j}}^m k \left( X_{ij}, X_{it} \right), \frac{1}{n \left(n - 1\right) m^2} \sum_{i=1}^n \sum_{j=1}^m \sum_{\substack{s=1 \\ s \neq i}}^n \sum_{t=1}^m k \left( X_{ij}, X_{st} \right) \right) \\
    & = \frac{1}{n^2 \left(n - 1\right) m^3 \left(m-1\right)} \sum_{i=1}^n \sum_{j=1}^m \sum_{\substack{t=1 \\ t \neq j}}^m \sum_{o=1}^n \sum_{p=1}^m \sum_{\substack{s=1 \\ s \neq o}}^n \sum_{r=1}^m \operatorname{Cov} \left( k \left( X_{ij}, X_{it} \right), k \left( X_{op}, X_{sr} \right) \right) \\
    \overset{\text{(i)}}&{=} \frac{1}{n^2 \left(n - 1\right) m^3 \left(m-1\right)} \sum_{i=1}^n \sum_{j=1}^m \sum_{\substack{t=1 \\ t \neq j}}^m \sum_{p=1}^m \sum_{\substack{s=1 \\ s \neq i}}^n \sum_{r=1}^m \\
    & \quad \quad \left( \operatorname{Cov} \left( k \left( X_{ij}, X_{it} \right), k \left( X_{ip}, X_{sr} \right) \right) + \operatorname{Cov} \left( k \left( X_{ij}, X_{it} \right), k \left( X_{sp}, X_{ir} \right) \right) \right) \\
    \overset{\text{(ii)}}&{=} \frac{2}{n^2 \left(n - 1\right) m^3 \left(m-1\right)} \sum_{i=1}^n \sum_{j=1}^m \sum_{\substack{t=1 \\ t \neq j}}^m \sum_{p=1}^m \sum_{\substack{s=1 \\ s \neq i}}^n \sum_{r=1}^m \operatorname{Cov} \left( k \left( X_{ij}, X_{it} \right), k \left( X_{ip}, X_{sr} \right) \right) \\
    \overset{\text{iid}}&{=} \frac{2m n \left(n-1 \right)}{n^2 \left(n - 1\right) m^3 \left(m-1\right)} \sum_{j=1}^m \sum_{\substack{t=1 \\ t \neq j}}^m \sum_{p=1}^m \operatorname{Cov} \left( k \left( X_{1j}, X_{1t} \right), k \left( X_{1p}, X_{21} \right) \right) \\
    & = \frac{2}{n m^2 \left(m-1\right)} \left( \sum_{j=1}^m \sum_{\substack{t=1 \\ t \neq j}}^m \sum_{\substack{p \in \{j,t\}}} \operatorname{Cov} \left( k \left( X_{1j}, X_{1t} \right), k \left( X_{1p}, X_{21} \right) \right) + \sum_{j=1}^m \sum_{\substack{t=1 \\ t \neq j}}^m \sum_{\substack{p=1 \\ p \neq j \\ p \neq t}}^m \operatorname{Cov} \left( k \left( X_{1j}, X_{1t} \right), k \left( X_{1p}, X_{21} \right) \right) \right) \\
    \overset{\text{(iii)}}&{=} \frac{2}{n m^2 \left(m-1\right)} \left( \underbrace{\sum_{j=1}^m \sum_{\substack{t=1 \\ t \neq j}}^m \sum_{\substack{p \in \{j,t\}}} \operatorname{Cov} \left( k \left( X_{11}, X_{12} \right), k \left( X_{11}, X_{21} \right) \right)}_{2 m \left(m-1\right) \text{ summands}} + \underbrace{\sum_{j=1}^m \sum_{\substack{t=1 \\ t \neq j}}^m \sum_{\substack{p=1 \\ p \neq j \\ p \neq t}}^m \operatorname{Cov} \left( k \left( X_{11}, X_{12} \right), k \left( X_{13}, X_{21} \right) \right)}_{m \left(m-1\right)\left(m-2\right) \text{ summands}} \right) \\
    & = \frac{4 m \left(m-1\right)}{n m^2 \left(m-1\right)} \operatorname{Cov} \left( k \left( X_{11}, X_{12} \right), k \left( X_{11}, X_{21} \right) \right) + \frac{2m \left(m-1\right)\left(m-2\right)}{n m^2 \left(m-1\right)} \operatorname{Cov} \left( k \left( X_{11}, X_{12} \right), k \left( X_{13}, X_{21} \right) \right) \\
    & = \frac{4}{n m} \underbrace{\operatorname{Cov} \left( k \left( X_{11}, X_{12} \right), k \left( X_{11}, X_{21} \right) \right)}_{\zeta_7 \coloneqq} + \frac{2 \left(m-2\right)}{n m} \underbrace{\operatorname{Cov} \left( k \left( X_{11}, X_{12} \right), k \left( X_{13}, X_{21} \right) \right)}_{\zeta_8 \coloneqq}. \\
\label{eq:var_var5}
\end{split}
\end{equation}

(i) when $i \neq o \neq s$, then $\operatorname{Cov} \left( k \left( X_{ij}, X_{it} \right), k \left( X_{op}, X_{sr} \right) \right) = 0$. \\
(ii) symmetry of $k$ and iid property result in $\operatorname{Cov} \left( k \left( X_{ij}, X_{it} \right), k \left( X_{ip}, X_{sr} \right) \right) = \operatorname{Cov} \left( k \left( X_{ij}, X_{it} \right), k \left( X_{sp}, X_{ir} \right) \right)$ as long as $i \neq s$ \\
(iii) iid and symmetry of $k$.

It follows from inserting \eqref{eq:var_var1}, \eqref{eq:var_var2}, \eqref{eq:var_var4}, and \eqref{eq:var_var5}, into \eqref{eq:var_var6} that

\begin{equation}
\begin{split}
    \mathbb{V} \left( \widehat{\operatorname{Var}}_k^{\left(n, m \right)} \right) & = \underbrace{\frac{1}{n} \mathbb{V} \left( \Braket{P |k| P} \right) + \frac{4 \left(n-2\right)}{n \left(n-1\right)} \zeta_3 - \frac{4 \left(m-2\right)}{n m} \zeta_8}_{\mathcal{O} \left( \frac{1}{n} \right)} + \underbrace{\frac{2}{n \left(n-1\right)} \zeta_4}_{\mathcal{O} \left( \frac{1}{n^2} \right)} \\
    & \quad \quad + \underbrace{\frac{4 \left( m - 2 \right)}{nm \left(m - 1 \right)} \zeta_1 - \frac{8}{n m} 
 \zeta_7 + \frac{2 \left( n-2 \right)}{n \left(n - 1\right) m} \zeta_9}_{\mathcal{O} \left( \frac{1}{nm} \right)} \\
    & \quad \quad + \underbrace{\frac{2}{nm \left(m-1\right)} \zeta_2}_{\mathcal{O} \left( \frac{1}{nm^2} \right)} + \underbrace{\frac{2 \left(m-1\right)}{n \left(n - 1\right) m^2} \zeta_5}_{\mathcal{O} \left( \frac{1}{n^2 m} \right)} + \underbrace{\frac{1}{n \left(n - 1\right) m^2} \zeta_6}_{\mathcal{O} \left( \frac{1}{n^2 m^2} \right)}. \\
\label{eq:var_var_result}
\end{split}
\end{equation}

In summary, our estimator is in $\mathcal{O} \left( \frac{1}{n} \left( 1 + \frac{1}{m} \right) \right)$ and consequently consistent w.r.t. $n$ but not $m$.
However, in Equation \eqref{eq:var_var_result} appear several terms linearly, or even quadratically, shrinking with $m$.
Consequently, it depends on the given setting how important $m$ is, which is also demonstrated in Figure \ref{fig:var_sim} on the right.
There, we used our estimator with the RBF kernel for InfiMNIST generations as described in Appendix \ref{app:experiments:img}.

\subsubsection{Illustrations of the Estimator}
\label{app:var_illus}

The estimator can also be visualized in different ways as the following.
We can interpret it as an operator on clusters in Figure \ref{fig:var_sim}.
Alternatively, we can also understand it in the context of Gram-like block matrices like:
For $i,s \in \left\{1, \dots, n \right\}$ define the quadratic matrices $\mathbf{K}_{is} = \left( k_{is_{jt}} \right)_{j,t=1 \dots m} \in \mathbb{R}^{m \times m}$ with entries $k_{is_{jt}} = k \left( X_{ij}, X_{st} \right)$.
Then we have the colored block matrix
\begin{equation}
\begin{pmatrix}
\textcolor{cyan}{\mathbf{K}_{11}} & \textcolor{black}{\cdots} & \textcolor{black}{\cdots} & \textcolor{black}{\mathbf{K}_{n1}} \\
\textcolor{black}{\vdots} & \textcolor{cyan}{\ddots} & \textcolor{black}{\mathbf{K}_{is}} & \textcolor{black}{\vdots} \\
\textcolor{black}{\vdots} & \textcolor{black}{\mathbf{K}_{si}} & \textcolor{cyan}{\ddots} & \textcolor{black}{\vdots} \\
\textcolor{black}{\mathbf{K}_{1n}} & \textcolor{black}{\cdots} & \textcolor{black}{\cdots} & \textcolor{cyan}{\mathbf{K}_{nn}} \\
\end{pmatrix}
=
\left(\begin{array}{@{}c|c@{}}
  \begin{matrix}
    \textcolor{red}{k_{11_{11}}} & \textcolor{cyan}{\cdots} & \textcolor{cyan}{\cdots} & \textcolor{cyan}{k_{11_{1m}}} \\
    \textcolor{cyan}{\vdots} & \textcolor{red}{\ddots}  &  \textcolor{cyan}{k_{11_{jt}}} & \textcolor{cyan}{\vdots}  \\
    \textcolor{cyan}{\vdots} & \textcolor{cyan}{k_{11_{tj}}} & \textcolor{red}{\ddots} & \textcolor{cyan}{\vdots}  \\
    \textcolor{cyan}{k_{11_{m1}}} & \textcolor{cyan}{\cdots} & \textcolor{cyan}{\cdots} & \textcolor{red}{k_{11_{mm}}} \\
  \end{matrix}
  & 
  \begin{matrix}
    \textcolor{black}{k_{12_{11}}}  &  \textcolor{black}{\cdots} \\
    \textcolor{black}{\vdots}  &  \textcolor{black}{\ddots}  \\
    \textcolor{black}{\vdots}  &  \textcolor{black}{\ddots}  \\
    \textcolor{black}{k_{12_{m1}}} &  \textcolor{black}{\cdots} \\
  \end{matrix} \\[5.0ex]
  \hline \\[-2.0ex]
  \begin{matrix}
    \textcolor{black}{k_{21_{11}}}  &  \;\;\;\textcolor{black}{\cdots}\;\;\;  &  \;\;\;\textcolor{black}{\cdots}\;\;\;  &  \textcolor{black}{k_{21_{1m}}} \\
    \textcolor{black}{\vdots}  &  \textcolor{black}{\ddots} & \textcolor{black}{\ddots} & \textcolor{black}{\vdots}  \\
  \end{matrix}
  &
  \begin{matrix}
    \textcolor{red}{k_{22_{11}}} & \textcolor{cyan}{\cdots}  \\
    \textcolor{cyan}{\vdots} & \textcolor{red}{\ddots} \\
  \end{matrix}
\end{array}\right).
\end{equation}

The proposed distributional variance estimator is then the average of all cyan entries without the red ones (blocks on diagonal without diagonal entries) minus the average of all black entries (off-diagonal blocks).

\subsection{Distributional Covariance Estimator}
\label{app:cov_estimator}

Figure \ref{fig:cov_ill} shows an illustration of the distributional covariance estimator for two distribution (=cluster) samples of joint $(P, Q)$ and two per-distribution samples.

\begin{figure*}[t]
\centering
    \begin{subfigure}{.4\textwidth}
    \centering
    \resizebox{\columnwidth}{!}{%
    \begin{tikzpicture}
    \begin{axis}[
    ticks=none,
      axis x line=center,
      axis y line=center,
      xmin=-0.3,
      xmax=3.3,
      ymin=-0.3,
      ymax=1.7]
    \draw[orange, line width=0.6mm] (0.5,0.5) -- (1.0,0.2);
    \draw[orange, line width=0.6mm] (0.5,0.5) -- (0.9,1.2);
    \draw[blue, line width=0.6mm, densely dotted] (0.5,0.5) -- (2.2,0.3);
    \draw[blue, line width=0.6mm, densely dotted] (0.5,0.5) -- (2.3,1.3);

    \draw[orange, line width=0.6mm] (0.5,0.9) -- (1.0,0.2);
    \draw[orange, line width=0.6mm] (0.5,0.9) -- (0.9,1.2);
    \draw[blue, line width=0.6mm, densely dotted] (0.5,0.9) -- (2.2,0.3);
    \draw[blue, line width=0.6mm, densely dotted] (0.5,0.9) -- (2.3,1.3);

    \draw[blue, line width=0.6mm, densely dotted] (2.7,0.9) -- (1.0,0.2);
    \draw[blue, line width=0.6mm, densely dotted] (2.7,0.9) -- (0.9,1.2);
    \draw[orange, line width=0.6mm] (2.7,0.9) -- (2.2,0.3);
    \draw[orange, line width=0.6mm] (2.7,0.9) -- (2.3,1.3);

    \draw[blue, line width=0.6mm, densely dotted] (1.9,0.8) -- (1.0,0.2);
    \draw[blue, line width=0.6mm, densely dotted] (1.9,0.8) -- (0.9,1.2);
    \draw[orange, line width=0.6mm] (1.9,0.8) -- (2.2,0.3);
    \draw[orange, line width=0.6mm] (1.9,0.8) -- (2.3,1.3);

    \filldraw[black] (0.5,0.5) circle (3.5pt) node[anchor=west] {};
    \filldraw[black] (0.5,0.9) circle (3.5pt) node[anchor=west] {};
    \node[fill=black,regular polygon, regular polygon sides=3,inner sep=2pt] at (1.0,0.2) {};
    \node[fill=black,regular polygon, regular polygon sides=3,inner sep=2pt] at (0.9,1.2) {};

    \filldraw[black] (1.9,0.8) circle (3.5pt) node[anchor=west] {};
    \filldraw[black] (2.7,0.9) circle (3.5pt) node[anchor=west] {};
    \node[fill=black,regular polygon, regular polygon sides=3,inner sep=2pt] at (2.2,0.3) {};
    \node[fill=black,regular polygon, regular polygon sides=3,inner sep=2pt] at (2.3,1.3) {};
    \node[] at (0.2,0.5) {\large $X_{11}$};
    \node[] at (0.2,0.9) {\large $X_{12}$};
    \node[] at (2.18,0.85) {\large $X_{21}$};
    \node[] at (3.0,0.9) {\large $X_{22}$};
    \node[] at (0.7,0.2) {\large $Y_{11}$};
    \node[] at (0.6,1.2) {\large $Y_{12}$};
    \node[] at (2.6,1.3) {\large $Y_{21}$};
    \node[] at (2.4,0.15) {\large $Y_{22}$};
    \node[] at (2.7,1.61) {\Large $\mathcal{X}$};
\end{axis}
    \end{tikzpicture}
    }
    \end{subfigure}%
\caption{
    Illustration of the estimator $\widehat{\operatorname{Cov}}_k^{\left(n,m\right)}$ in the sample space $\mathcal{X}$ for $n=2$ outer samples and $m=2$ inner samples. The estimator computes the average similarity within clusters (solid orange lines) minus the average similarity between clusters (dotted blue lines). Shorter lines indicate higher similarity and larger kernel values.
}
\label{fig:cov_ill}
\end{figure*}

Note that we have under the given i.i.d. assumptions that

\begin{equation}
\begin{split}
    \mathbb{E} \left[ k \left( X_{ij}, Y_{it} \right) \right]
    & = \mathbb{E} \left[ \mathbb{E} \left[ k \left( X_{ij}, Y_{it} \right) \mid P_i, Q_i \right] \right] \\
    \overset{\text{iid}}&{=} \mathbb{E} \left[ \mathbb{E} \left[ k \left( X_{i1}, Y_{i1} \right) \mid P_i, Q_i \right] \right] \\
    \overset{\text{iid}}&{=} \mathbb{E} \left[ \Braket{ P_i | k | Q_i } \right] \\
    \overset{\text{iid}}&{=} \mathbb{E} \left[ \Braket{ P | k | Q } \right]
\label{eq:cov_est1}
\end{split}
\end{equation}

as well as for $i \neq s$

\begin{equation}
\begin{split}
    \mathbb{E} \left[ k \left( X_{ij}, Y_{st} \right) \right]
    & = \mathbb{E} \left[ \mathbb{E} \left[ k \left( X_{ij}, Y_{st} \right) \mid P_i, Q_s \right] \right] \\
    \overset{\text{iid}}&{=} \mathbb{E} \left[ \mathbb{E} \left[ k \left( X_{i1}, Y_{s1} \right) \mid P_i, Q_s \right] \right] \\
    \overset{\text{iid}}&{=} \mathbb{E} \left[ \Braket{ P_i | k | Q_s } \right] \\
    \overset{\text{iid}}&{=} \Braket{ \mathbb{E} \left[ P \right] | k | \mathbb{E} \left[ Q \right] }.
\label{eq:cov_est2}
\end{split}
\end{equation}

\subsubsection{Expectation of the Estimator}

Now, we can prove that the covariance estimator is unbiased, i.e.

\begin{equation}
\begin{split}
    & \mathbb{E} \left[ \widehat{\operatorname{Cov}}_k^{\left(n, m \right)} \left( \mathbf{X}, \mathbf{Y} \right) \right] \\
    & = \mathbb{E} \left[ \frac{1}{n m^2} \sum_{i=1}^n \sum_{j=1}^m \sum_{t=1}^m \left( k \left( X_{ij}, Y_{it} \right) -  \frac{1}{n - 1} \sum_{\substack{s=1 \\ s \neq i}}^n k \left( X_{ij}, Y_{st} \right) \right) \right] \\
    & = \frac{1}{n m^2} \sum_{i=1}^n \sum_{j=1}^m \sum_{t=1}^m \left( \mathbb{E} \left[ k \left( X_{ij}, Y_{it} \right) \right] - \frac{1}{n - 1} \sum_{\substack{s=1 \\ s \neq i}}^n \mathbb{E} \left[ k \left( X_{ij}, Y_{st} \right) \right] \right) \\
    \overset{\text{Eq. \eqref{eq:cov_est1} \& \eqref{eq:cov_est2}}}&{=}  \frac{1}{n m^2} \sum_{i=1}^n \sum_{j=1}^m \sum_{t=1}^m \left( \mathbb{E} \left[ \Braket{ P | k | Q } \right] - \frac{1}{n - 1} \sum_{\substack{s=1 \\ s \neq i}}^n \Braket{ \mathbb{E} \left[ P \right] | k | \mathbb{E} \left[ Q \right] } \right) \\
    & = \mathbb{E} \left[ \Braket{ P | k | Q } \right] - \Braket{ \mathbb{E} \left[ P \right] | k | \mathbb{E} \left[ Q \right] } \\
    & = \mathbb{E} \left[ \Braket{ P - \mathbb{E} \left[ P \right] | k | Q - \mathbb{E} \left[ Q \right] } \right] \\
    & = \operatorname{Cov}_k \left( P, Q \right).
\end{split}
\end{equation}

\subsubsection{Variance of the Estimator}

Similar to the variance case, we also analyse its convergence rate:

\begin{equation}
\begin{split}
    & \mathbb{V} \left( \widehat{\operatorname{Cov}}_k^{\left(n, m \right)} \left( \mathbf{X}, \mathbf{Y} \right) \right) \\
    & = \mathbb{V} \left( \frac{1}{n m^2} \sum_{i=1}^n \sum_{j=1}^m \sum_{t=1}^m \left( k \left( X_{ij}, Y_{it} \right) -  \frac{1}{n - 1} \sum_{\substack{s=1 \\ s \neq i}}^n k \left( X_{ij}, Y_{st} \right) \right) \right) \\
    & = \mathbb{V} \left( \frac{1}{n m^2} \sum_{i=1}^n \sum_{j=1}^m \sum_{t=1}^m k \left( X_{ij}, Y_{it} \right) \right) + \mathbb{V} \left( \frac{1}{n \left( n-1 \right) m^2} \sum_{i=1}^n \sum_{j=1}^m \sum_{t=1}^m \sum_{\substack{s=1 \\ s \neq i}}^n k \left( X_{ij}, Y_{st} \right) \right) \\
    & \quad \quad - 2 \operatorname{Cov} \left( \frac{1}{n m^2} \sum_{i=1}^n \sum_{j=1}^m \sum_{t=1}^m k \left( X_{ij}, Y_{it} \right),  \frac{1}{n \left( n-1 \right) m^2} \sum_{i=1}^n \sum_{j=1}^m \sum_{t=1}^m \sum_{\substack{s=1 \\ s \neq i}}^n k \left( X_{ij}, Y_{st} \right) \right) \\
\label{eq:var_cov0}
\end{split}
\end{equation}

\begin{equation}
\begin{split}
    & \mathbb{V} \left( \frac{1}{n m^2} \sum_{i=1}^n \sum_{j=1}^m \sum_{t=1}^m k \left( X_{ij}, Y_{it} \right) \right) \\
    \overset{\text{TV}}&{=} \mathbb{V} \left( \mathbb{E} \left[ \frac{1}{n m^2} \sum_{i=1}^n \sum_{j=1}^m \sum_{\substack{t=1}}^m k \left( X_{ij}, Y_{it} \right) \mid P_1 \dots P_n, Q_1 \dots Q_n \right] \right) \\
    & \quad \quad + \mathbb{E} \left[ \mathbb{V} \left( \frac{1}{n m^2} \sum_{i=1}^n \sum_{j=1}^m \sum_{\substack{t=1}}^m k \left( X_{ij}, Y_{it} \right) \mid P_1 \dots P_n, Q_1 \dots Q_n \right) \right] \\
    \overset{\text{iid}}&{=} \mathbb{V} \left( \frac{1}{n m^2} \sum_{i=1}^n \sum_{j=1}^m \sum_{\substack{t=1}}^m \mathbb{E} \left[ k \left( X_{ij}, Y_{it} \right) \mid P_i, Q_i \right] \right) \\
    & \quad \quad + \mathbb{E} \left[ \frac{1}{n^2} \sum_{i=1}^n \mathbb{V} \left( \frac{1}{ m^2} \sum_{j=1}^m \sum_{\substack{t=1}}^m k \left( X_{ij}, Y_{it} \right) \mid P_i, Q_i \right) \right] \\
    \overset{\text{iid}}&{=} \mathbb{V} \left( \frac{1}{n} \sum_{i=1}^n \mathbb{E} \left[ k \left( X_{i1}, Y_{i1} \right) \mid P_i, Q_i \right] \right) + \frac{1}{n} \mathbb{E} \left[ \mathbb{V} \left( \frac{1}{ m^2} \sum_{j=1}^m \sum_{\substack{t=1}}^m k \left( X_{1j}, Y_{1t} \right) \mid P_1, Q_1 \right) \right] \\
    \overset{\text{iid}}&{=} \frac{1}{n} \mathbb{V} \left[ \Braket{P |k| Q} \right] + \frac{1}{n} \mathbb{E} \left[ \mathbb{V} \left( \frac{1}{ m^2} \sum_{j=1}^m \sum_{\substack{t=1}}^m k \left( X_{1j}, Y_{1t} \right) \mid P_1, Q_1 \right) \right] \\
    \overset{\text{iid}}&{=} \frac{1}{n} \underbrace{\mathbb{V} \left( \Braket{P |k| Q} \right)}_{\eta_1 \coloneqq} + 
    \frac{1}{n m^2} \underbrace{\mathbb{E} \left[ \mathbb{V} \left( k \left( X_{11}, Y_{11} \right) \mid P_1, Q_1 \right) \right]}_{\eta_2 \coloneqq}
\label{eq:var_cov1}
\end{split}
\end{equation}

and for the second term we have

\begin{equation}
\begin{split}
    & \mathbb{V} \left( \frac{1}{n \left( n-1 \right) m^2} \sum_{i=1}^n \sum_{j=1}^m \sum_{t=1}^m \sum_{\substack{s=1 \\ s \neq i}}^n k \left( X_{ij}, Y_{st} \right) \right) \\
    \overset{\text{TV}}&{=} \underbrace{\mathbb{V} \left( \mathbb{E} \left[ \frac{1}{n \left(n - 1\right) m^2} \sum_{i=1}^n \sum_{j=1}^m \sum_{\substack{s=1 \\ s \neq i}}^n \sum_{t=1}^m k \left( X_{ij}, Y_{st} \right) \mid P_1 \dots P_n, Q_1 \dots Q_n \right] \right)}_{(\mathrm{III}) \coloneqq} \\
    & \quad \quad + \underbrace{\mathbb{E} \left[ \mathbb{V} \left( \frac{1}{n \left(n - 1\right) m^2} \sum_{i=1}^n \sum_{j=1}^m \sum_{\substack{s=1 \\ s \neq i}}^n \sum_{t=1}^m k \left( X_{ij}, Y_{st} \right) \mid P_1 \dots P_n, Q_1 \dots Q_n \right) \right]}_{(\mathrm{IV}) \coloneqq}. \\
\end{split}
\end{equation}

Due to the length of the expression, we first solve (III) and then (IV).

\begin{equation}
\begin{split}
    (\mathrm{III}) \overset{\text{iid}}&{=} \mathbb{V} \left( \frac{1}{n \left(n - 1\right) m^2} \sum_{i=1}^n \sum_{j=1}^m \sum_{\substack{s=1 \\ s \neq i}}^n \sum_{t=1}^m \mathbb{E} \left[ k \left( X_{ij}, Y_{st} \right) \mid P_i, Q_s \right] \right) \\
    \overset{\text{iid}}&{=} \mathbb{V} \left( \frac{1}{n \left(n - 1\right) m^2} \sum_{i=1}^n \sum_{j=1}^m \sum_{\substack{s=1 \\ s \neq i}}^n \sum_{t=1}^m \mathbb{E} \left[ k \left( X_{i1}, Y_{s1} \right) \mid P_i, Q_s \right] \right) \\
    & = \mathbb{V} \left( \frac{1}{n \left(n - 1\right)} \sum_{i=1}^n \sum_{\substack{s=1 \\ s \neq i}}^n \Braket{P_i | k | Q_s} \right) \\
    & = \mathbb{E} \left[ \left( \frac{1}{n \left(n - 1\right)} \sum_{i=1}^n \sum_{\substack{s=1 \\ s \neq i}}^n \Braket{P_i | k | Q_s} \right)^2 \right] - \left( \mathbb{E} \left[ \frac{1}{n \left(n - 1\right)} \sum_{i=1}^n \sum_{\substack{s=1 \\ s \neq i}}^n \Braket{P_i | k | Q_s} \right] \right)^2 \\
    \overset{\text{iid}}&{=} \frac{1}{n^2 \left(n - 1\right)^2} \sum_{i=1}^n \sum_{\substack{s=1 \\ s \neq i}}^n \sum_{j=1}^n \sum_{\substack{t=1 \\ t \neq j}}^n \mathbb{E} \left[ \Braket{P_i | k | Q_s} \Braket{P_j | k | Q_t} \right] - \Braket{\mathbb{E} \left[ P \right] | k | \mathbb{E} \left[ Q \right]}^2 \\
    & = \frac{1}{n^2 \left(n - 1\right)^2} \sum_{i=1}^n \sum_{\substack{s=1 \\ s \neq i}}^n \mathbb{E} \left[ \Braket{P_i | k | Q_s}^2 \right]  + \frac{1}{n^2 \left(n - 1\right)^2} \sum_{i=1}^n \sum_{\substack{s=1 \\ s \neq i}}^n \sum_{\substack{t=1 \\ t \neq j \\ t \neq s}}^n \mathbb{E} \left[ \Braket{P_i | k | Q_s} \Braket{P_i | k | Q_t} \right] \\
    & \quad \quad + \frac{1}{n^2 \left(n - 1\right)^2} \sum_{i=1}^n \sum_{\substack{s=1 \\ s \neq i}}^n \sum_{\substack{j=1 \\ j \neq i \\ j \neq s}}^n \mathbb{E} \left[ \Braket{P_i | k | Q_s} \Braket{P_j | k | Q_s} \right] \\
    & \quad \quad + \frac{1}{n^2 \left(n - 1\right)^2} \sum_{i=1}^n \sum_{\substack{s=1 \\ s \neq i}}^n \sum_{\substack{j=1 \\ j \neq i \\ j \neq s}}^n \sum_{\substack{t=1 \\ t \neq i \\ t \neq s \\ t \neq j}}^n \mathbb{E} \left[ \Braket{P_i | k | Q_s} \Braket{P_j | k | Q_t} \right] - \Braket{\mathbb{E} \left[ P \right] | k | \mathbb{E} \left[ Q \right]}^2 \\
    \overset{\text{iid}}&{=} \frac{1}{n \left(n - 1\right)} \mathbb{E} \left[ \Braket{P_1 | k | Q_2}^2 \right] + \frac{n - 2}{n \left(n - 1\right)} \mathbb{E} \left[ \Braket{P_1 | k | Q_2} \Braket{P_1 | k | Q_3} \right] \\
    & \quad \quad + \frac{n - 2}{n \left(n - 1\right)} \mathbb{E} \left[ \Braket{P_2 | k | Q_1} \Braket{P_3 | k | Q_1} \right] + \frac{\left(n - 2\right) \left(n - 3\right)}{n \left(n - 1\right)} \Braket{\mathbb{E} \left[ P \right] | k | \mathbb{E} \left[ Q \right]}^2 - \Braket{\mathbb{E} \left[ P \right] | k | \mathbb{E} \left[ Q \right]}^2 \\
    & = \frac{1}{n \left(n - 1\right)} \mathbb{E} \left[ \Braket{P_1 | k | Q_2}^2 \right] + \frac{n - 2}{n \left(n - 1\right)} \mathbb{E} \left[ \Braket{P_1 | k | Q_2} \Braket{P_1 | k | Q_3} \right] \\
    & \quad \quad + \frac{n - 2}{n \left(n - 1\right)} \mathbb{E} \left[ \Braket{P_2 | k | Q_1} \Braket{P_3 | k | Q_1} \right] + \frac{\left(n - 2\right) \left(n - 3\right) - n \left(n - 1\right)}{n \left(n - 1\right)} \Braket{\mathbb{E} \left[ P \right] | k | \mathbb{E} \left[ Q \right]}^2 \\
    & = \frac{1}{n \left(n - 1\right)} \mathbb{E} \left[ \Braket{P_1 | k | Q_2}^2 \right] + \frac{n - 2}{n \left(n - 1\right)} \mathbb{E} \left[ \Braket{P_1 | k | Q_2} \Braket{P_1 | k | Q_3} \right] \\
    & \quad \quad + \frac{n - 2}{n \left(n - 1\right)} \mathbb{E} \left[ \Braket{P_2 | k | Q_1} \Braket{P_3 | k | Q_1} \right] + \frac{\left(n - 2\right) \left(n - 3\right) - n \left(n - 1\right)}{n \left(n - 1\right)} \Braket{\mathbb{E} \left[ P \right] | k | \mathbb{E} \left[ Q \right]}^2 \\
    & = \frac{1}{n \left(n - 1\right)} \mathbb{E} \left[ \Braket{P_1 | k | Q_2}^2 \right] + \frac{n - 2}{n \left(n - 1\right)} \mathbb{E} \left[ \Braket{P_1 | k | Q_2} \Braket{P_1 | k | Q_3} \right] \\
    & \quad \quad + \frac{n - 2}{n \left(n - 1\right)} \mathbb{E} \left[ \Braket{P_2 | k | Q_1} \Braket{P_3 | k | Q_1} \right] - \frac{4n - 6}{n \left(n - 1\right)} \Braket{\mathbb{E} \left[ P \right] | k | \mathbb{E} \left[ Q \right]}^2 \\
    \overset{\text{(i)}}&{=} \frac{1}{n \left(n - 1\right)} \eta_3 +  \frac{n - 2}{n \left(n - 1\right)} \eta_4
\label{eq:var_cov2}
\end{split}
\end{equation}

(i) with $\eta_3 \coloneqq \mathbb{E} \left[ \Braket{P_1 | k | Q_2}^2 \right] - 2 \Braket{\mathbb{E} \left[ P \right] | k | \mathbb{E} \left[ Q \right]}^2$ and $\eta_4 \coloneqq \mathbb{E} \left[ \Braket{P_1 | k | Q_2} \Braket{P_1 | k | Q_3} \right] + \mathbb{E} \left[ \Braket{P_2 | k | Q_1} \Braket{P_3 | k | Q_1} \right] - 4 \Braket{\mathbb{E} \left[ P \right] | k | \mathbb{E} \left[ Q \right]}^2$.

Due to analogous reasons as in the distributional variance case, we have

\begin{equation}
\begin{split}
    (\mathrm{IV}) \overset{\text{}}&{=} \underbrace{\mathbb{E} \left[ \frac{1}{n^2 \left(n - 1\right)^2} \sum_{i=1}^n \sum_{\substack{s=1 \\ s \neq i}}^n \mathbb{V} \left( \frac{1}{m^2} \sum_{j=1}^m \sum_{t=1}^m k \left( X_{ij}, Y_{st} \right) \mid P_i, Q_s \right) \right]}_{(\mathrm{IVa}) \coloneqq} \\
    & \quad \quad + \underbrace{\mathbb{E} \left[ \frac{1}{n^2 \left(n - 1\right)^2} \sum_{i=1}^n \sum_{\substack{s=1 \\ s \neq i}}^n \sum_{\substack{b=1 \\ b \neq i \\ b \neq s}}^n \operatorname{Cov} \left( \frac{1}{m^2} \sum_{j=1}^m \sum_{t=1}^m k \left( X_{ij}, Y_{st} \right), \frac{1}{m^2} \sum_{c=1}^m \sum_{d=1}^m k \left( X_{ic}, Y_{bd} \right) \mid P_i, Q_b, Q_s \right) \right]}_{(\mathrm{IVb}) \coloneqq} \\
    & \quad \quad + \underbrace{\mathbb{E} \left[ \frac{1}{n^2 \left(n - 1\right)^2} \sum_{i=1}^n \sum_{\substack{s=1 \\ s \neq i}}^n \sum_{\substack{a=1 \\ a \neq s \\ a \neq i}}^n \operatorname{Cov} \left( \frac{1}{m^2} \sum_{j=1}^m \sum_{t=1}^m k \left( X_{ij}, Y_{st} \right), \frac{1}{m^2} \sum_{c=1}^m \sum_{d=1}^m k \left( X_{ac}, Y_{sd} \right) \mid P_i, P_a, Q_s \right) \right]}_{(\mathrm{IVc}) \coloneqq}. \\
\end{split}
\end{equation}

Due to the length of the expressions, we again first look at (IVa), then (IVb), and then (IVc).

In the following equation, we use almost the identical steps as in Equation \eqref{eq:var_var3}:

\begin{equation}
\begin{split}
    (\mathrm{IVa}) & = \mathbb{E} \left[ \frac{1}{n^2 \left(n - 1\right)^2} \sum_{i=1}^n \sum_{\substack{s=1 \\ s \neq i}}^n \mathbb{V} \left( \frac{1}{m^2} \sum_{j=1}^m \sum_{t=1}^m k \left( X_{ij}, Y_{st} \right) \mid P_i, Q_s \right) \right] \\
    \overset{\text{iid}}&{=} \frac{1}{n \left(n - 1\right)} \mathbb{E} \left[ \mathbb{V} \left( \frac{1}{m^2} \sum_{j=1}^m \sum_{t=1}^m k \left( X_{1j}, Y_{2t} \right) \mid P_1, Q_2 \right) \right] \\
    \overset{}&{=} \frac{1}{n \left(n - 1\right)} \mathbb{E} \left[ \operatorname{Cov} \left( \frac{1}{m^2} \sum_{j=1}^m \sum_{t=1}^m k \left( X_{1j}, Y_{2t} \right), \frac{1}{m^2} \sum_{i=1}^m \sum_{s=1}^m k \left( X_{1i}, Y_{2s} \right) \mid P_1, Q_2 \right) \right] \\
    \overset{}&{=} \frac{1}{n \left(n - 1\right) m^4} \sum_{j=1}^m \sum_{t=1}^m \sum_{i=1}^m \sum_{s=1}^m \mathbb{E} \left[ \operatorname{Cov} \left( k \left( X_{1j}, Y_{2t} \right), k \left( X_{1i}, Y_{2s} \right) \mid P_1, Q_2 \right) \right] \\
    \overset{\text{iid}}&{=} \frac{1}{n \left(n - 1\right) m^4} \sum_{j=1}^m \sum_{t=1}^m \mathbb{E} \left[ \operatorname{Cov} \left( k \left( X_{1j}, Y_{2t} \right), k \left( X_{1j}, Y_{2t} \right) \mid P_1, Q_2 \right) \right] \\
    & \quad \quad + \frac{1}{n \left(n - 1\right) m^4} \sum_{j=1}^m \sum_{t=1}^m \sum_{\substack{i=1 \\ i \neq j}}^m \mathbb{E} \left[ \operatorname{Cov} \left( k \left( X_{1j}, Y_{2t} \right), k \left( X_{1i}, Y_{2t} \right) \mid P_1, Q_2 \right) \right] \\
    & \quad \quad + \frac{1}{n \left(n - 1\right) m^4} \sum_{j=1}^m \sum_{t=1}^m \sum_{\substack{s=1 \\ s \neq t}}^m \mathbb{E} \left[ \operatorname{Cov} \left( k \left( X_{1j}, Y_{2t} \right), k \left( X_{1j}, Y_{2s} \right) \mid P_1, Q_2 \right) \right] \\
    \overset{\text{iid}}&{=} \frac{1}{n \left(n - 1\right) m^2} \underbrace{\mathbb{E} \left[ \operatorname{Cov} \left( k \left( X_{11}, Y_{21} \right), k \left( X_{11}, Y_{21} \right) \mid P_1, Q_2 \right) \right]}_{\eta_5 \coloneqq } \\
    & \quad \quad + \frac{m-1}{n \left(n - 1\right) m^2} \underbrace{\mathbb{E} \left[ \operatorname{Cov} \left( k \left( X_{11}, Y_{21} \right), k \left( X_{12}, Y_{21} \right) \mid P_1, Q_2 \right) \right]}_{\eta_6 \coloneqq } \\
    & \quad \quad + \frac{m-1}{n \left(n - 1\right) m^2} \underbrace{\mathbb{E} \left[ \operatorname{Cov} \left( k \left( X_{11}, Y_{21} \right), k \left( X_{11}, Y_{22} \right) \mid P_1, Q_2 \right) \right]}_{\eta_7 \coloneqq }. \\
\label{eq:var_cov3}
\end{split}
\end{equation}

Next, we have analogous to Equation \eqref{eq:var_var4}

\begin{equation}
\begin{split}
    (\mathrm{IVb}) & = \mathbb{E} \left[ \frac{1}{n^2 \left(n - 1\right)^2} \sum_{i=1}^n \sum_{\substack{s=1 \\ s \neq i}}^n \sum_{\substack{b=1 \\ b \neq i \\ b \neq s}}^n \operatorname{Cov} \left( \frac{1}{m^2} \sum_{j=1}^m \sum_{t=1}^m k \left( X_{ij}, Y_{st} \right), \frac{1}{m^2} \sum_{c=1}^m \sum_{d=1}^m k \left( X_{ic}, Y_{bd} \right) \mid P_i, Q_s, Q_b \right) \right] \\
    & = \mathbb{E} \left[ \frac{1}{n^2 \left(n - 1\right)^2 m^4} \sum_{i=1}^n \sum_{\substack{s=1 \\ s \neq i}}^n \sum_{\substack{b=1 \\ b \neq i \\ b \neq s}}^n \sum_{j=1}^m \sum_{t=1}^m \sum_{d=1}^m \operatorname{Cov} \left( k \left( X_{ij}, Y_{st} \right), k \left( X_{ij}, Y_{bd} \right) \mid P_i, Q_s, Q_b \right) \right] \\
    \overset{\text{iid}}&{=} \mathbb{E} \left[ \frac{1}{n^2 \left(n - 1\right)^2 m^4} \sum_{i=1}^n \sum_{\substack{s=1 \\ s \neq i}}^n \sum_{\substack{b=1 \\ b \neq i \\ b \neq s}}^n \sum_{j=1}^m \sum_{t=1}^m \sum_{d=1}^m \operatorname{Cov} \left( k \left( X_{11}, Y_{21} \right), k \left( X_{11}, Y_{31} \right) \mid P_1, Q_2, Q_3 \right) \right] \\
    \overset{\text{}}&{=} \frac{n-2}{n \left(n - 1\right) m} \underbrace{\mathbb{E} \left[ \operatorname{Cov} \left( k \left( X_{11}, Y_{21} \right), k \left( X_{11}, Y_{31} \right) \mid P_1, Q_2, Q_3 \right) \right]}_{\eta_9 \coloneqq} \\
\label{eq:var_cov4}
\end{split}
\end{equation}

and in an almost identical manner

\begin{equation}
\begin{split}
    (\mathrm{IVc}) & = \mathbb{E} \left[ \frac{1}{n^2 \left(n - 1\right)^2} \sum_{i=1}^n \sum_{\substack{s=1 \\ s \neq i}}^n \sum_{\substack{a=1 \\ a \neq s \\ a \neq i}}^n \operatorname{Cov} \left( \frac{1}{m^2} \sum_{j=1}^m \sum_{t=1}^m k \left( X_{ij}, Y_{st} \right), \frac{1}{m^2} \sum_{c=1}^m \sum_{d=1}^m k \left( X_{ac}, Y_{sd} \right) \mid P_i, P_a, Q_s \right) \right] \\
    & = \mathbb{E} \left[ \frac{1}{n^2 \left(n - 1\right)^2 m^4} \sum_{i=1}^n \sum_{\substack{s=1 \\ s \neq i}}^n \sum_{\substack{b=1 \\ b \neq i \\ b \neq s}}^n \sum_{j=1}^m \sum_{t=1}^m \sum_{d=1}^m \operatorname{Cov} \left( k \left( X_{ij}, Y_{st} \right), k \left( X_{ac}, Y_{sd} \right) \mid P_i, P_a, Q_s \right) \right] \\
    \overset{\text{iid}}&{=} \mathbb{E} \left[ \frac{1}{n^2 \left(n - 1\right)^2 m^4} \sum_{i=1}^n \sum_{\substack{s=1 \\ s \neq i}}^n \sum_{\substack{b=1 \\ b \neq i \\ b \neq s}}^n \sum_{j=1}^m \sum_{t=1}^m \sum_{d=1}^m \operatorname{Cov} \left( k \left( X_{11}, Y_{31} \right), k \left( X_{21}, Y_{31} \right) \mid P_1, P_2, Q_3 \right) \right] \\
    \overset{\text{}}&{=} \frac{n-2}{n \left(n - 1\right) m} \underbrace{\mathbb{E} \left[ \operatorname{Cov} \left( k \left( X_{11}, Y_{31} \right), k \left( X_{21}, Y_{31} \right) \mid P_1, P_2, Q_3 \right) \right]}_{\eta_{10} \coloneqq}. \\
\label{eq:var_cov5}
\end{split}
\end{equation}

The only term left is
\begin{equation}
\begin{split}
    & \operatorname{Cov} \left( \frac{1}{n m^2} \sum_{i=1}^n \sum_{j=1}^m \sum_{t=1}^m k \left( X_{ij}, Y_{it} \right),  \frac{1}{n \left( n-1 \right) m^2} \sum_{i=1}^n \sum_{j=1}^m \sum_{t=1}^m \sum_{\substack{s=1 \\ s \neq i}}^n k \left( X_{ij}, Y_{st} \right) \right) \\
    & = \frac{1}{n^2 \left(n - 1\right) m^4} \sum_{i=1}^n \sum_{j=1}^m \sum_{\substack{t=1}}^m \sum_{o=1}^n \sum_{p=1}^m \sum_{\substack{s=1 \\ s \neq o}}^n \sum_{r=1}^m \operatorname{Cov} \left( k \left( X_{ij}, Y_{it} \right), k \left( X_{op}, Y_{sr} \right) \right) \\
    \overset{\text{(i)}}&{=} \frac{1}{n^2 \left(n - 1\right) m^4} \sum_{i=1}^n \sum_{j=1}^m \sum_{\substack{t=1 \\ t \neq j}}^m \sum_{p=1}^m \sum_{\substack{s=1 \\ s \neq i}}^n \sum_{r=1}^m \operatorname{Cov} \left( k \left( X_{ij}, Y_{it} \right), k \left( X_{ip}, Y_{sr} \right) \right) \\
    & \quad \quad + \frac{1}{n^2 \left(n - 1\right) m^4} \sum_{i=1}^n \sum_{j=1}^m \sum_{\substack{t=1 \\ t \neq j}}^m \sum_{p=1}^m \sum_{\substack{s=1 \\ s \neq i}}^n \sum_{r=1}^m \operatorname{Cov} \left( k \left( X_{ij}, Y_{it} \right), k \left( X_{sp}, Y_{ir} \right) \right) \\
    \overset{\text{iid}}&{=} \frac{1}{n m^3} \sum_{j=1}^m \sum_{\substack{t=1 \\ t \neq j}}^m \sum_{p=1}^m \operatorname{Cov} \left( k \left( X_{1j}, Y_{1t} \right), k \left( X_{1p}, Y_{21} \right) \right) + \frac{1}{n m^3} \sum_{j=1}^m \sum_{\substack{t=1 \\ t \neq j}}^m \sum_{r=1}^m \operatorname{Cov} \left( k \left( X_{1j}, Y_{1t} \right), k \left( X_{21}, Y_{1r} \right) \right) \\
    \overset{\text{}}&{=} \frac{1}{n m^3} \sum_{j=1}^m \sum_{\substack{t=1 \\ t \neq j}}^m \operatorname{Cov} \left( k \left( X_{1j}, Y_{1t} \right), k \left( X_{1j}, Y_{21} \right) \right) + \frac{1}{n m^3} \sum_{j=1}^m \sum_{\substack{t=1 \\ t \neq j}}^m \operatorname{Cov} \left( k \left( X_{1j}, Y_{1t} \right), k \left( X_{1t}, Y_{21} \right) \right) \\
    & \quad \quad + \frac{1}{n m^3} \sum_{j=1}^m \sum_{\substack{t=1 \\ t \neq j}}^m \sum_{\substack{p=1 \\ p \neq j \\ p \neq t}}^m \operatorname{Cov} \left( k \left( X_{1j}, Y_{1t} \right), k \left( X_{1p}, Y_{21} \right) \right) + \frac{1}{n m^3} \sum_{j=1}^m \sum_{\substack{t=1 \\ t \neq j}}^m \operatorname{Cov} \left( k \left( X_{1j}, Y_{1t} \right), k \left( X_{21}, Y_{1j} \right) \right) \\
    & \quad \quad + \frac{1}{n m^3} \sum_{j=1}^m \sum_{\substack{t=1 \\ t \neq j}}^m \operatorname{Cov} \left( k \left( X_{1j}, Y_{1t} \right), k \left( X_{21}, Y_{1t} \right) \right) + \frac{1}{n m^3} \sum_{j=1}^m \sum_{\substack{t=1 \\ t \neq j}}^m \sum_{\substack{p=1 \\ p \neq j \\ p \neq t}}^m \operatorname{Cov} \left( k \left( X_{1j}, Y_{1t} \right), k \left( X_{21}, Y_{1p} \right) \right) \\
    \overset{\text{iid}}&{=} \frac{1}{n m^3} \sum_{j=1}^m \sum_{\substack{t=1 \\ t \neq j}}^m \operatorname{Cov} \left( k \left( X_{11}, Y_{12} \right), k \left( X_{11}, Y_{21} \right) \right) + \frac{1}{n m^3} \sum_{j=1}^m \sum_{\substack{t=1 \\ t \neq j}}^m \operatorname{Cov} \left( k \left( X_{11}, Y_{12} \right), k \left( X_{12}, Y_{21} \right) \right) \\
    & \quad \quad + \frac{1}{n m^3} \sum_{j=1}^m \sum_{\substack{t=1 \\ t \neq j}}^m \sum_{\substack{p=1 \\ p \neq j \\ p \neq t}}^m \operatorname{Cov} \left( k \left( X_{11}, Y_{12} \right), k \left( X_{13}, Y_{21} \right) \right) + \frac{1}{n m^3} \sum_{j=1}^m \sum_{\substack{t=1 \\ t \neq j}}^m \operatorname{Cov} \left( k \left( X_{11}, Y_{12} \right), k \left( X_{21}, Y_{11} \right) \right) \\
    & \quad \quad + \frac{1}{n m^3} \sum_{j=1}^m \sum_{\substack{t=1 \\ t \neq j}}^m \operatorname{Cov} \left( k \left( X_{11}, Y_{12} \right), k \left( X_{21}, Y_{12} \right) \right) + \frac{1}{n m^3} \sum_{j=1}^m \sum_{\substack{t=1 \\ t \neq j}}^m \sum_{\substack{p=1 \\ p \neq j \\ p \neq t}}^m \operatorname{Cov} \left( k \left( X_{11}, Y_{12} \right), k \left( X_{21}, Y_{13} \right) \right) \\
    \overset{\text{}}&{=} \frac{m-1}{n m^2} \underbrace{\left( \operatorname{Cov} \left( k \left( X_{11}, Y_{12} \right), k \left( X_{11}, Y_{21} \right) \right) + \operatorname{Cov} \left( k \left( X_{11}, Y_{12} \right), k \left( X_{21}, Y_{11} \right) \right) \right)}_{\eta_{11} \coloneqq} \\
    & \quad \quad + \frac{m-1}{n m^2} \underbrace{\left( \operatorname{Cov} \left( k \left( X_{11}, Y_{12} \right), k \left( X_{21}, Y_{12} \right) \right) + \operatorname{Cov} \left( k \left( X_{11}, Y_{12} \right), k \left( X_{12}, Y_{21} \right) \right) \right)}_{\eta_{12} \coloneqq} \\
    & \quad \quad + \frac{\left( m-1 \right)\left( m-2 \right)}{n m^2} \underbrace{ \left(\operatorname{Cov} \left( k \left( X_{11}, Y_{12} \right), k \left( X_{13}, Y_{21} \right) \right) + \operatorname{Cov} \left( k \left( X_{11}, Y_{12} \right), k \left( X_{21}, Y_{13} \right) \right) \right)}_{\eta_{13} \coloneqq}. \\
\label{eq:var_cov6}
\end{split}
\end{equation}

By combining all previous equations, we get

\begin{equation}
\begin{split}
    & \mathbb{V} \left( \widehat{\operatorname{Cov}}_k^{\left(n, m \right)} \left( \mathbf{X}, \mathbf{Y} \right) \right) \\
    & = \underbrace{\frac{1}{n} \eta_1 + \frac{n-2}{n \left(n-1 \right)} \eta_4 + \frac{-2 \left( m-1 \right) \left( m - 2 \right)}{nm^2} \eta_{13}}_{\mathcal{O} \left( \frac{1}{n} \right)} + \underbrace{\frac{1}{n \left(n-1\right)} \eta_3}_{\mathcal{O} \left( \frac{1}{n^2} \right)} \\
    & \quad \quad + \underbrace{\frac{-2 \left( m-1 \right)}{nm^2} \left( \eta_{11} + \eta_{12} \right) + \frac{n-2}{n \left(n - 1\right) m} \left( \eta_9 + \eta_{10} \right)}_{\mathcal{O} \left( \frac{1}{nm} \right)} \\
    & \quad \quad + \underbrace{\frac{1}{n m^2} \eta_2}_{\mathcal{O} \left( \frac{1}{nm^2} \right)} + \underbrace{\frac{m-1}{n \left(n - 1\right) m^2} \left( \eta_6 + \eta_7 \right)}_{\mathcal{O} \left( \frac{1}{n^2 m} \right)} + \underbrace{\frac{1}{n \left(n - 1\right) m^2} \eta_5}_{\mathcal{O} \left( \frac{1}{n^2 m^2} \right)}. \\
\label{eq:var_cov_result}
\end{split}
\end{equation}


\end{document}